\RequirePackage{caption}

\documentclass[journal,lettersize,twoside]{IEEEtran}
\usepackage{amsmath,amsfonts}
\usepackage{algorithmic}
\usepackage{algorithm}
\usepackage{array}
\usepackage[caption=false,font=footnotesize]{subfig}
\usepackage{textcomp}
\usepackage{stfloats}
\usepackage{url}
\usepackage{verbatim}
\usepackage{graphicx}
\usepackage{cite}

\usepackage{tikz}
\usetikzlibrary{arrows}
\usepackage{smartdiagram}
\usesmartdiagramlibrary{additions} 

\newcommand\copyrighttext{%
\footnotesize \copyright 2022 IEEE. Personal use of this material is permitted.  Permission from IEEE must be obtained for all other uses, in any current or future media, including reprinting/republishing this material for advertising or promotional purposes, creating new collective works, for resale or redistribution to servers or lists, or reuse of any copyrighted component of this work in other works.\\
\url{https://doi.org/10.1109/TRO.2022.3181004}
}
\newcommand\copyrightnotice{%
	\begin{tikzpicture}[remember picture,overlay]
		\node[anchor=south,yshift=0pt] at (current page.south) {\fbox{\parbox{\dimexpr\textwidth-\fboxsep-\fboxrule\relax}{\copyrighttext}}};
	\end{tikzpicture}%
}

\begin{document}

\title{Innate Motivation for Robot Swarms by \\ Minimizing Surprise: From Simple Simulations \\ to Real-World Experiments}

\author{Tanja~Katharina~Kaiser and~Heiko~Hamann%
\thanks{T. K. Kaiser and H. Hamann are with the Institute of Computer Engineering, University of L\"ubeck, Germany, e-mail: \{kaiser,hamann\}@iti.uni-luebeck.de.}
}

\markboth{}%
{Kaiser and Hamann: Innate Motivation for Robot Swarms by Minimizing Surprise}

\IEEEpubid{}

\maketitle
\copyrightnotice 

\begin{abstract}
Applications of large-scale mobile multi-robot systems can be beneficial over monolithic robots because of higher potential for robustness and scalability.
Developing controllers for multi-robot systems is challenging because the multitude of interactions is hard to anticipate and difficult to model.
Automatic design using machine learning or evolutionary robotics seem to be options to avoid that challenge, but bring the challenge of designing reward or fitness functions. 
Generic reward and fitness functions seem unlikely to exist and task-specific rewards often have undesired side effects. 
Approaches of so-called innate motivation try to avoid the specific formulation of rewards and work instead with different drivers, such as curiosity. 
Our approach to innate motivation is to minimize surprise, which we implement by maximizing the accuracy of the swarm robot's sensor predictions using neuroevolution. 
A~unique advantage of the swarm robot case is that swarm members populate the robot's environment and can trigger more active behaviors in a self-referential loop.
We summarize our previous simulation-based results concerning behavioral diversity, robustness, scalability, and engineered self-organization, and put them into context.
In several new studies, we analyze the influence of the optimizer's hyperparameters, the scalability of evolved behaviors, and the impact of realistic robot simulations. Finally, we present results using real robots that show how the reality gap can be bridged. 
\end{abstract}


\section{Introduction}

\IEEEPARstart{R}{obot} swarms are decentralized collective systems consisting of simple embodied agents that act autonomously and rely on local information only~\cite{Hamann2018}.
These large-scale multi-robot systems generally have potential for increased robustness and scalability over single-robot systems. 
The global, collective behavior (i.e., macro-level) emerges from local interactions between individual robots and between robots and the environment (i.e., micro-level)~\cite{brambilla2013swarm}. 
This makes the manual design of swarm robotics systems challenging as the desired collective behavior is defined on the macro-level, but the robot controller has to be implemented on the micro-level while considering hard to anticipate interactions and feedback processes~\cite{bonabeau1999swarm}. 
A~promising alternative is the automatic design of swarm robot controllers~\cite{Francesca2016}.
For this purpose, methods of machine learning can be used~\cite{panait2005cooperative}, but researchers most frequently rely on methods of evolutionary robotics~\cite{nolfi2000evolutionary,bongard13,trianni08}. 
The automatic design process is guided by the optimization of a fitness function.
Usually, task-specific fitness functions are used that quantify the intended task or behavior. 
But the specification of goal-directed fitness functions is difficult, as evolutionary algorithms could potentially maximize fitness in every possible way, which might lead to unexpected, unwanted behaviors when the fitness function is not specified accurately enough~\cite{doncieux2014beyond} or original solutions may not be found, as too specific behavioral features are rewarded, which restricts the search space~\cite{nelson2009fitness}. 
A~different strategy is to use a task-independent fitness function which puts selection pressure for aspects that are not directly related to a desired task.
For example, those can be implicit fitness functions, that is, agents have to survive long enough to be able to reproduce, or use information-theoretic measures as rewards to quantify how interesting behaviors are based on an agent's sensor and motor values~\cite{doncieux2014beyond}. 
While the evolutionary process has the freedom to find original solutions here, there is no guarantee that desired behaviors are found. 

\enlargethispage{1\baselineskip}
In our minimize surprise approach~\cite{hamann14a}, we evolve swarm behaviors with a task-independent fitness function as an innate motivation. 
We are loosely inspired by Friston's information-theoretic `free-energy principle'~\cite{friston10}.
It states that organisms constantly try to minimize free energy, which is the deviation between observed and predicted sensor values in the simplest case and thus is considered as prediction error or surprise~\cite{friston2010action}. 
The minimization of free energy can then be achieved by an organism either adjusting its actions, such that they lead to sensor values matching the predictions, or optimizing the internal world model or 
predictor~\cite{Schwartenbeck2013}. 
Approaches based on the minimization of surprise are often criticized on the basis of the `Dark-Room Problem'~\cite{friston12}. 
This problem states that, while the easiest way to minimize surprises is to search for and stay in a dark, unchanging room, organisms do not do so. 
However, organisms will be surprised by a dark room if their world model does not expect it. 
\IEEEpubidadjcol
Thus, surprise is minimized for an agent's own econiche or, in our case, the world in which agents live. 
In contrast to a scenario with a single agent living in a static environment, swarm members populate an agent's environment in our approach. 
This can trigger more active behaviors in a self-referential loop as an agent's sensor inputs, that is, 
its perceptions, are based both on its own actions and the actions of other swarm members. 
Here, we use the minimization of surprise or, put differently, the maximization of prediction accuracy, as selection pressure in the evolutionary process.  
We equip our homogeneous robot swarms with actor-predictor pairs of artificial neural networks (ANN). 
Direct selection pressure from minimizing surprise is applied only on the predictor while the actor indirectly receives selection pressure by being paired with the predictor. 
Surprise is minimized over generations and swarm behaviors emerge as a by-product~\cite{hamann14a}. 

\begin{figure}
\includegraphics[width=\linewidth]{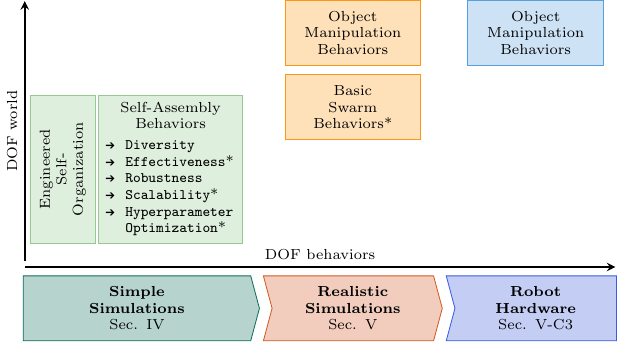}
    \caption{Overview of this paper's studies. We showcase our minimize surprise approach in scenarios of increasing complexity: simple simulations, realistic simulations, and on real robot hardware. In our simple simulations, we allow different degrees of freedom (DOF) for behaviors and  observe major advantages, such as behavior diversity, scalability, and robustness. Experiments in realistic simulations are run in environments with and without manipulable boxes. We conclude with real-world multi-robot experiments. Studies presented for the first time in this paper are marked with a * \cite{kaiser2021a}. \label{fig:studyOverview}}
\end{figure}

With this paper we summarize and extend our work on evolving swarm behaviors with our minimize surprise approach to give a holistic overview and perspective~\cite{kaiser18,kaiser19c,kaiser20a,kaiser20b,kaiser21}, as visualized in Fig.~\ref{fig:studyOverview}.
We discuss related work in Sec.~\ref{section:relatedwork} and introduce the minimize surprise approach in Sec.~\ref{section:approach}. 
In Sec.~\ref{section:AnalysisApproach}, we provide an in-depth analysis of the approach based on a sample scenario in which we aim for the emergence of self-assembly behaviors~\cite{Whitesides2418} on 2D~torus grids~\cite{kaiser18,kaiser19c,kaiser20a,kaiser20b}.
Sec.~\ref{section:SAscenario} presents the experimental setup, metrics for the classification of the emergent behaviors, and used statistical tests as used in previous works~\cite{kaiser18,kaiser19c}. 
We summarize our previous results on emergent behaviors (Secs.~\ref{section:SApredictionaccuracy}~-~\ref{section:SApredictions})~\cite{kaiser19c,kaiser20b}, their robustness regarding sensor noise and damage (Sec.~\ref{section:SArobustness})~\cite{kaiser19c}, engineering self-organization to push evolution towards the emergence of desired behaviors (Sec.~\ref{section:SAengineeredSO})~\cite{kaiser18,kaiser19c}, and comparing the behavioral diversity generated by minimize surprise to novelty search (Sec.~\ref{section:SANovelty})~\cite{kaiser20a,kaiser20b}.
In this paper, we provide for the first time a comprehensive in-depth analysis of our approach by including several new studies with significant novel contributions:
a comparison of the evolved ANN pairs and randomly generated ANN pairs (Sec.~\ref{section:SAeffectivity}) to show that minimize surprise outperforms random search, a study of scalability of emergent behaviors in swarm density (Sec.~\ref{section:SAscalability}), and a study on the influence of the hyperparameters of the evolutionary algorithm on the resulting ANN pairs (Sec.~\ref{section:SAhyperparameter}). 
In Sec.~\ref{section:reality}, we cross the reality gap by evolving swarm behaviors with minimize surprise in realistic simulations and on real robots. 
We present new results on the evolution of basic collective behaviors in realistic simulations (Sec.~\ref{section:REALbasic}) and summarize our previous results using realistic simulations and real robots for object manipulation behaviors 
(Sec.~\ref{section:REALbp})~\cite{kaiser21}.
By adding the new scenario, we show that our approach generally works in the real world.
The paper concludes with a summary and a discussion of future work.
All figures, a video~\cite{kaiser2021a}, and the code for all experiments\footnote{https://gitlab.iti.uni-luebeck.de/minimize-surprise} are online.

\section{Related Work} \label{section:relatedwork}

In this section, we present three categories of work related to our minimize surprise approach: evolutionary swarm robotics (Sec.~\ref{section:EvoRobo}), evolutionary divergent search (Sec.~\ref{section:divergentsearch}), and  intrinsically motivated learning (Sec.~\ref{section:intrinsicmotivation}). 

\subsection{Evolutionary Swarm Robotics} \label{section:EvoRobo}

Evolutionary swarm robotics~\cite{trianni08} applies evolutionary robotics~\cite{nolfi2000evolutionary} to swarm robotics~\cite{Hamann2018} for the automatic generation of swarm robot controllers~\cite{Francesca2016}. 
Most evolutionary algorithms are gradient-free, that is, optimization is done in a trial-and-error style~\cite{Eiben2015}.  
We optimize ANNs as robot controllers using evolutionary algorithms~\cite{Eiben2015} in our minimize surprise approach.
Other possible control architectures include behavior trees~\cite{Jones2019} and finite state machines~\cite{hecker2012}. 
A~variety of swarm behaviors were successfully evolved both in simulation and on real robots, examples are aggregation and coordinated motion~\cite{dorigo2004evolving}, self-assembly~\cite{gross2006}, and foraging~\cite{Jones2019}.
In our work, we aim for self-assembly behaviors (Sec.~\ref{section:AnalysisApproach}), basic swarm behaviors, and object manipulation behaviors (Sec.~\ref{section:reality}). 

In offline evolution, controllers are first optimized in simulation and then transferred to the real robot for task execution~\cite{silva2014}. 
Optimizing robot controllers completely in simulation speeds up the search process and prevents robot hardware from wearing out. 
But the reality gap~\cite{jakobi1995} is a challenge.
As simulations differ from the real world, evolution may exploit simulation-specific features and
the performance on the real robot may be worse than in simulation. 
A potential solution is to conduct some or all evaluations on real robots, but then we lose the optimization advantages from simulation.  

In online evolution, controllers are evolved on real robots in their operational environment during task execution, which avoids the reality gap~\cite{silva2014}. 
It allows robots to adapt online to environmental or task-related changes. 
While our first works were done in simple simulations (Sec.~\ref{section:AnalysisApproach}), we also show that swarm behaviors can be evolved online in realistic simulations and on real robots using our minimize surprise approach~(Sec.~\ref{section:reality}). 

\subsection{Evolutionary Divergent Search} \label{section:divergentsearch}

Open-ended evolution leads to constant morphological and behavioral innovation in nature~\cite{doncieux2015evolutionary} and is considered one of the grand open challenges of artificial life~\cite{bedau2000}. 
Inspired by this open-endedness, divergent search mechanisms push towards behavioral diversity instead of a task-specific objective in the evolutionary process~\cite{pughQA2016}.
Novelty search~\cite{lehman2011}, 
for example, aims for behavioral diversity by driving the search process using the novelty of individuals with respect to current and past individuals. 
In contrast, surprise search~\cite{gravina2016} rewards the deviation from expected behaviors. 
Quality-Diversity (QD) algorithms~\cite{pughQA2016}, such as MAP-Elites~\cite{mouret2015illuminating} and novelty search with local competition~\cite{lehmann2011_NSLC}, extend purely divergent search and guide search towards collections of behaviors that are both maximally diverse and performing high. 
These algorithms also proved to be successful in the generation of diverse swarm behaviors. 
For example, Gomes et al.~\cite{gomes13} used novelty search to evolve controllers for aggregation and resource sharing, and Cazenille et al.~\cite{cazenille2019exploring} used MAP-Elites to evolve self-assembly behaviors. 
In our minimize surprise approach, we aim for diverse swarm behaviors by using a task-independent reward. 
We compare minimize surprise and novelty search with respect to behavioral diversity to show the competitiveness of our approach in Sec.~\ref{section:SANovelty}. 

Approaches that are inspired by open-ended evolution, like novelty search, are conceptually similar to intrinsically motivated approaches in machine learning and developmental robotics~\cite{lehman2011, Oudeyer2009}. 
We discuss those in more detail next. 

\subsection{Intrinsically Motivated Learning} \label{section:intrinsicmotivation}

Intrinsically motivated artificial systems combine machine learning approaches with intrinsic motivation mechanisms to realize open-ended learning~\cite{Mirolli2013}. 
These intrinsic motivation mechanisms are inspired by the psychological concept of intrinsic motivation, that is, activities are done for the joy or challenge of performing them and not for a reward or due to external pressures~\cite{RYAN2000}. 
Computational approaches of intrinsic motivation are generally considered to be task-independent, free of semantics (i.e., meaning of sensor values), and applicable to any agent embodiment including the sensory-motoric configuration. 
Measures for intrinsic motivation can be calculated from the agent's perspective and relate to its knowledge or competence~\cite{Oudeyer2009,Mirolli2013}.  
Examples for intrinsic motivations are curiosity~\cite{schmidhuber1991possibility}, novelty~\cite{Metzen2013}, empowerment~\cite{salge2014}, homeokinesis~\cite{Der2012}, predictive information~\cite{scheunemann2020human}, and surprise that can be based on the prediction error~\cite{achiam2017surprisebased} or on how unexpected states are~\cite{berseth2020smirl}.
In most cases, model uncertainty is maximized to encourage open-ended learning. 
However, Berseth et al.~\cite{berseth2020smirl} show that the minimization of surprise (i.e., rewarding familiar states here) can also lead to increasingly complex behaviors in highly dynamic environments. 

Most computational models of intrinsic motivation use reinforcement learning in single agent scenarios both in simulation and on real robots, while swarm and multi-agent settings are only rarely studied. 
Khan et al.~\cite{Khan2018} discuss that multi-agent settings may allow for the emergence of new functionality, like communication, and for new models of motivation, like sharing motivations with other agents.
In our work, we combine an intrinsic motivation measure with methods of evolutionary computation in a swarm scenario. 
The selection pressure for evolution is calculated based on the intrinsic motivation measures summed over all swarm members.

\section{The Minimize Surprise Approach} \label{section:approach}

\tikzset{%
  every neuron/.style={
    circle,
    draw,
    minimum size=0.7cm
  },
  neuron missing/.style={
    draw=none, 
    scale=2,
    text height=0.333cm,
    execute at begin node=\color{black}$\vdots$
  },
}

\begin{figure}[t]
    \centering
    \subfloat[]{
\resizebox {0.47\linewidth} {!} {
\begin{tikzpicture}[x=0.9cm, y=0.9cm, >=stealth,]
\foreach \m/\l [count=\y] in {1,missing,2,3,missing,4}
  \node [every neuron/.try, neuron \m/.try] (input-\m) at (0,2 -\y*1.1) {};
\foreach \m [count=\y] in {1,missing,2}
  \node [every neuron/.try, neuron \m/.try ] (hidden-\m) at (1.75,0.2-\y) {};
\foreach \m [count=\y] in {1,missing,2}
  \node [every neuron/.try, neuron \m/.try ] (output-\m) at (3.5,0.2-\y) {};
\draw [<-] (input-1) -- ++(-2.5,0)
    node [above, midway] {$s_{0}(t)$};
\draw [<-] (input-2) -- ++(-2.5,0)
    node [above, midway] {$s_{R-1}(t)$};
\draw [<-] (input-3) -- ++(-2.5,0)
    node [above, midway] {$a_0(t-1)$};
\draw [<-] (input-4) -- ++(-2.5,0)
    node [above, midway] {$a_{M-1}(t-1)$};
\draw [->] (output-1) -- ++(2.0,0)
    node [above, midway] {$a_{0}(t)$};
\draw [->] (output-2) -- ++(2.0,0)
    node [above, midway] {$a_{M-1}(t)$};
\foreach \i in {1,2,3,4}
  \foreach \j in {1,...,2}
    \draw [->] (input-\i) -- (hidden-\j);
\foreach \i in {1,...,2}
  \foreach \j in {1,2}
    \draw [->] (hidden-\i) -- (output-\j);
\end{tikzpicture}\label{fig:actor} }}
\subfloat[]{
    \resizebox {0.45\linewidth} {!} {
  \begin{tikzpicture}[x=0.9cm, y=0.9cm, >=stealth,]
\foreach \m/\l [count=\y] in {1,missing,2,3,missing,4}
  \node [every neuron/.try, neuron \m/.try] (input-\m) at (0,2 -\y*1.1) {};
\foreach \m [count=\y] in {1,missing,2}
  \node [every neuron/.try, neuron \m/.try ] (hidden-\m) at (1.75,0.2-\y) {};
\foreach \m [count=\y] in {1,missing,2}
  \node [every neuron/.try, neuron \m/.try ] (output-\m) at (3.5,0.2-\y) {};
\draw [<-] (input-1) -- ++(-2.0,0)
    node [above, midway] {$s_0(t)$};
\draw [<-] (input-2) -- ++(-2.0,0)
    node [above, midway] {$s_{R-1}(t)$};
\draw [<-] (input-3) -- ++(-2.0,0)
    node [above, midway] {$a_0(t)$};
\draw [<-] (input-4) -- ++(-2.0,0)
    node [above, midway] {$a_{M-1}(t)$};
\foreach \l [count=\i] in {0, R-1}
  \draw [->] (output-\i) -- ++(2.5,0)
    node [above, midway] {$p_{\l}(t+1)$};
\foreach \i in {1,2,3,4}
  \foreach \j in {1,...,2}
    \draw [->] (input-\i) -- (hidden-\j);
\foreach \i in {1,...,2}
  \foreach \j in {1,...,2}
    \draw [->] (hidden-\i) -- (output-\j);
 \draw[->,shorten >=1pt] (hidden-1) to [out=45,in=90,loop,looseness=5.8] (hidden-1);
 \draw[->,shorten >=1pt] (hidden-2) to [out=315,in=270,loop,looseness=5.8] (hidden-2);
\end{tikzpicture}  }\label{fig:predictor}
    }
\caption{Actor-predictor ANN pair of each swarm member in minimize surprise. The actor~(a) is a single hidden layer feedforward ANN that outputs M~action values $a_0(t), \dots, a_{M-1}(t)$ (e.g. motor speeds). 
The predictor~(b) has one recurrent hidden layer and outputs R~sensor value predictions ${p_0(t+1),\dots,p_{R-1}(t+1)}$ for time step~$t+1$. Inputs are R~sensor values $s_0(t),\dots,s_{R-1}(t)$ at time step~t and all or a subset of the action values $a_0(t-1), \dots, a_{M-1}(t-1)$ of time step $t-1$ or $a_0(t), \dots, a_{M-1}(t)$ of time step $t$, respectively~\cite{kaiser2021a}.}
\label{fig:ANNs}
\end{figure}
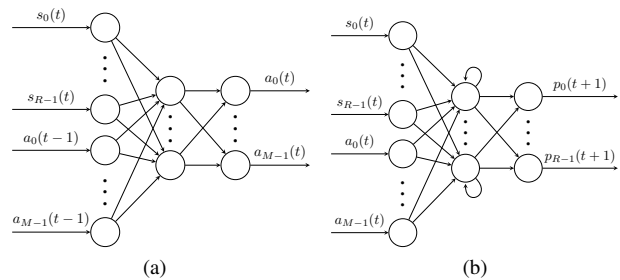

With our minimize surprise approach~\cite{hamann14a}, we aim for the evolution of diverse swarm behaviors using a task-independent fitness function. 
Each swarm member (i.e., robot or agent) is equipped with a pair of ANNs that are evolved together. 
An actor network serves as a regular controller (see~Fig.~\ref{fig:actor}). 
It is implemented as a feedforward network, which has proven to be a suitable controller for a variety of swarm robotics tasks, such as foraging~\cite{heinerman2016}, or aggregation and coordinated motion~\cite{dorigo2004evolving}.
The actor has one hidden layer and 
outputs M~action values that encode the swarm member's next action, for example, as discrete actions, like straight motion and turning, or motor speeds.  
Additionally, an ANN with one recurrent hidden layer serves as a predictor network (see Fig.~\ref{fig:predictor}) that forecasts the sensor values of the next time step. 
Recurrent neural networks have feedback connections (i.e., they have memory) and are especially suitable for sequential data, such as time-series of sensor values.
The predictor can be seen as a world model because the prediction of exteroceptive sensors equates to predicting the future state of the environment.
Inputs to both networks are the current sensor values of the swarm member. 
Additionally, all or a subset of the swarm member's last chosen action values, that is, $a_0(t-1), \dots, a_{M-1}(t-1)$ of the previous time step $t-1$, are fed into the actor ANN. 
Thereby, we make the actor recurrent and introduce limited memory, although it is implemented as a feedforward network. 
The next intended actions, that is, action values chosen in this time step, are given to the predictor ANN as additional inputs because the agent's action influences its future sensor readings. 

A powerful advantage of our minimize surprise approach is that a task-specific fitness function is not required. 
We define an innate motivation by giving reward for high prediction accuracy (i.e., minimal surprise) instead. 
We define fitness as 
\begin{equation}
F = \frac{1}{TNR} \sum_{t=0}^{T-1} \sum_{n=0}^{N-1} \sum_{r=0}^{R-1} (1 -  | p_{r}^{n}(t) - s_{r}^{n}(t) |)\, ,
\label{equ:fitness}
\end{equation}

where $T$ is the evaluation length in time steps, $N$ is the swarm size, $R$~is the number of sensors per swarm member, $p^{n}_{r}(t)$ is the prediction for sensor~$r$ of swarm member~$n$, and~$s^n_r(t)$ is the value of sensor~$r$ of swarm member~$n$ at time step~$t$.
The fitness is normalized to a theoretical maximum of~$1$.
This fitness function is generic and will stay the same for all experiments in this paper. 

While self-supervised learning~\cite{ha_2018, Nava2019} could be used to train the predictor network, we cannot manually or automatically generate labeled data to train the actor network as our approach is task-independent. 
Thus, we rely on neuroevolution of the ANN weights for the fixed topologies of both networks using a simple genetic algorithm~\cite{holland75}. 
Genomes directly encode the weights of the actor-predictor ANNs that are evolved in pairs~\cite{ronald1994}. 
We chose not to investigate the impact of sophisticated methods of evolutionary computation to simplify this study. 
We randomly generate the initial population of genomes by drawing the weights from a uniform distribution in $[-0.5,0.5]$. 
All swarm members of an evaluation share the same genome (i.e., application of the same ANN weights), that is, we use a homogeneous swarm. 
Thus, we have two different population concepts within this paper: a population of swarm members forming the homogeneous swarm in an evaluation of a genome applying the actor-predictor ANN pair as well as a population of genomes encoding the weights of ANN pairs in the evolutionary process.
Selective pressure during evolution is put on the predictor by rewarding prediction accuracy~(Eq.~\ref{equ:fitness}). 
The actor is not directly rewarded and receives only indirectly selective pressure, as it is paired with a predictor.
High prediction accuracy (i.e., fitness) is reached by an actor-predictor pair when the behaviors
resulting from the selected actions of the actor lead to sensor values as predicted by the predictor. 
In turn, higher fitness values result in a higher likelihood to survive in the evolutionary process. 
Hence, it is left to the evolutionary dynamics and generally difficult to analyze whether actors adapt agent behaviors to predictions or predictors adapt their predictions to behaviors.

\section{In-Depth Analysis of the Approach using Simple Simulations} \label{section:AnalysisApproach}

\subsection{Sample Scenario: Self-Assembly}
\label{section:SAscenario} 

Here, we define a sample scenario that allows us to study our minimize surprise approach in depth while keeping computational resources within reasonable limits. 
Sec.~\ref{section:SAExperimentalSetup} defines the experimental setup in detail, while Sec.~\ref{section:SAmetrics} provides the metrics for the analysis of the emergent behaviors. 

\subsubsection{Experimental Setup} \label{section:SAExperimentalSetup}

Self-assembly behaviors lead to the autonomous organization of system components into patterns or structures by local interactions~\cite{Whitesides2418} and were already implemented on robot swarms of up to one thousand Kilobots~\cite{Rubenstein795}.  
These behaviors are more complex than the basic collective behaviors that emerged in our first works~\cite{hamann14a, borkowski17}. 
We increase the potential complexity by aiming for self-assembly in our minimize surprise setup here. 
We govern the complexity of the study by restricting us to a simple, 2D grid world with periodic boundary conditions (i.e., a torus) to simplify sensing and equidistant agent positioning. 

\begin{figure}[t]
    \centering
    \includegraphics[width=0.2\textwidth]{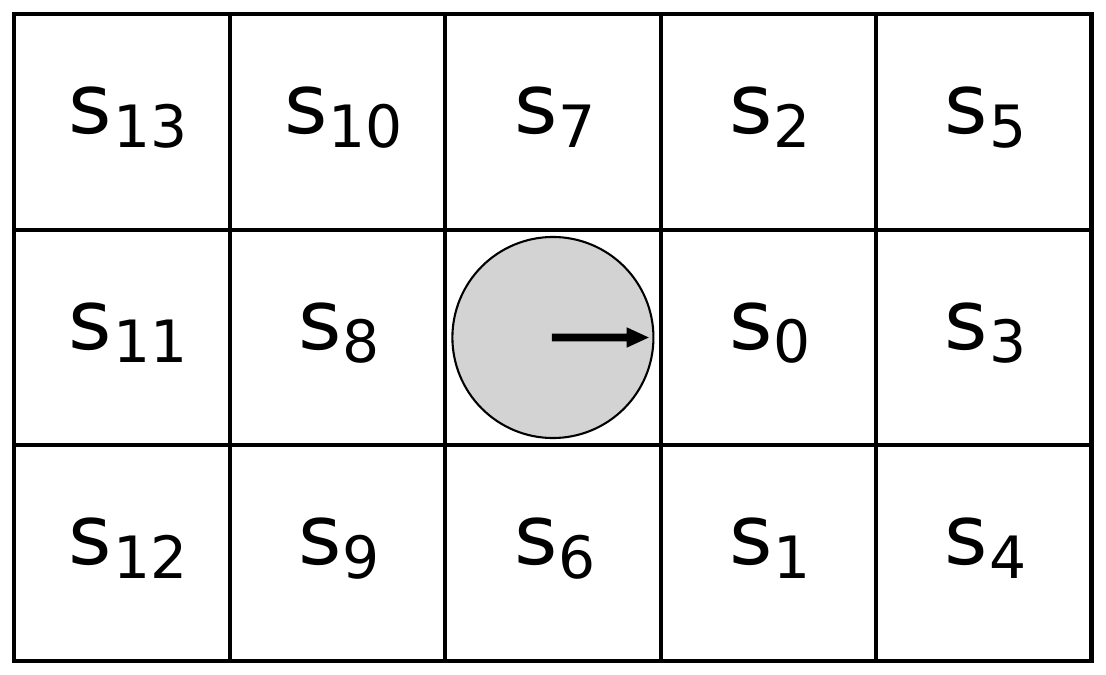}
    \caption{Sensor model for the simple simulations (Sec.~\ref{section:AnalysisApproach}) with labels per sensor. The \textit{gray circle} represents the agent and the \textit{arrow} its heading~\cite{kaiser2021a}. \label{fig:SAsensormodel}}
\end{figure}

\begin{table}[t]
\caption{Hyperparameters for the self-assembly scenario.\label{tab:SAparameters}}
\centering
\begin{tabular}{ll}
\hline
 \textbf{hyperparameter} & \textbf{value} \\ \hline 
 grid side length $L$ & $[11, 30]$ \\ 
 \# of sensors $R$& 14 \\ 
 swarm size $N$ &  100 \\ \hline 
 population size & 50 \\ 
 number of generations & 100 \\ 
 evaluation length $T$ (time steps) & 500 \\ 
 \# of simulation runs per fitness evaluation & 10 \\  
 elitism & 1 \\ 
 mutation rate & 0.1 \\  \hline 
\end{tabular}
\end{table}

We simulate a swarm of simple agents with discrete headings (i.e., North, East, South, and West) and binary sensors that cover 14~neighboring grid cells (see Fig.~\ref{fig:SAsensormodel}). 
A sensor value of `0' indicates the respective grid cell is empty while a `1' means that it is occupied by another agent. 
A study of three different sensor models and the justification for the chosen sensor model are found in~\cite{kaiser19c}.  
An agent can either rotate by $\pm 90^\circ$ or move one grid cell forward in each time step. 
A~grid cell can only be occupied by one agent at a time.
A~move forward is hence blocked if the targeted grid cell is already occupied. 
This can be compared to a hardware protection layer in real robots that prevents collisions with other robots or obstacles.

\begin{figure*}[t]
    \centering
    \subfloat[ ]{\includegraphics[width=0.19\textwidth, trim={0.8cm 0.8cm 0.8cm 0.8cm}, clip]{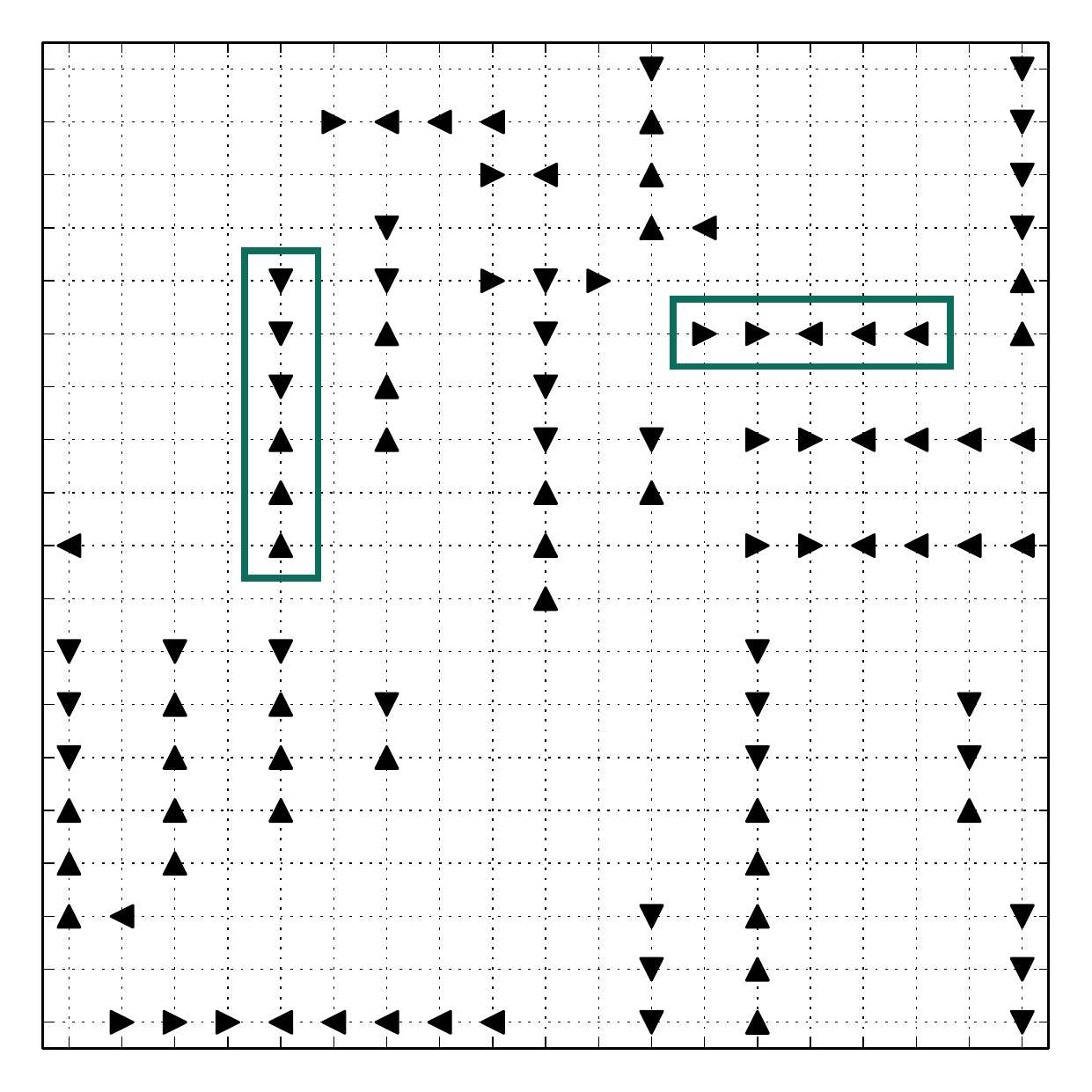}\label{fig:SAline}} 
    \hspace{1mm}
    \subfloat[ ]{\includegraphics[width=0.19\textwidth, trim={0.8cm 0.8cm 0.8cm 0.8cm}, clip]{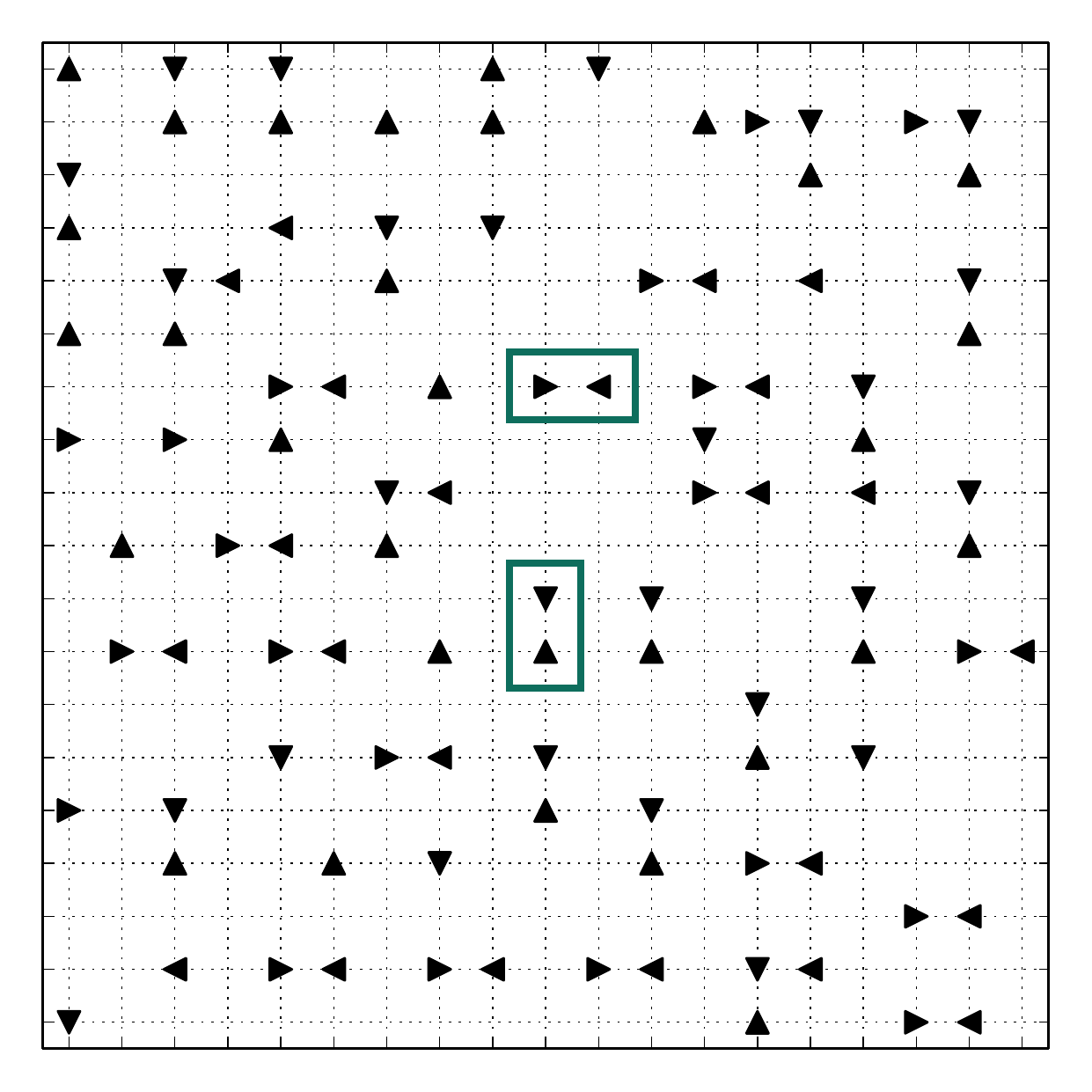}\label{fig:SApairs}}
    \hspace{1mm}
    \subfloat[ ]{\includegraphics[width=0.19\textwidth, trim={0.8cm 0.8cm 0.8cm 0.8cm}, clip]{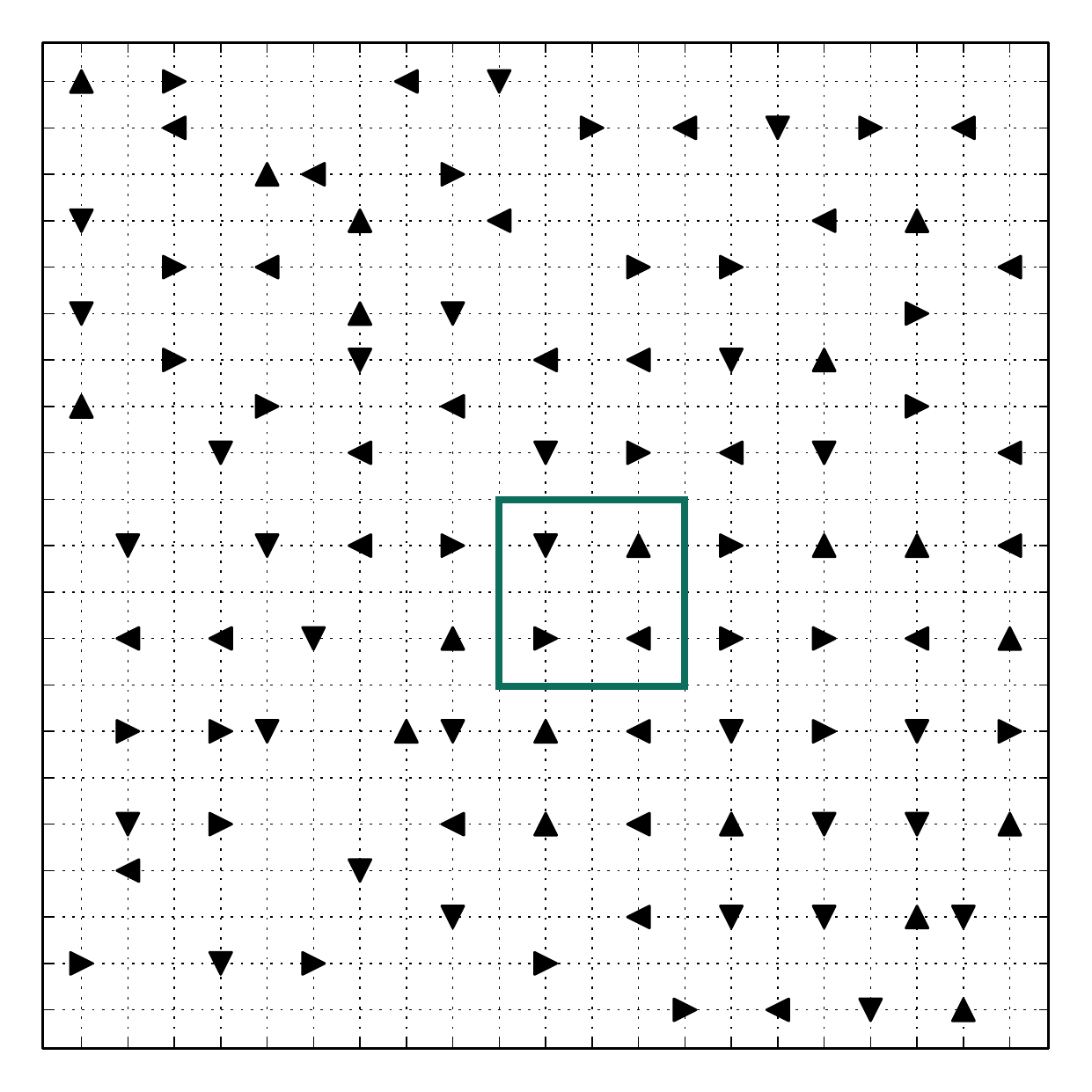}\label{fig:SAsquares}} 
    \hspace{1mm}
    \subfloat[ ]{\includegraphics[width=0.19\textwidth, trim={0.8cm 0.8cm 0.8cm 0.8cm}, clip]{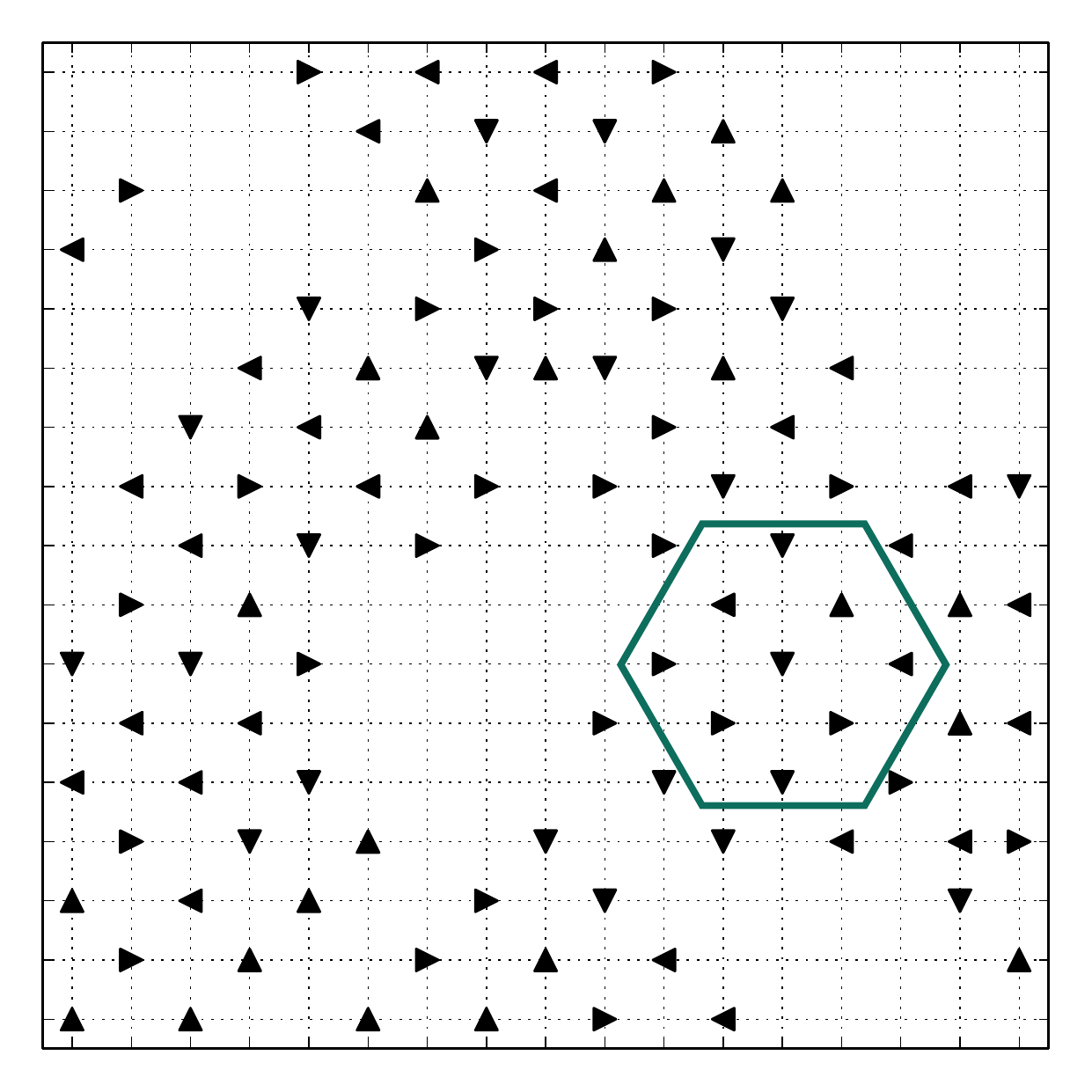}\label{fig:SAtriangular}} 
    \hspace{1mm}
    \subfloat[ ]{\includegraphics[width=0.19\textwidth, trim={0.8cm 0.8cm 0.8cm 0.8cm}, clip]{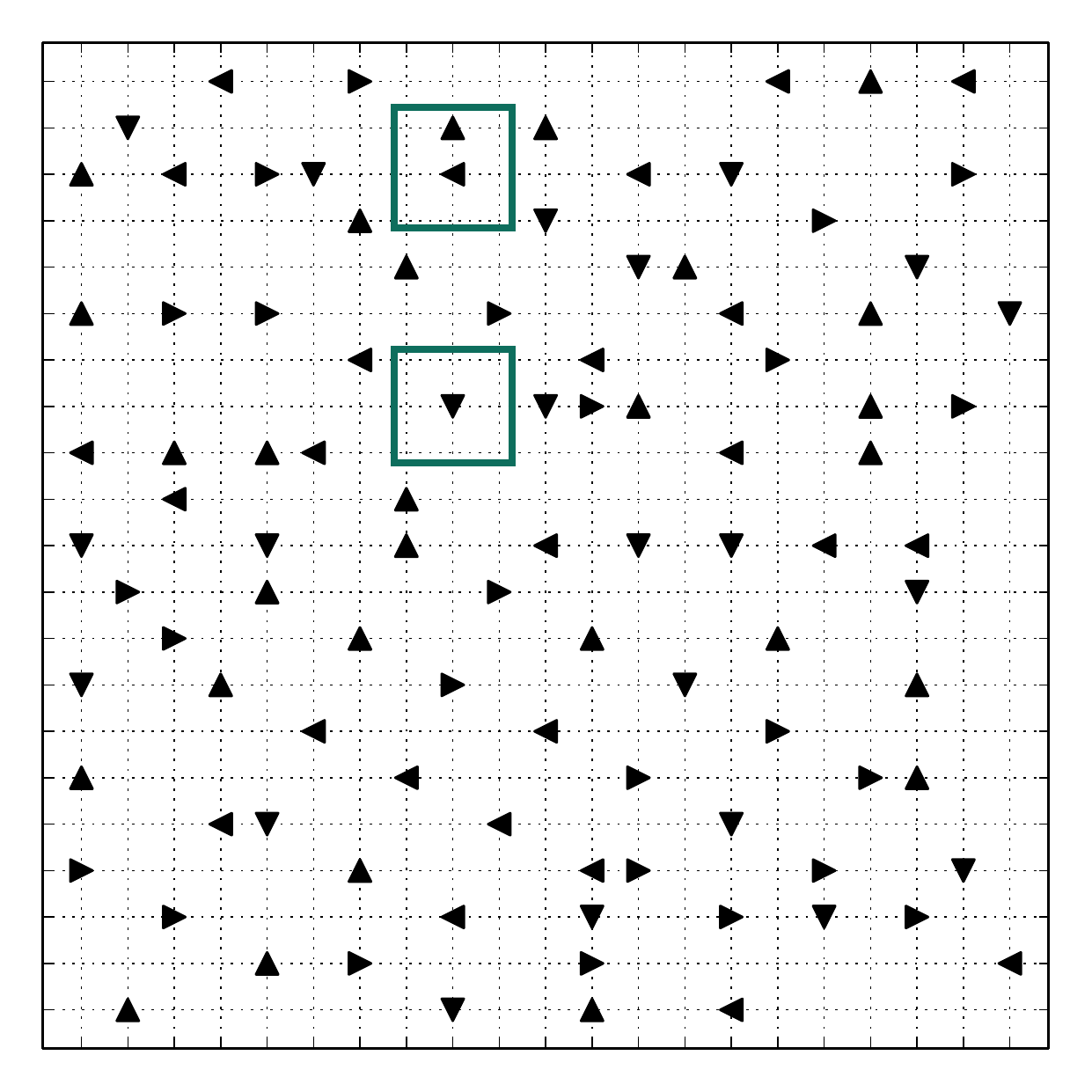}\label{fig:SArandomdisp}}\\ 
    \subfloat[ ]{\includegraphics[width=0.19\textwidth, trim={0.8cm 0.8cm 0.8cm 0.8cm}, clip]{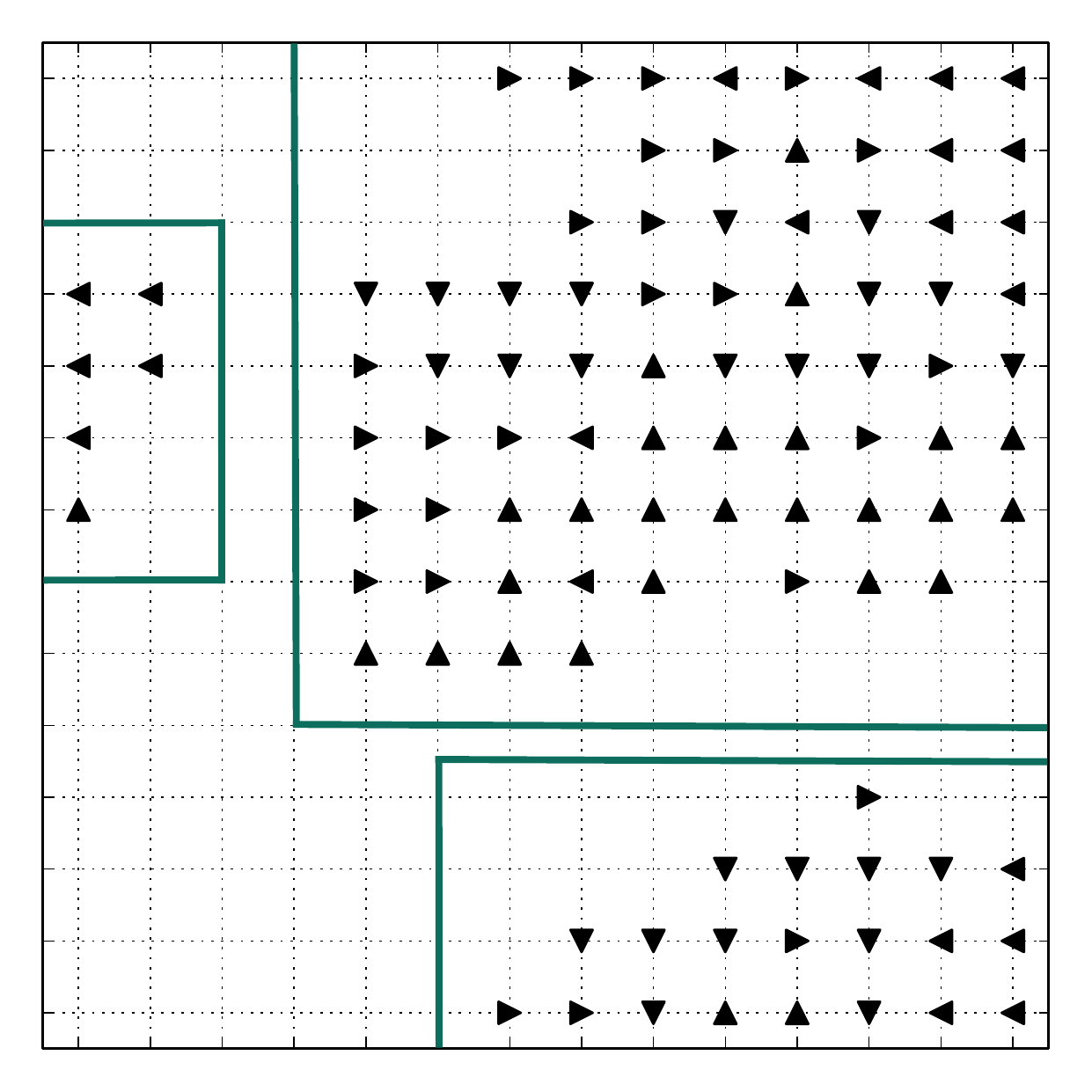}\label{fig:SAaggregation}}  
    \hspace{1mm}
    \subfloat[ ]{\includegraphics[width=0.19\textwidth, trim={0.8cm 0.8cm 0.8cm 0.8cm}, clip]{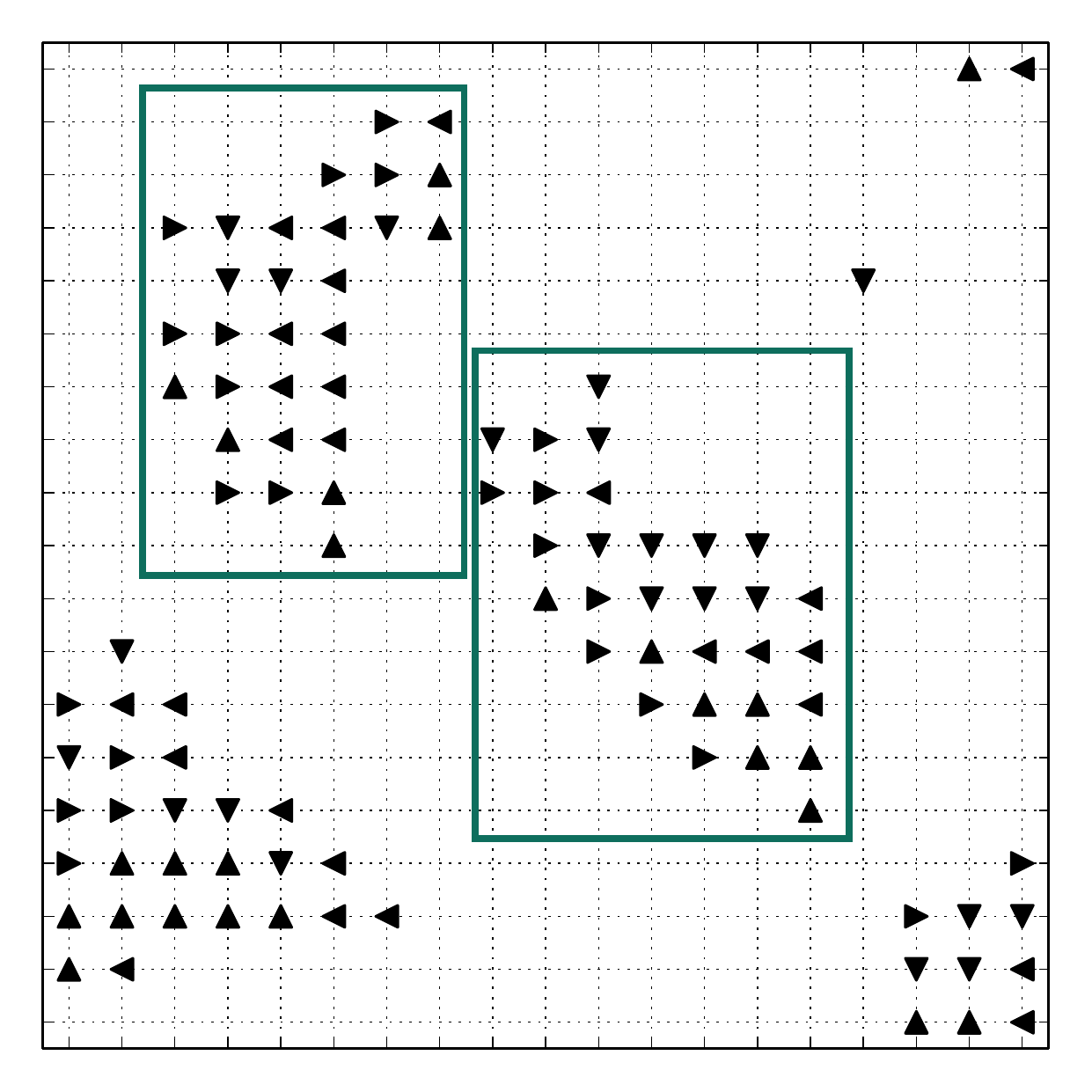}\label{fig:SAclustering}} 
   \hspace{1mm}
    \subfloat[ ]{\includegraphics[width=0.19\textwidth, trim={0.8cm 0.8cm 0.8cm 0.8cm}, clip]{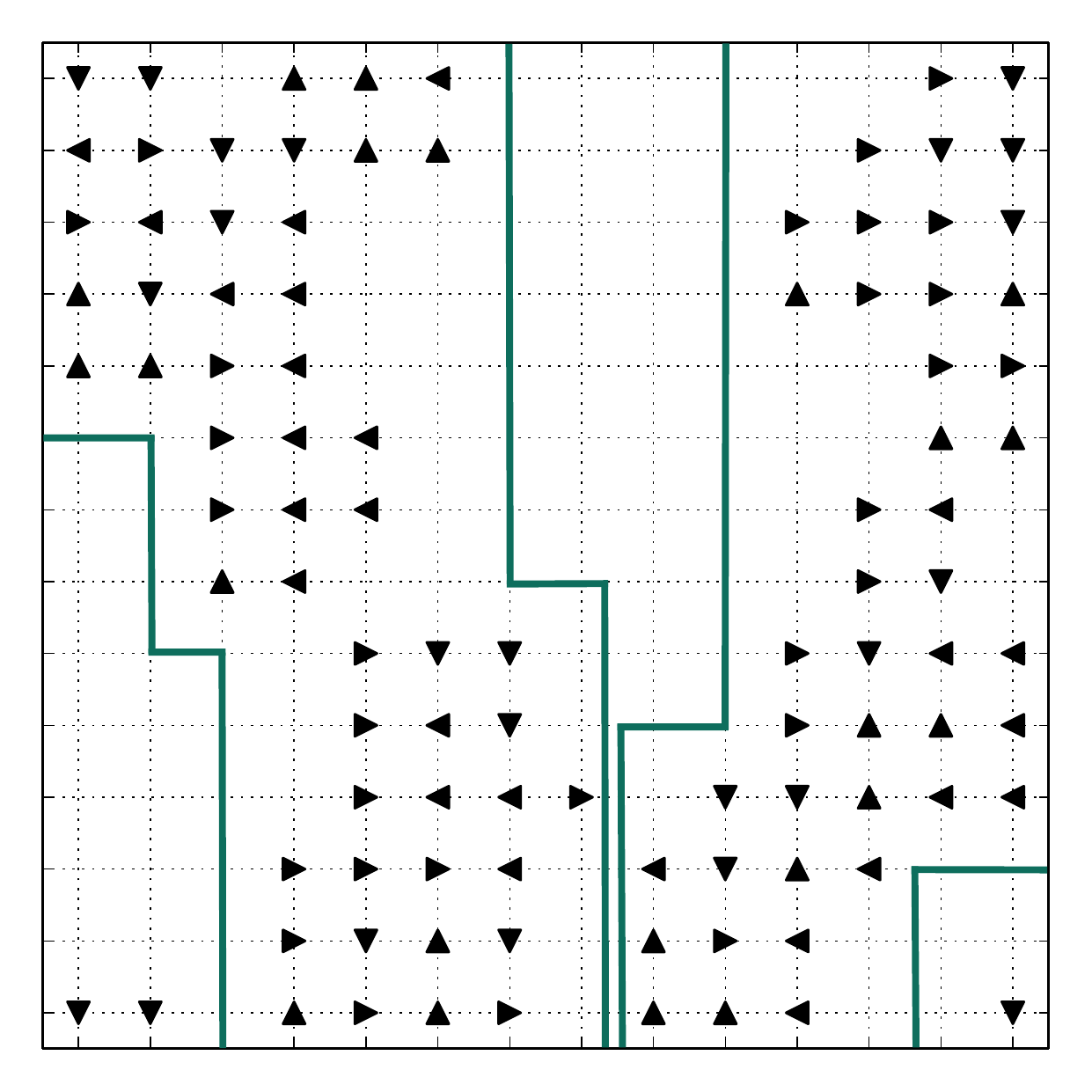}\label{fig:SAloosegroup}} 
    \hspace{1mm}
    \subfloat[ ]{\includegraphics[width=0.19
    \textwidth, trim={0.8cm 0.8cm 0.8cm 0.8cm}, clip]{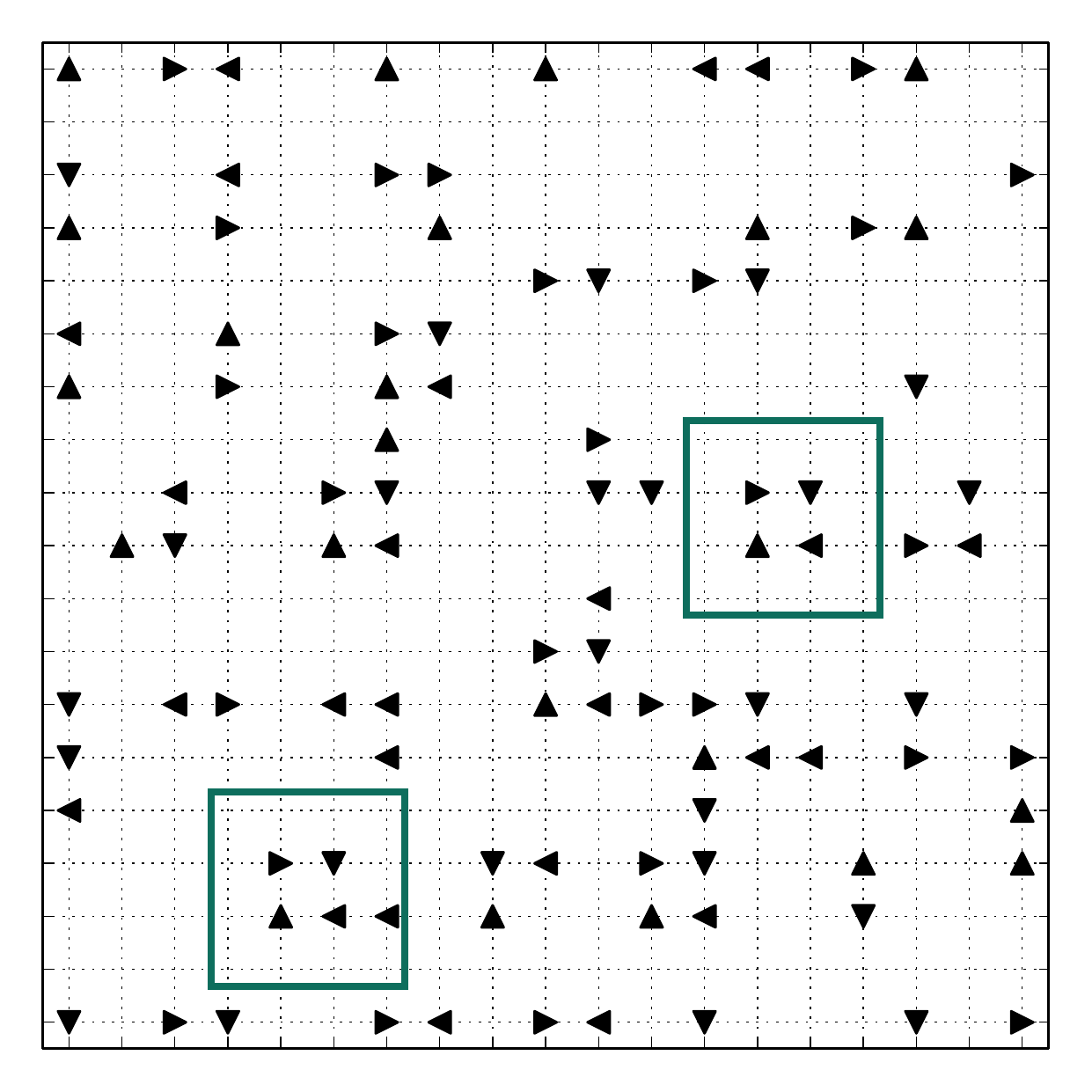}\label{fig:SAswirl}} 
    \caption{Examples of emergent patterns in the self-assembly scenario: (a)~lines, (b)~pairs, (c)~squares, (d)~triangular lattice, (e)~random dispersion, (f)~aggregation, (g)~clustering, (h)~loose grouping, and~(i)~swirls. Agents and their headings are represented by \textit{black triangles}. \textit{Green boxes} represent arbitrary examples of the respective pattern to guide the eye~\cite{kaiser2021a}.}
    \label{fig:SApatterns}
\end{figure*}

Each agent is equipped with an actor-predictor ANN pair (see Fig.~\ref{fig:ANNs}). 
Here, the actor determines the next action of the agent by selecting two action values: $a_0(t)$ deciding whether to move or turn and $a_1(t)$ determining the turning direction. 
We use only $a_0(t)$ as input to the ANNs as in previous work~\cite{borkowski17}.
While the actor always outputs a turning direction $a_1(t)$, it is only relevant when $a_0(t)$ selects to turn. 
Thus, actor and predictor ANN have 15~inputs each: 14~binary sensor values and one action value. 
The predictor outputs predictions for the 14~sensors. 
We evolve the ANN pairs with a simple genetic algorithm using a population size of~$50$, proportionate selection, age-based replacement, elitism of one, no crossover, and run evolution for~$100$ generations. 
Each value of the genome is mutated with a probability of~$0.1$ and by adding a uniformly random number from~$[-0.4,0.4]$. 
We evaluate each genome ten~times for $500$~time steps each and take the minimum fitness~(Eq.~\ref{equ:fitness}) from these ten repetitions as the overall fitness of that genome. 
Agents are uniformly 
randomly placed on the grid at the beginning of each evaluation. 

We keep a fixed swarm size $N$ of $100$ and vary the side lengths~$L\in[11,30]$ (here, number of grid cells) of our squared grid world to vary the swarm density (i.e., agents per area: $\frac{N}{L \times L}$) and to study the effects on the emerging swarm behaviors. 
The resulting densities are between $11.1~\%$~($\frac{100}{30\times 30}$) and $82.6~\%$~($\frac{100}{11\times 11}$).
A~grid size of $10\times 10$ would lead to a swarm density of $100\%$ and obviously only allows for aggregation. 
In that case, the actor's outputs are negligible and evolution would lead to a predictor constantly outputting `1' for all sensor predictions. 
Table~\ref{tab:SAparameters} summarizes all parameters.

\subsubsection{Metrics for Classification of Emergent Behaviors}
\label{section:SAmetrics} 

We define metrics for nine different patterns to analyze the emergent self-assembly behaviors (i.e., phenotypes) quantitatively using an empirical approach. 
The definition of these patterns is based on an initial qualitative analysis of the results. 
Thus, we cannot guarantee that the set is complete, but we are still confident that all distinguishable, relevant patterns are covered by our metrics. 
We define lines, pairs, aggregation, clustering, loose grouping, random dispersion, squares, swirls, and triangular lattices (see Fig.~\ref{fig:SApatterns}). 
Compared to our previous studies~\cite{kaiser19c, kaiser20b}, we add swirls as a new pattern. 
All nine behaviors are either rotation symmetric (e.g., squares) or exploit that agents are stopped when the grid cell in front is occupied in order to form repetitive patterns (e.g., lines). 
In the case of rotation symmetric patterns, agents stay on their current grid cell by turning constantly. 
Thus, the swarm creates a `boring', structured environment that allows for simple sensor predictions and high prediction accuracy, as all agents in the pattern have similar sensor readings. 
We classify resulting behaviors based on the highest resemblance of the emergent pattern formed by the agents in the last time step of the run to one of the nine defined patterns 
and solution quality gives the percentage of agents assembled into this structure. 

Lines~(Fig.~\ref{fig:SAline}) and pairs (Fig.~\ref{fig:SApairs}) consist of agents with parallel headings. 
To ensure a static structure, the agents at the ends of the structure have to point inwards if the length of the line is less than the grid side length~$L$. 
This guarantees that all intended forward movement is blocked. 
On each side next to the structure, we allow a maximum of half the structure's length of neighbors whereby these are not allowed to be positioned on two adjacent grid cells. 
The two patterns are differentiated by their lengths: pairs consist of exactly two agents while lines are formed by at least three agents. 

We define three dispersion behaviors: squares (Fig.~\ref{fig:SAsquares}), triangular lattices (Fig.~\ref{fig:SAtriangular}), and random dispersion (Fig.~\ref{fig:SArandomdisp}). 
The patterns are rotation symmetric rendering the agent headings irrelevant here. 
Agents in triangular lattices are positioned in a 2D diagonal square lattice. 
In squares, agents are one grid cell apart in each direction. 
Randomly dispersed agents have either no neighbors in their von~Neumann neighborhood or maximally one neighbor in their Moore~neighborhood.

We differentiate four grouping behaviors: aggregation (Fig.~\ref{fig:SAaggregation}), clustering (Fig.~\ref{fig:SAclustering}), loose grouping (Fig.~\ref{fig:SAloosegroup}), and swirls (Fig.~\ref{fig:SAswirl}). 
To be part of a cluster, an agent needs at least six neighbors in its Moore neighborhood whereof three have to be in its von~Neumann neighborhood.
Neighbors are also considered part of the cluster independent from fulfilling these criteria themselves. 
We classify the formation of one single cluster as aggregation, several separate clusters as clustering, and interconnected clusters as loose grouping.
Swirls consist of four agents positioned in a square with no free cells in between. 
Each agent's heading points to one of its neighbors whereby the neighbor's heading differs by $\pm 90^\circ$.

\subsubsection{Statistical Tests} \label{section:SAStatisticalTests}

We test for statistically significant differences between the different scenarios using non-parametric tests (i.e., no normality assumption) as recommended for evolutionary robotics~\cite{doncieux2011, garcia2008}. 
Fitness and solution quality are tested group-wise using the Kruskal-Wallis test (hereafter abbreviated KW)~\cite{kruskal1952use} and the Mann-Whitney U test (MW-U)~\cite{mann1947test} with Bonferroni correction (BC)~\cite{bonferroni} for post-hoc tests. 
We compare emergent behavior distributions (i.e., how often did we observe clustering, dispersion, etc.) both pairwise and between several groups with Fisher's Exact test (FE)~\cite{fisher1992statistical}. 
For group-wise comparisons, we use FE with BC as a post-hoc test.
All test results are online~\cite{kaiser2021a}.

\subsection{Emergent Behaviors} \label{section:SAbehaviors}

In a first step, we investigate the emergent behaviors when applying our minimize surprise approach to the self-assembly scenario (see Sec.~\ref{section:SAscenario}). 
We analyze the fitness (Sec.~\ref{section:SApredictionaccuracy}), the structures formed by the resulting behaviors (Sec.~\ref{section:SABehaviorClassification}), the relation between the formed patterns and the sensor value predictions (Sec.~\ref{section:SApredictions}), and the effectiveness of our minimize surprise approach in comparison to randomly generated controllers (Sec.~\ref{section:SAeffectivity}).

\subsubsection{Prediction Accuracy} \label{section:SApredictionaccuracy}

\begin{figure}[t]
    \centering
    \subfloat[]{\includegraphics[width=0.9\linewidth]{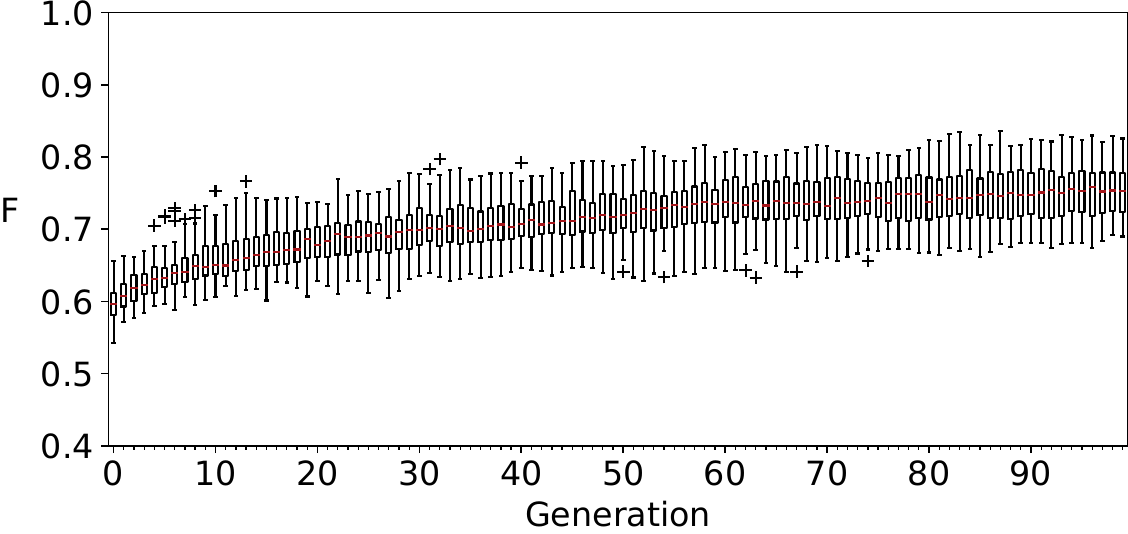}\label{fig:SAfit12}}\\
    \subfloat[]{\includegraphics[width=0.9\linewidth]{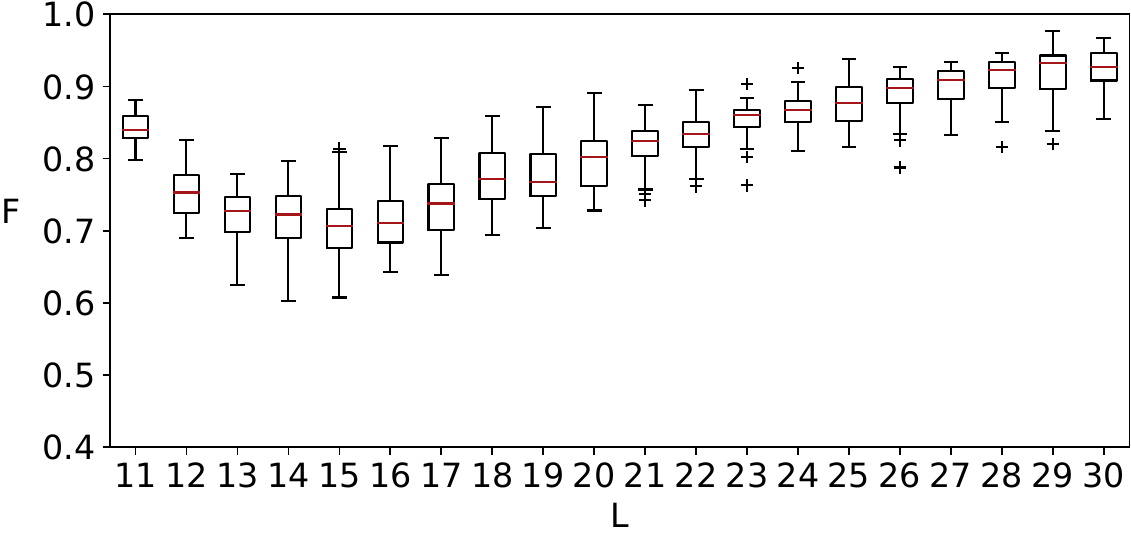}\label{fig:SAfitness_evo}}\\
    \subfloat[]{\includegraphics[width=0.9\linewidth]{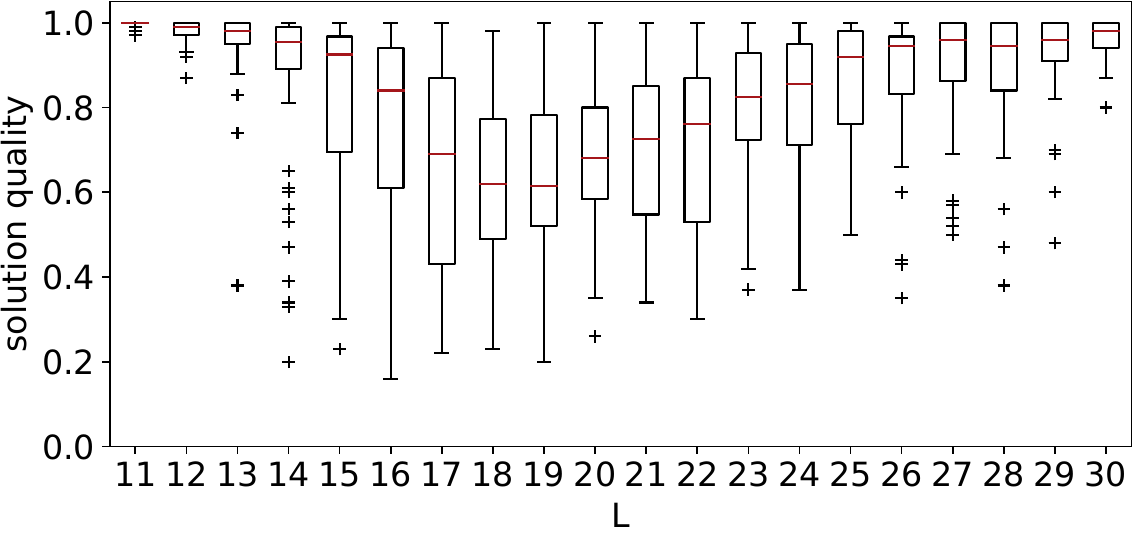}\label{fig:SAquality}}
    \caption{Best fitness F of 50~independent minimize surprise runs (a)~over generations on the $12\times 12$ grid ($L=12$) and (b)~for the last generation per grid size $L$ as well as (c)~solution quality (i.e., percentage of agents positioned in dominant structure) of best individual per grid size $L$. Medians are indicated by the \textit{red bars}~\cite{kaiser2021a}.}
    \label{fig:SAfit}
\end{figure}

Fitness (Eq.~\ref{equ:fitness}), that is, prediction accuracy, is a measure of success for our minimize surprise approach. 
Fig.~\ref{fig:SAfit12} shows exemplarily for all grid sizes the increase of best fitness of 50~independent evolutionary runs over generations for the $12\times 12$ grid. 
In Fig.~\ref{fig:SAfitness_evo}, we show the best fitness for grid sizes $L \in [11, 30]$, that is, different swarm densities. 
The median best fitness in the last generation ranges from $0.71$~($L=15$) to $0.93$ ($L=29$). 
That means at least $71\%$ of the sensor values were predicted correctly by the evolved predictors. 
There are statistically significant differences in the best fitness of the various grid sizes (KW, $p<0.001$). 
The prediction task is easiest for dense ($L\in [11,12]$) and sparse ($L\in [18,30]$) swarm densities, that is, we find high fitness values. 
For intermediate densities ($L\in [13, 17]$), we find statistically significantly lower fitness values (MW-U with BC, $p < 0.05$) and thus the prediction task is harder here. 

\subsubsection{Behavior Classification} \label{section:SABehaviorClassification}

\begin{figure*}[t]
    \centering
    \subfloat[ ]{\includegraphics[width=0.12\textwidth]{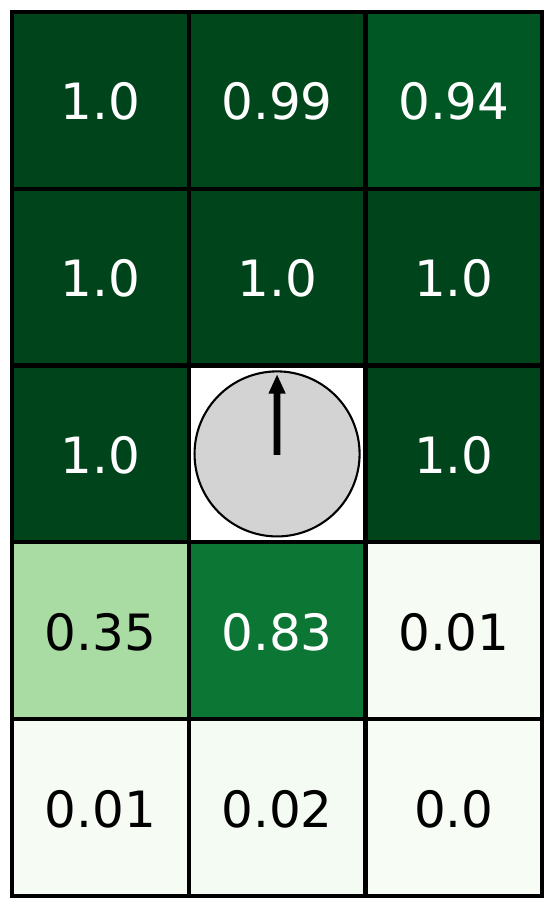}\label{fig:SAaggPred}}
    \hspace{3mm}
     \subfloat[ ]{\includegraphics[width=0.12\textwidth]{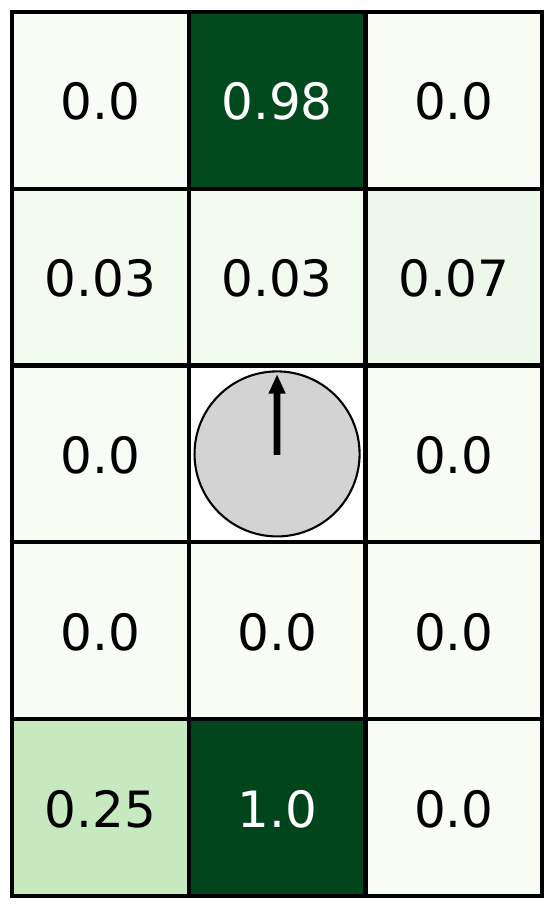}\label{fig:SAsquaresPred}} 
    \hspace{3mm} 
     \subfloat[ ]{\includegraphics[width=0.12\textwidth]{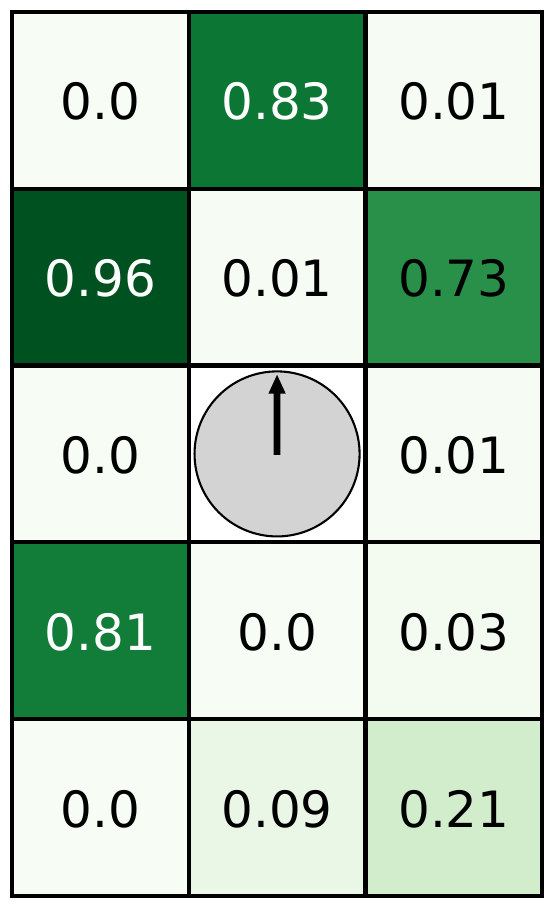}\label{fig:SAtriangularPred}}
     \hspace{3mm} 
    \subfloat[ ]{\includegraphics[width=0.12\textwidth]{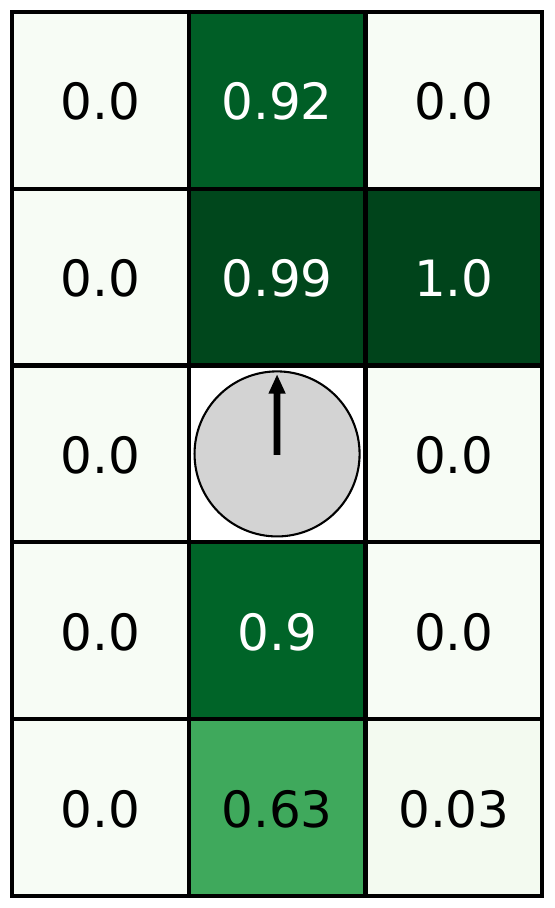}\label{fig:SAlinePred}}
    \hspace{3mm}
    \subfloat[ ]{\includegraphics[width=0.12\textwidth]{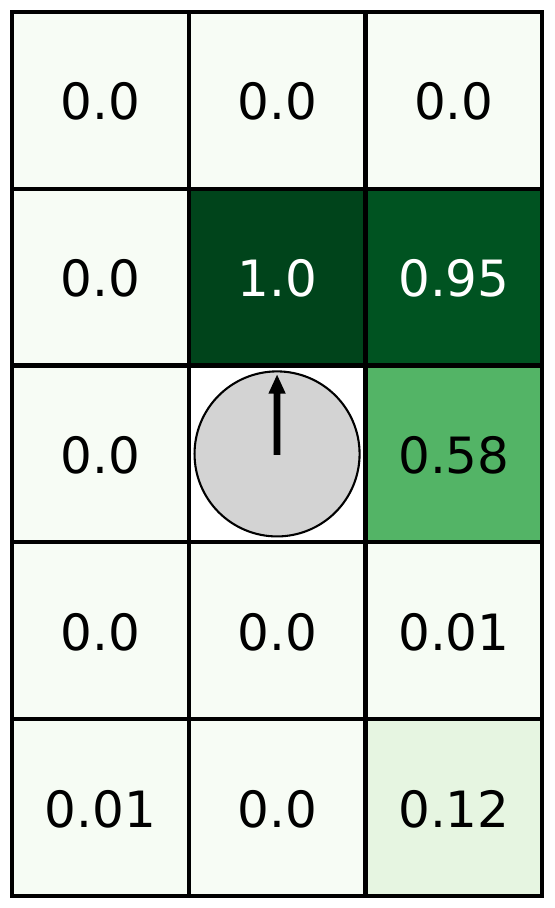}\label{fig:SAswirlPred}}
    \caption{Mean sensor predictions (1: always predict occupied, 0: always predict non-occupied) of the agents' $14$ sensor values for (a)~aggregation (behavior in Fig.~\ref{fig:SAaggregation}), (b)~squares (behavior in Fig.~\ref{fig:SAsquares}), (c)~triangular lattice (behavior in Fig.~\ref{fig:SAtriangular}), (d)~lines (behavior in Fig.~\ref{fig:SAline}), and (e)~swirls (behavior in Fig.~\ref{fig:SAswirl}).
    Agents are represented by \textit{gray circles}, \textit{black arrows} indicate their headings~\cite{kaiser2021a}.}
    \label{fig:SApredictions}
\end{figure*}

We classify the patterns formed by the best evolved individuals in their evaluation's last time step~$T$ using the metrics presented in Sec.~\ref{section:SAmetrics}. 
Fig.~\ref{fig:SA_MS_structures} shows the percentage of formed structures over grid size.
A video of emergent behaviors is online~\cite{kaiser2021a}. 
Overall, the behavior distributions are statistically significantly different, that is, grid size affects the emergent behavior distribution (FE, $p<0.01$). 
This significant difference occurs with increasing difference in swarm density (FE with BC, $p < 0.05$), that is, neighboring distributions usually do not vary to a statistically significant degree. 
Grouping behaviors (i.e., aggregation, loose grouping, and clustering) prevail on small grid sizes, as these structures form easily in high swarm densities. 
Movement is barely possible on the smallest grid sizes $L\in [10,12]$ and thus only aggregation and loose grouping can emerge. 
We note a shift in the distributions towards pairs, lines, and dispersion behaviors with increasing grid size.
The decrease in swarm density allows for the formation of patterns that require agents to be distributed in space.
Across all grid sizes, triangular lattices, squares, and swirls emerge rarely most likely as they require exact positioning of several agents. 
We find the most diverse behavior distributions with seven and six out of nine possible structures on the $16\times 16$ grid and $15\times 15$ grid, respectively. 
Next, we compare the solution quality (i.e., percentage of agents forming the dominant structure) over grid size (see Fig.~\ref{fig:SAquality}) and find statistically significant differences (KW, $p<0.001$). 
As expected, solution quality is highest on the $11\times 11$ grid, as the great majority of agents ends up aggregated due to the high swarm density. 
In general, sparse and dense settings lead to higher solution quality than intermediate densities.

\subsubsection{Predictions} \label{section:SApredictions}

To reach both high solution quality and fitness, a high percentage of agents needs to be positioned in a pattern and the predictions have to match this formed structure closely. 
Therefore, we study the mean sensor predictions of the best individuals and compare them to the formed structure. 
In aggregation (Fig.~\ref{fig:SAaggPred}) and other grouping behaviors (i.e., clustering, loose grouping), most of an agent's adjacent grid cells are predicted to be occupied. 
In contrast, only few or no neighbors are predicted for dispersion behaviors like squares (Fig.~\ref{fig:SAsquaresPred}; square lattice with one empty grid cell between agents in all directions), triangular lattices (Fig.~\ref{fig:SAtriangularPred}) or random dispersion. 
In line structures (Fig.~\ref{fig:SAlinePred}), agents predict the grid cells in front and behind them to be occupied. 
Similarly, for pairs a neighbor is predicted only on the grid cell directly in front of the agent. 
Predictions of swirls (Fig.~\ref{fig:SAswirlPred}) contain three neighbors, whereby one neighbor is directly in front of the agent and the other two to their right (or left).  
Overall, we find that the formed patterns and the mean sensor predictions match closely. 
As we have 14~sensors, small deviations between mean predictions and the sensor values induced by the formed structures will not significantly affect prediction accuracy. 
For example, the mean predictions of the line structure in Fig.~\ref{fig:SAlinePred} include the prediction of a neighbor to the front right of the agent that does not match the formed pattern, which leads to a reduction of prediction accuracy by only $\frac{1}{14}$th.

\subsubsection{Effectiveness of the Approach} \label{section:SAeffectivity}

\begin{figure}[t]
    \centering
    \includegraphics[width=\linewidth]{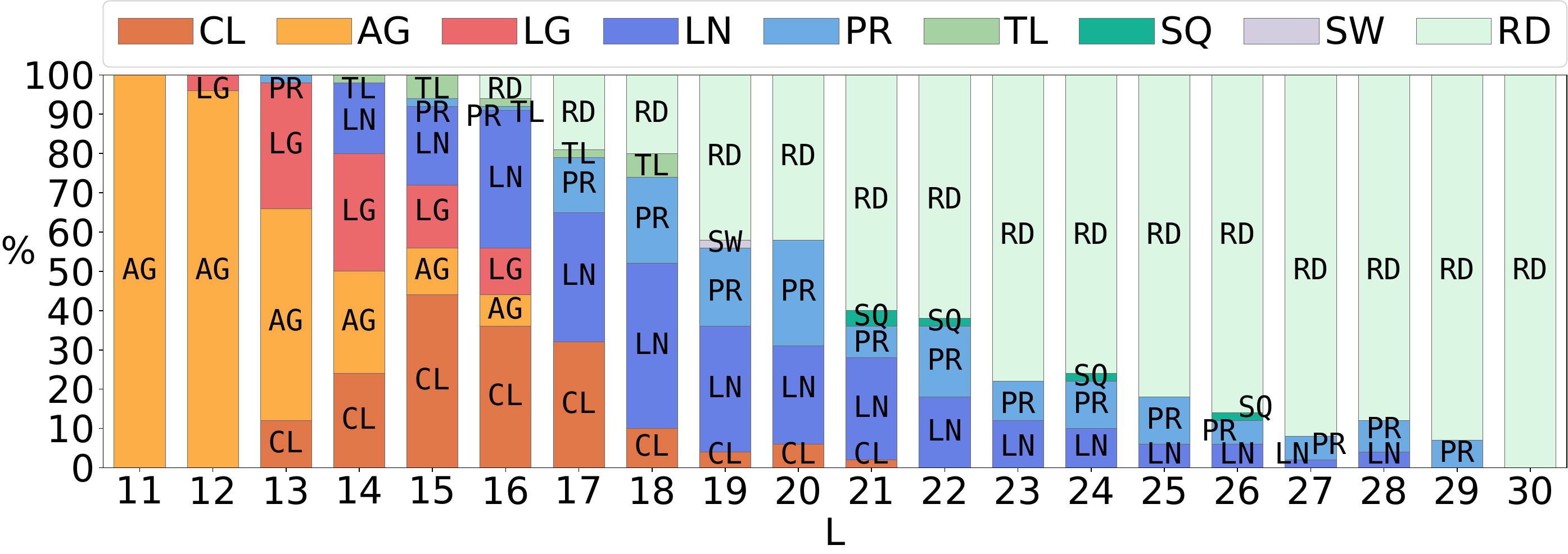}
    \caption{Percentage of resulting structures in minimize surprise for grid sizes~$L\times L$ with $L\in[11,30]$ with clustering~(CL), aggregation~(AG), loose grouping~(LG), lines~(LN), pairs~(PR), triangular lattice~(TL), squares~(SQ), swirls~(SW) and random dispersion~(RD). Adapted from~\cite{kaiser20b}.}
    \label{fig:SA_MS_structures}
\end{figure}

\begin{figure}[t]
    \centering
    \subfloat[]{\includegraphics[width=\linewidth]{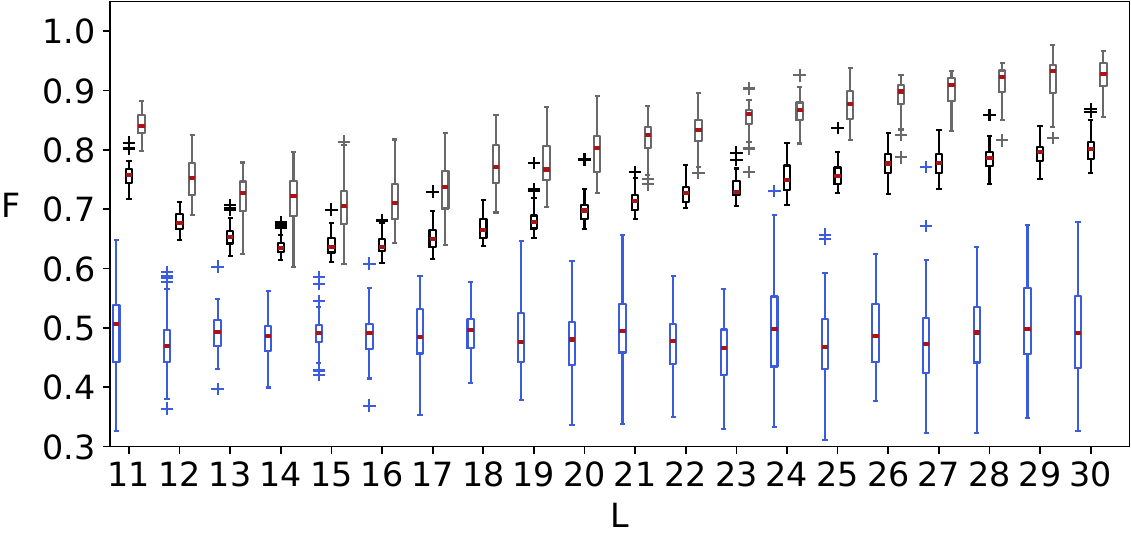}\label{fig:SAfitness_rand}}\\
    \subfloat[]{\includegraphics[width=\linewidth]{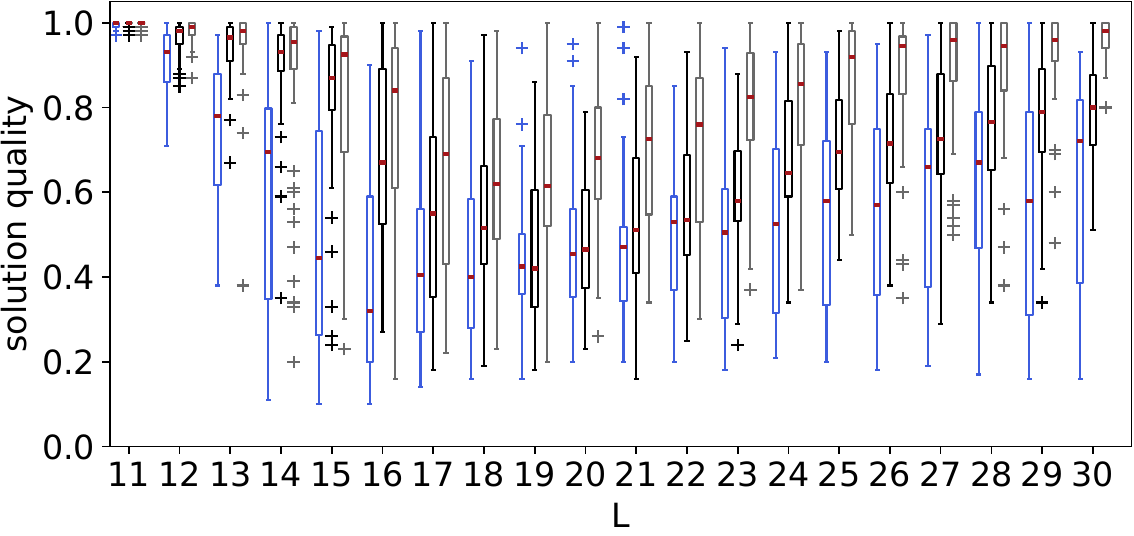}\label{fig:SAquality_rand}} \\
    \subfloat[]{\includegraphics[width=\linewidth]{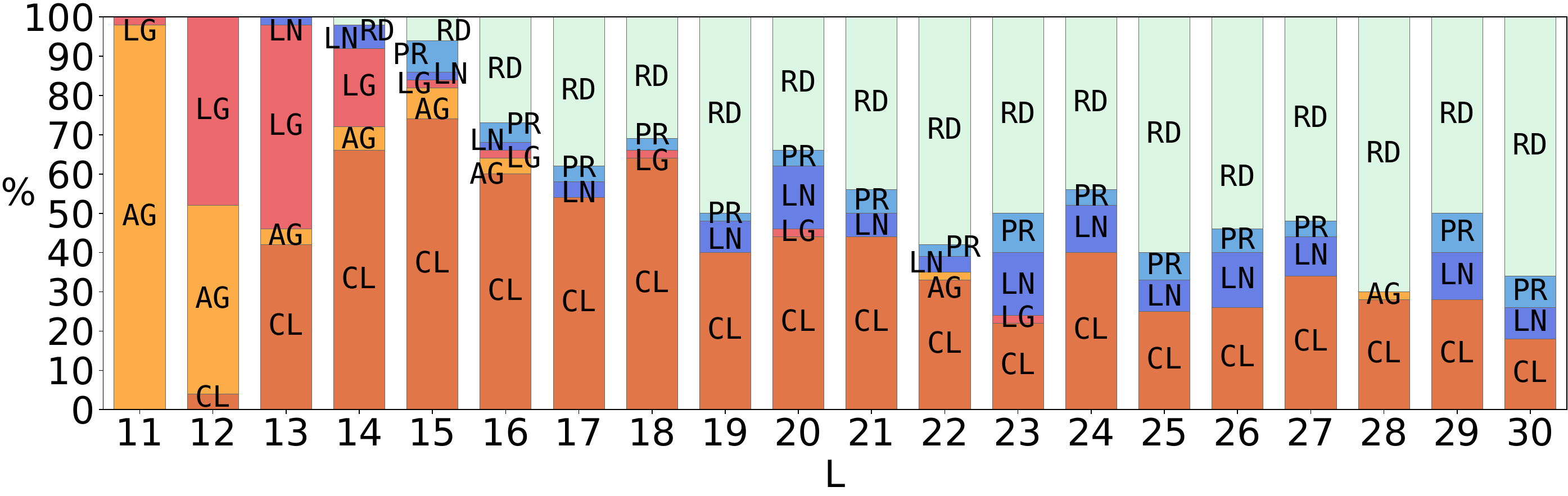}\label{fig:SAbehavrandom}}\\
    \subfloat[ ]{\includegraphics[width=\linewidth]{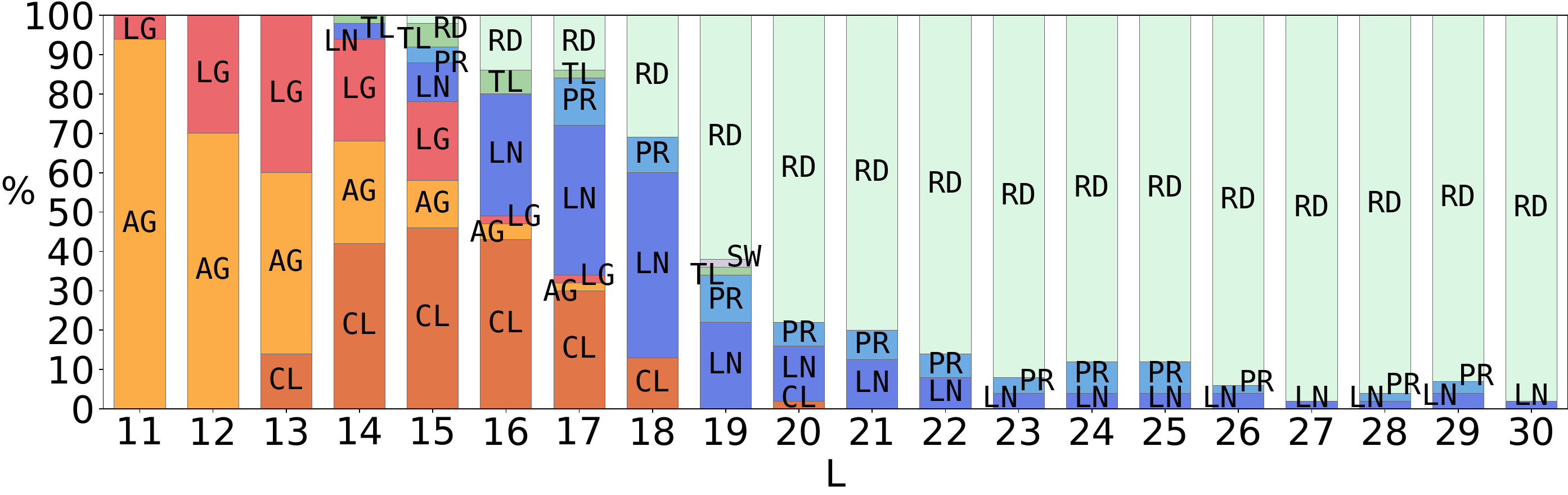}\label{fig:SAbehavrandomnoselect}}
    \caption{Best fitness~(a) and solution quality~(b) (i.e., percentage of agents positioned in dominant structure) of 50~random individuals (\textit{blue boxes}, left), 50 best individuals of one population of randomly generated controllers (selected random individuals; \textit{black boxes}, middle), and 50~best evolved individuals (\textit{gray boxes}, right), and percentage of structures formed by the random individuals~(c) and by the selected random individuals~(d) with clustering (CL), aggregation (AG), loose grouping (LG), lines (LN), pairs (PR), triangular lattice (TL), swirls (SW), and random dispersion (RD) for grid sizes $L\times L$ with $L\in[11,30]$. Medians are indicated by the \textit{red bars}~\cite{kaiser2021a}.}
    \label{fig:SAfitRand}
\end{figure}

Since minimize surprise does not work towards solutions for a given task but rather implements an exploratory search, we want to test its effectiveness in finding relevant and interesting behaviors. 
For this purpose, we compare the best evolved individuals of the evolutionary runs with minimize surprise (see Sec.~\ref{section:SAbehaviors}) to randomly generated actor-predictor pairs. 
We generate two sets of random ANN pairs per grid side length $L\in [11,30]$: (i)~by creating 50~random individuals, and (ii)~by creating 50~times a population of $5{,}000$~random individuals and selecting the best individual from each population based on prediction success (Eq.~\ref{equ:fitness}; hereafter referred to as selected random individuals).
The number of evaluations in the latter approach is equal to that in minimize surprise (i.e., 50~runs with $100$~generations $\times~50$~individuals).
We compare best fitness (Fig.~\ref{fig:SAfitness_rand}), solution quality (i.e., percentage of agents in the structure; Fig.~\ref{fig:SAquality_rand}), and behavior distributions (Figs.~\ref{fig:SA_MS_structures},~\ref{fig:SAbehavrandom}) of the best evolved and the two sets of random ANN pairs per grid size. 

For all grid sizes, the ANN pairs evolved with minimize surprise reach statistically significantly greater fitness than the random individuals and the selected random individuals. 
As expected, the selected random individuals reach also significantly greater fitness than the random individuals (MW-U with BC, $p<0.01$; Fig.~\ref{fig:SAfitness_rand}).
We find that the best evolved individuals of our minimize surprise runs have significantly better solution quality than the random individuals for all grid sizes except for $L=11$ (MW-U with BC, $p<0.05$; Fig.~\ref{fig:SAquality_rand}).
The selected random individuals have significantly better solution quality than the random individuals for all grid sizes except for $L \in \{11, 17, 18, 19, 20, 21, 22, 28\}$.
The best evolved individuals reach significantly better solution quality than the selected random individuals for $L \ge 19$, while we do not find significant differences on smaller grid sizes.
The behavior distributions of the random individuals and the selected random individuals differ significantly for all grid sizes except for $L \in \{11,12,14\}$ (FE with BC, $p<0.05$; Figs.~\ref{fig:SAbehavrandom},~\ref{fig:SAbehavrandomnoselect}).
The random individuals and the best evolved individuals of our minimize surprise approach lead to significantly different behavior distributions for all grid sizes except for $L=11$. 
For the selected random individuals and the best evolved individuals, we find significant differences for $L \in \{12,20\}$ while the behavior distributions on all other grid sizes are not significantly different.

Next, we investigate if some of the behaviors are found more or less likely in evolution than in randomly generated individuals.
We study if evolution's selection and improvement over generations influences the resulting behaviors in general. 
For each behavior we compare its frequency over all grid sizes $L\in [11,30]$ between the  best evolved individuals, the random individuals, and the selected random individuals (FE with BC, $p<0.05$).     
The likelihood for clusters is highest for random individuals and similarly likely for the selected random individuals and the best evolved individuals.  
By contrast, aggregation is more likely for the best evolved individuals and the selected random individuals than for the random individuals. 
Loose grouping and swirls are found equally likely in all three scenarios. 
Lines, pairs, and squares are found most likely in evolution.
These behaviors require correct positioning to reach high fitness (see Sec.~\ref{section:SAmetrics}) and thus seem to rely on selection and improvement over generations of the evolutionary process. 
Triangular lattices are more likely for the selected random individuals and the best evolved individuals than for the random individuals.
Random dispersion is most likely for selected random individuals and least likely for random individuals. 

Overall, this indicates that evolution successfully enables the adaptation of actors and predictors.
Actor outputs lead to behaviors that are correctly predicted by the predictors and predictors are adapted to behaviors to optimize prediction accuracy. 
The resulting behaviors seem consistent in themselves and generate reasonable patterns that are recognized as such by a human observer. 
Complex behaviors (e.g., lines, squares) that are rarely generated in random runs are selected and improved
over generations to reach high fitness and solution quality. 
For low swarm densities, random runs lead to greater behavioral diversity but significantly lower solution quality than evolution. 
Only a part of the swarm assembles into the dominant structure while the rest forms random structures.
Thus, we conclude that minimize surprise is an effective approach that outperforms random search. 
Our approach can be used to evolve a variety of reasonable and potentially useful 
behaviors that have high solution quality and potential applicability in real-world scenarios.

\subsection{Behavioral Diversity in Comparison to Novelty Search} \label{section:SANovelty} 

A variety of behaviors emerges across independent evolutionary runs using minimize surprise (see Sec.~\ref{section:SAbehaviors}).
By contrast, divergent search algorithms (see Sec.~\ref{section:divergentsearch}) generate behavioral diversity within one evolutionary run.
Here, we compare these different approaches to the generation of behavioral diversity by showing that minimize surprise is competitive to novelty search~\cite{kaiser20b, kaiser20a}.\footnote{An extensive comparison of minimize surprise, minimize surprise with predefined predictions (cf. Sec.~\ref{section:SAengineeredSO}), novelty search with a task-specific behavioral characteristic, novelty search with a task-independent characteristic, and a standard genetic algorithm can be found in~\cite{kaiser20b}.}
We compare to novelty search as it is representative of divergent search algorithms due to its competitiveness~\cite{lehman2010, lehman2011} and easy implementation. 
In this study, we base both minimize surprise and novelty search on the same genetic algorithm to avoid influences due to different evolutionary algorithms.

\subsubsection{Method}

Our implementation of novelty search is based on the genetic algorithm presented in Sec.~\ref{section:SAscenario} with the exclusion of elitism. 
In novelty search, individuals (here only the actor ANN as in Fig.~\ref{fig:actor}) are evaluated based on their novelty~$\rho$ by comparing the current population to an archive of past individuals (i.e., previously visited regions of behavior space).
Novelty $\rho$ is given by
\begin{equation}
    \rho(i) = \frac{1}{K} \sum^{K-1}_{k=0} \mathrm{b\_dist}(i, \mu_k)\;,
    \label{eq:novelty}
\end{equation}

where $\mu_k$ is individual $i$'s $k$th-nearest neighbor with respect to the behavioral distance metric~$\mathrm{b\_dist}(\cdot,\cdot)$ and considering \textit{K}~nearest neighbors. 
Our behavioral distance~$\mathrm{b\_dist}(\cdot,\cdot)$ is the Euclidean distance between the behavioral feature vector of individual $i$ and its $k$th-nearest neighbor and we use up to the tenth nearest neighbor ($K=10$). 
Individuals are added with a low probability of~$2\%$ to the archive of past individuals~\cite{lehman2010}. 
We define a task-independent behavioral feature vector (i.e., behavioral characteristic) as an R-dimensional vector 
\begin{equation}
    \mathbf{b} = \frac{1}{N}\left[\sum^{N-1}_{n=0} s_0^n(T),\,\dots\,,\sum^{N-1}_{n=0} s_{R-1}^n(T)\right]\; 
\end{equation}

of $R$~sensor values in the last time step of the evaluation averaged over swarm size~$N$.
These vectors of mean sensor values can serve as local pattern templates.
Thus, a variety of sensor patterns may emerge as novelty search tries to explore all variants. 
We average the behavioral vectors obtained in ten repetitions per individual. 
For the comparison of novelty search with minimize surprise, we restrict ourselves to grid sizes $15\times 15$ and $20\times 20$ as sample scenarios for denser and sparser swarm density settings to keep this study reasonably short. 
As before, emergent behaviors are classified with the metrics presented in Sec.~\ref{section:SAmetrics}. 
The structure classification is a different, discretized representation of the behavioral feature vectors that allows comparison with the minimize surprise runs.
For minimize surprise, we keep the best evolved individual each of the 50~evolutionary runs from Sec.~\ref{section:SAbehaviors} as solutions.
For novelty search, we do 50~independent runs per grid size
and consider two sets of potential solutions: (i)~\textbf{all} 250,000 individuals (50~individuals, 100~generations, 50~experiments; N-SV\textbf{A}), and (ii)~the 50~individuals with the best solution \textbf{quality} out of all 250,000 individuals (N-SV\textbf{Q}).

\subsubsection{Results}

\begin{figure}[!t]
    \centering
    \subfloat[]{\includegraphics[width=0.7\linewidth]{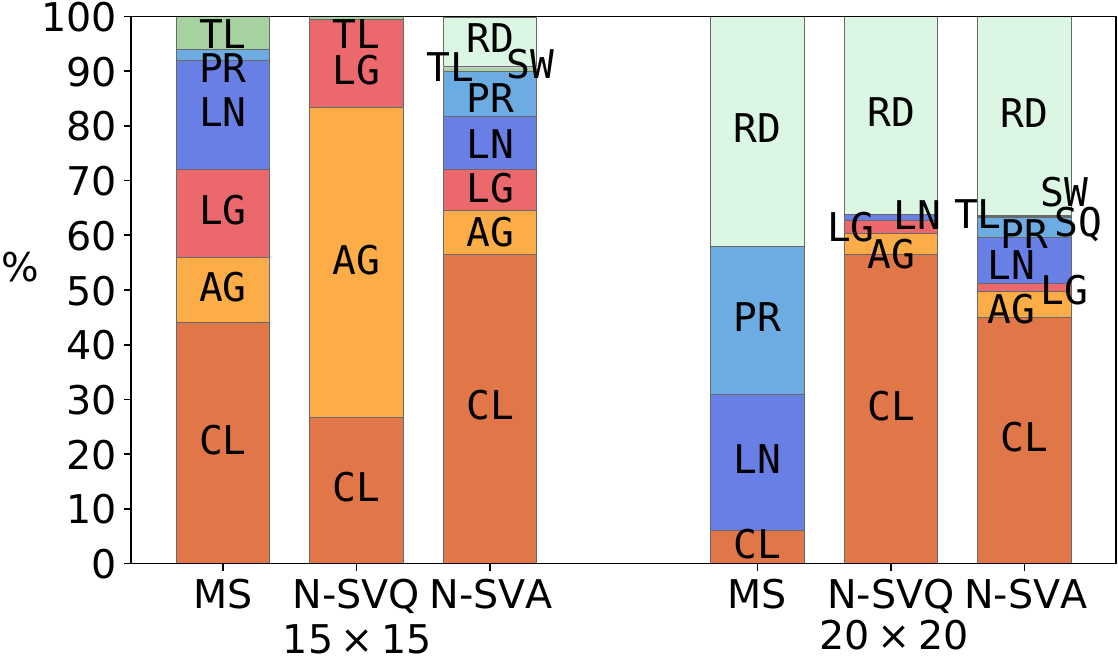}\label{fig:SAbehdist}}\\
    \subfloat[]{\includegraphics[width=0.7\linewidth]{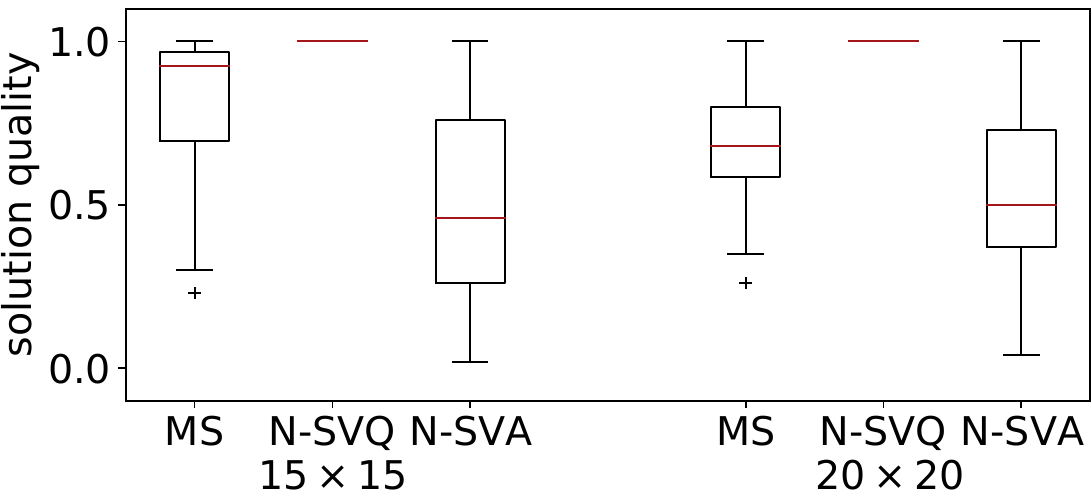}\label{fig:SAcompquali}}
    \caption{Comparison of minimize surprise~(MS) and novelty search with behavioral characteristic~$b$ considering the 50~individuals with the best solution quality (N-SVQ) and all 250,000 individuals (N-SVA) regarding (a)~percentage of resulting structures with clustering~(CL), aggregation~(AG), loose grouping~(LG), lines~(LN), pairs~(PR), triangular lattice~(TL), squares~(SQ), swirls~(SW), and random dispersion~(RD) and (b)~solution quality (i.e. percentage of agents positioned in dominant structure) for grid sizes $15\times 15$ and $20\times 20$~\cite{kaiser2021a}.}
    \label{fig:SAcomp}
\end{figure}

We find statistically significant differences in the behavior distributions (Fig.~\ref{fig:SAbehdist}) of minimize surprise and both solution sets of novelty search per grid size (FE with BC, $p<0.001$). 
Minimize surprise leads to fewer different structures than 
N-SVA on both grid sizes. 
However, additional behaviors are probably generated by chance as we consider 5,000~times more individuals as potential solutions for N-SVA than for minimize surprise. 
Compared to N-SVQ, minimize surprise leads to more different structures on the $15\times 15$ grid and to less structures on the $20\times 20$~grid.
As for minimize surprise (see Sec.~\ref{section:SAbehaviors}), we find more grouping behaviors on the smaller grid and more dispersion behaviors, lines,  and pairs on the larger grid for N-SVA. 
For N-SVQ, grouping behaviors prevail on both grid sizes.
Solution quality, that is, the percentage of agents in the dominant structure (see Sec.~\ref{section:SAmetrics}), indicates how consistently a structure was formed by the swarm. 
Minimize surprise leads to statistically significantly higher solution quality than N-SVA on both grid sizes (MW-U with BC, $p<0.001$). 
A~possible explanation is how solutions are generated in each of the two methods.
Novelty search creates a large number of diverse solutions within one run without putting weight on improving those over generations by only aiming for novel behaviors.
Minimize surprise, in contrast, produces behavioral diversity across independent evolutionary runs while aiming to improve of prediction accuracy (Eq.~\ref{equ:fitness}) of its individuals over generations in a single run. 
Hence, novelty search requires post-evaluation to select the best performing individuals, while minimize surprise already provides optimized individuals as solutions. 
When selecting only the best 50~individuals based on solution quality as solutions of novelty search (N-SVQ), we find higher solution quality than in minimize surprise. 
But in this case, grouping behaviors prevail and structures that require correct positioning, such as triangular lattices, are rare.

Overall, we find that novelty search and minimize surprise each have their unique advantages. 
Novelty search leads to more behavioral variety when considering all individuals as potential solutions (N-SVA).
But novelty search requires post-evaluation to select high-quality solutions, which in turn may reduce behavioral variety as in our case of selecting the best individuals based on solution quality (N-SVQ).
An alternative to the two presented solution sets for novelty search (i.e., N-SVA and N-SVQ) is to select the highest quality individual per pattern type as solutions. 
This gives us a maximally diverse and high quality solution set that outperforms minimize surprise. 
Nevertheless, the post-evaluation required in novelty search is a drawback that would probably be resolved when using a QD method, such as MAP-Elites.
Minimize surprise directly provides high quality solutions but generates less variety.

\subsection{Robustness of Behaviors} \label{section:SArobustness}

Next, we study the robustness of self-assembly behaviors evolved with minimize surprise. 
We introduce sensor noise as a non-deterministic element into our otherwise completely deterministic setup first (Sec.~\ref{section:SAsensornoise}) and asses the robustness of the behaviors against damage in the formed structure afterwards (Sec.~\ref{section:SAdamage}).

\subsubsection{Sensor Noise} \label{section:SAsensornoise}

\begin{table}[!t]
\caption{Impact of sensor noise on a (a)~$15 \times 15$ grid and (b)~$20\times 20$ grid;  
median best fitness (Eq.~\ref{equ:fitness}) of the last generation and resulting structures. Reprinted from \cite{kaiser19c}, with permission from Elsevier.\label{tab:SAnoise}}
\centering
\subfloat[\label{tab:SAnoise15}]{
\begin{tabular}{ccccccc}
\hline
 \begin{tabular}{@{}c@{}}
                   \textbf{sensor} \\ 
                   \textbf{noise} 
                 \end{tabular}
 & \textbf{fitness} & \textbf{lines} & \begin{tabular}{@{}c@{}}
                   \textbf{aggre-} \\ 
                   \textbf{gation} 
                 \end{tabular} &  \begin{tabular}{@{}c@{}}
                   \textbf{cluster-} \\ 
                   \textbf{ing} 
                 \end{tabular} &
 \begin{tabular}{@{}c@{}}
                   \textbf{loose} \\ 
                   \textbf{group.} 
                 \end{tabular} & \begin{tabular}{@{}c@{}}
                   \textbf{triangular} \\ 
                   \textbf{lattice} 
                 \end{tabular} \\ \hline 
 $0\%$ & 0.698 & $15\%$ & $10\%$ & $55\%$ & $15\%$ & $5\%$ \\
 $5\%$ & 0.724 & $10\%$ & $55\%$ & $30\%$ & $5\%$ & $0\%$ \\ 
 $10\%$ & 0.694 & $5\%$ & $40\%$ & $30\%$ & $15\%$ & $10\%$  \\ 
 $15\%$ & 0.654 & $0\%$ & $50\%$ & $30\%$ & $15\%$ & $5\%$  \\ \hline 
\end{tabular}
}\\
\subfloat[\label{tab:SAnoise20}]{
\begin{tabular}{cccccc}
\hline
 \begin{tabular}{@{}c@{}}
                   \textbf{sensor} \\ 
                   \textbf{noise} 
                 \end{tabular} & \textbf{fitness} & \textbf{pairs} & \textbf{lines} & \textbf{clustering} &
 \begin{tabular}{@{}c@{}}
                   \textbf{random} \\ 
                   \textbf{dispersion} 
                 \end{tabular}  \\ \hline 
 $0\%$ & 0.805 & $27.5\%$ & $22.5\%$ & $10\%$ & $40\%$   \\
 $5\%$ & 0.768 & $0\%$ & $15\%$ & $5\%$ & $80\%$  \\ 
 $10\%$ & 0.725 & $0\%$ & $0\%$ & $15\%$ & $85\%$  \\ 
 $15\%$ & 0.699 & $0\%$ & $0\%$ & $0\%$ & $100\%$ \\ \hline 
\end{tabular}
}
\end{table}

Our simulation environment in the previous 
Sec.~\ref{section:SAscenario} is fully deterministic.
Here, we introduce sensor noise as a first step towards more realistic environments and study its influence on the resulting behaviors~\cite{kaiser19c}. 
We compare the results of our previous, fully deterministic setup (i.e., sensor noise of $0\%$) with the results of non-deterministic setups where we flip the binary sensor values of agents with a probability of $5\%$, $10\%$ or~$15\%$. 
Again, we restrict ourselves to grid sizes $15\times 15$ and $20\times 20$ as examples for dense and sparse swarm density settings and do 20~independent evolutionary runs each. 
Tables~\ref{tab:SAnoise15} and~\ref{tab:SAnoise20} provide median best fitness and behavior distribution per sensor noise for the two grid sizes. 

We find statistically significantly different fitness for the different sensor noise levels on both grid sizes (KW, $p<0.01$).
On the $15\times 15$~grid, we find significantly lower fitness for $15\%$ sensor noise (MW-U with BC, $p<0.05$). 
On the $20\times 20$~grid, fitness decreases significantly with increasing sensor noise (MW-U with BC, $p<0.05$).
The behavioral diversity decreases with increasing sensor noise, as 
more robust structures tend to dominate 
in non-deterministic environments. 
Robust structures, such as grouping behaviors and random dispersion, do not rely on the exact positioning of agents and thus cannot be destroyed by the behavior of an individual agent.
In grouping behaviors, most agents are surrounded by other agents and cannot leave the structure quickly in any event, including false sensor readings. 
In randomly dispersed structures, agents rarely sense neighbors and a few false sensor readings only marginally affect the overall fitness. 
A forward move or turn caused by a false sensor reading does not immediately lead to the formation of a different structure or a big change in the sensor values in most cases. 
Triangular lattices seem to emerge rather independently of sensor noise~(see Table~\ref{tab:SAnoise15}). 
The formed structure is rotation symmetric as in all dispersion behaviors (i.e., triangular lattices, random dispersion, squares) and agents turn constantly to stay on their grid cell. 
Sensor values might already change slightly with the change of heading when turning and thus the behaviors are possibly evolved to withstand small variations in sensor readings caused by sensor noise. 
Lines, pairs, and swirls rely on the correct positioning and orientation of agents and are less robust to disturbances. 
Here, agents may turn based on false negative sensor readings, which immediately breaks the structure. 
We find that grouping behaviors are favored on the $15\times 15$ grid and random dispersion on the $20\times 20$ grid with increasing sensor noise.
In summary, our approach is not only robust to noise but even adapts to noise by selecting for swarm behaviors that are robust to sensor errors.

\subsubsection{Damage and Repair} \label{section:SAdamage}

\begin{figure}[!t]
    \centering
    \subfloat[ ]{\includegraphics[width=0.4\linewidth]{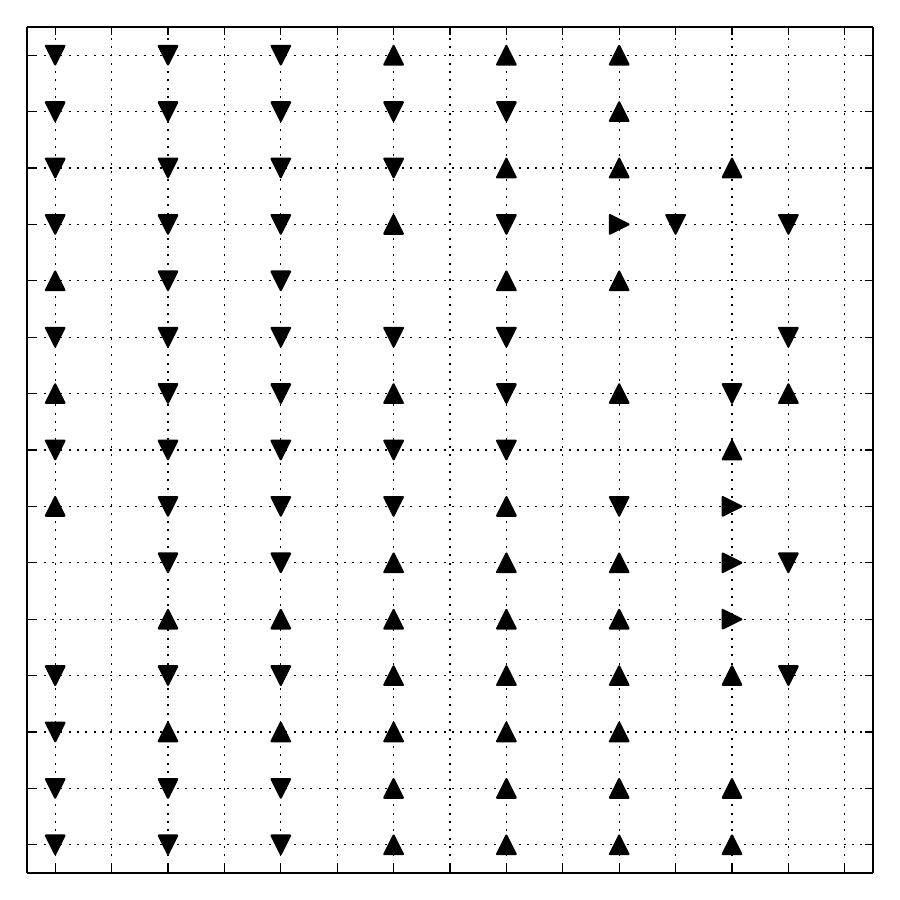}\label{fig:SAdamageLine}} 
    \hspace{3mm}
    \subfloat[ ]{\includegraphics[width=0.4\linewidth]{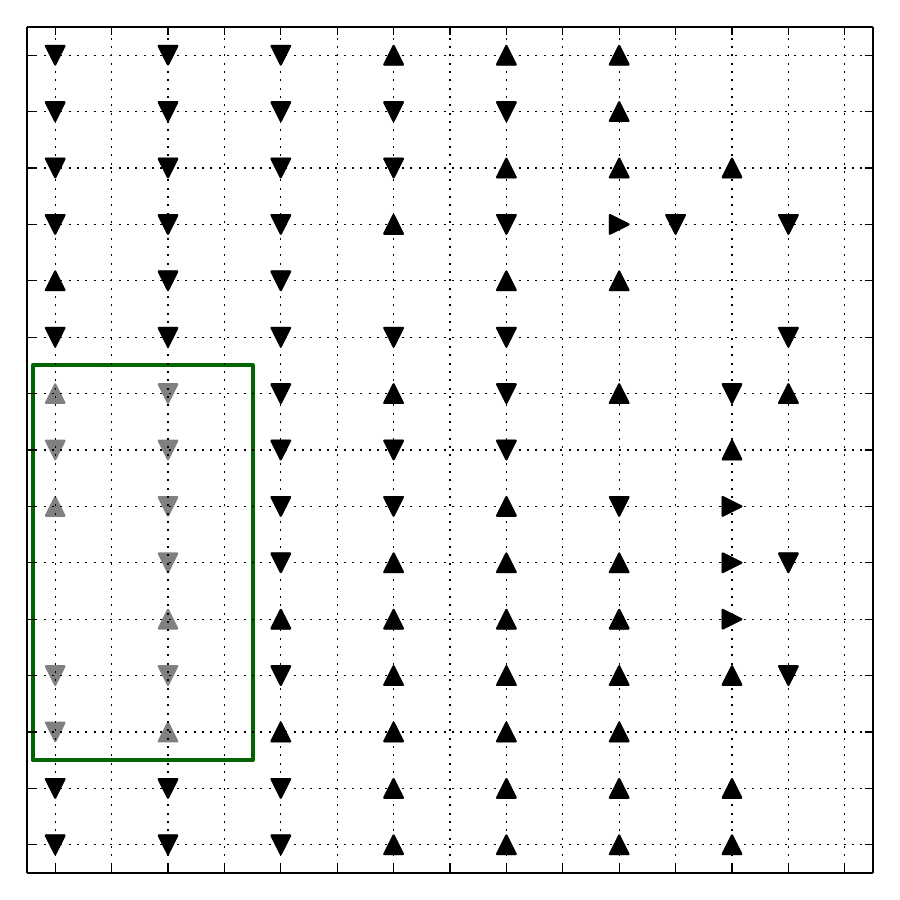}\label{fig:SAremoveArea}}
    \hspace{3mm}
    \subfloat[]{\includegraphics[width=0.4\linewidth]{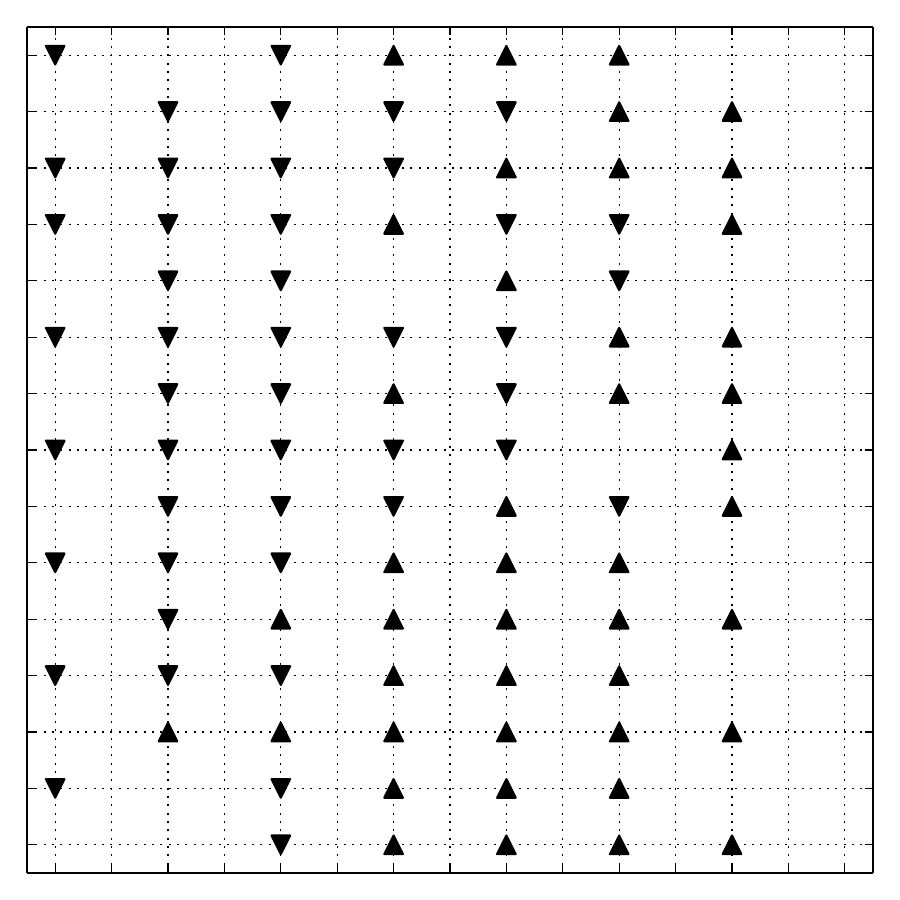}\label{fig:SArunRemoveArea}} 
    \caption{Damage and repair of a line structure: (a)~structure formed in the initial run, (b)~the structure with the removed area (12 removed agents), and (c)~the repaired structure after running the simulation for another 500~time steps. Reprinted (a)~by permission from Springer Nature:~\cite{kaiser18} \copyright Springer Nature Switzerland AG 2019 and (c)~from \cite{kaiser19c}, with permission from Elsevier.}
    \label{fig:SAdamage}
\end{figure}

Here, we study the robustness of the emergent self-assembly behaviors against damage. 
As a representative example we use the line structures shown in Fig.~\ref{fig:SAdamageLine} that form on the $15\times 15$ grid. 
A~more detailed study including additional damage scenarios was previously published~\cite{kaiser19c}.

We damage the structure by removing agents from the initially complete structure or by repositioning agents within that structure. 
Afterwards, the respective actor-predictor pair is evaluated for another 500~time steps (hereafter referred to as repair run). 
We measure fitness (Eq.~\ref{equ:fitness}) and solution quality (i.e., percentage of agents that form the line patterns) at the start and end of runs. 
Additionally, we introduce similarity~$S$ that measures the amount of agents with equal poses (i.e., agents with the same position (x,y) and orientation) in the last time step~$T$ of the initial and the repair run.
We define similarity~$S$ to the initial structure as 
\begin{equation}
    S(C_\text{repair}, C_\text{initial}) = \frac{1}{N} \sum_{u_i\in C_\text{repair}} m(u_i)\, ,  \label{equ:similarity}
\end{equation}

with lists of final agent poses $C_\textrm{repair}$ for the repair run and $C_\textrm{initial}$ for the initial run, initial swarm size~$N$, and matching between agent poses of before ($q_j \in C_\text{initial}$) and after ($u_i \in C_\text{repair}$) the damage 
\begin{equation}
    m(u_i) = \left\{
\begin{array}{ll}
1, & 
\mathrm{if}~\exists q_j\in C_\text{initial}: u_i=q_j  \\
0, & \, \text{else} \\
\end{array}
\right. \;.
\end{equation}

A similarity of~$1.0$ indicates that the initial structure was completely reassembled after damage. Consequently, the maximum possible similarity is less than~$1.0$ when reducing swarm size as only parts of the initial structure can reform.

\begin{table}[!t]
\caption{Robustness against damage of line structures: mean fitness, mean solution quality at the start and the end of the run, and mean similarity (Eq.~\ref{equ:similarity}). Median values in brackets~\cite{kaiser19c}. \label{tab:SAdamage}}
\centering
\begin{tabular}{lcccc}
\hline
 \space  & \textbf{fitness} & \textbf{pct start} & \textbf{pct end} & \textbf{similarity} \\ \hline 
initial run & 0.814 & 0 & 82.0 &  1.0 \\ \hline
remove area  & 0.958 & 69.3 & 79.5 &  0.76 \\ 
reposition area  & 0.915 & 47.9 & 83.5 & 0.588\\
 & (0.917) & (54.0) & (86.5) & (0.6)\\ \hline 
\end{tabular}
\end{table}

First, we completely remove 12~agents from the area of the line structure shown in Fig.~\ref{fig:SAremoveArea}. 
This causes the swarm density to decrease from $44.4\%$ to $39.1\%$ and the solution quality to decrease by~$12.7$~percentage points~(pp). 
Here, we do one repair run as our simulation environment is deterministic (see Sec.~\ref{section:SAscenario}). 
During this repair run, solution quality increases as lines reform in the damaged area. 
At the end of the run, the recovered pattern has a solution quality that is only $2.5$~pp less than before, see Table~\ref{tab:SAdamage}.
The similarity to the initial structure is a high~$S=0.76$, which corresponds to 76~matching agent poses, out of a possible maximum similarity of~$S=0.88$. 

In a second test, we randomly reposition 12~agents uniformly outside of the area to keep the same swarm density as in the initial run. 
We do 20~independent repair runs. 
Replacing agents leads not only to the destruction of lines on the now-empty part of the grid, but randomly placed agents may also corrupt lines in other parts of the grid. 
As a result we observe  
a decrease in solution quality by~$34.1$~pp (see Table~\ref{tab:SAdamage}).
We find a slightly higher solution quality ($1.5$~pp) at the end of the repair run than at the end of the initial run. 
However, the similarity to the initial structure is lower than when removing the agents completely. 
The self-assembled structure is possibly more severely disturbed by randomly placed agents. 

In both scenarios, we find higher median fitness than in the initial run. 
Sensor values and predictions match more closely from the start of the run as agents are already partly assembled into the line structure initially in the evaluation. 
Overall, removing or repositioning agents triggers a reassembly of the pattern and we conclude that the evolved self-assembly behaviors are resilient to damage.

\begin{figure}[!t]
    \centering
    \subfloat[]{\includegraphics[width=\linewidth]{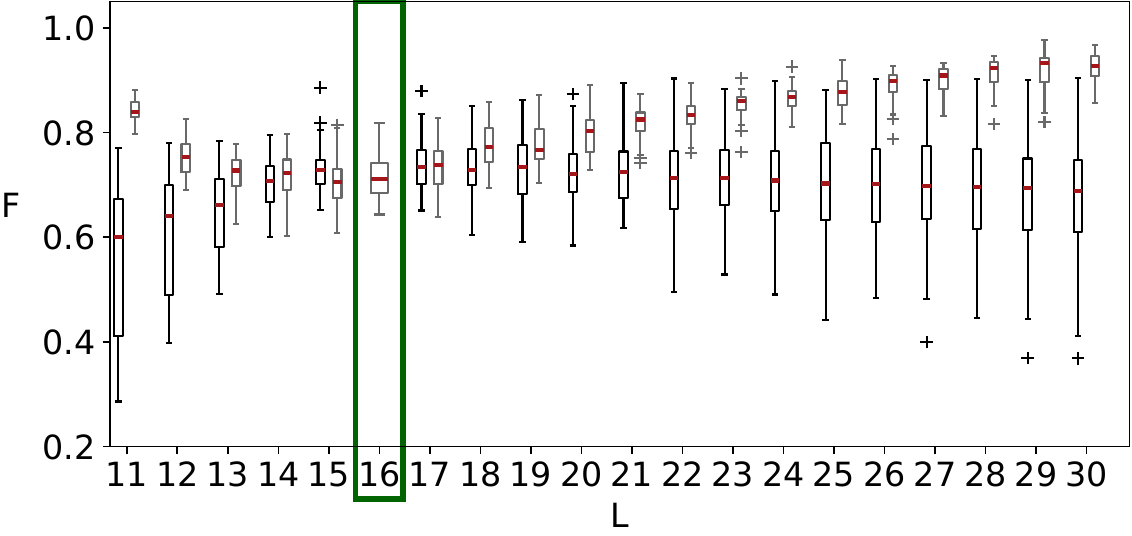}\label{fig:SAfitness_rerun}}\\
    \subfloat[]{\includegraphics[width=\linewidth]{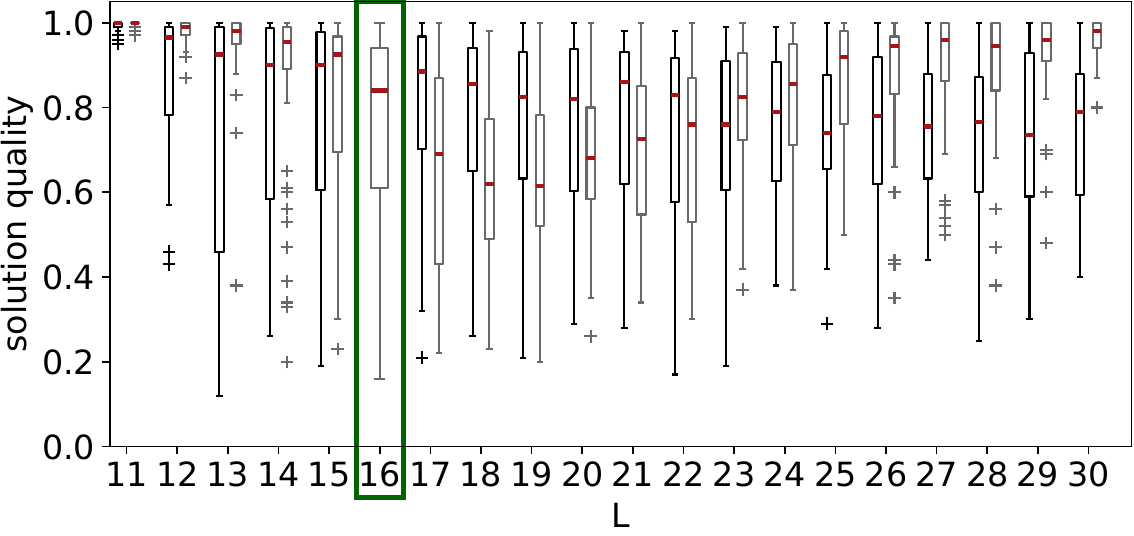}\label{fig:SAquality_rerun}} \\
    \subfloat[]{\includegraphics[width=\linewidth]{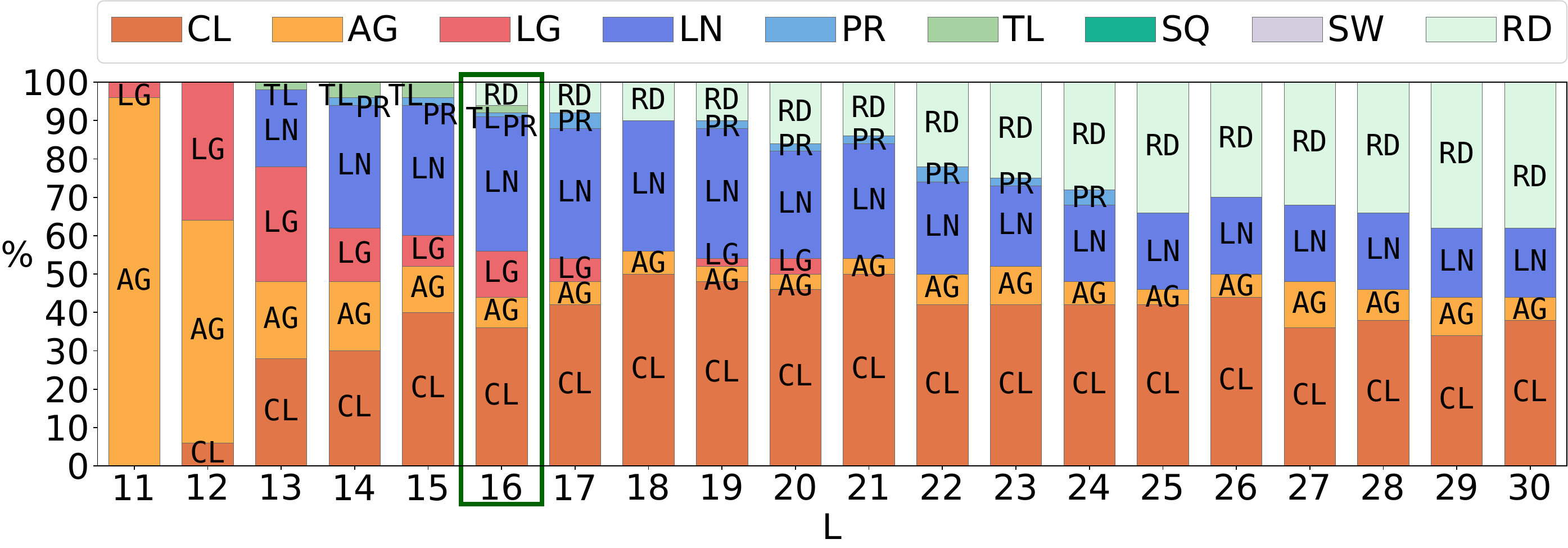}\label{fig:SAbehavrerun}}
    \caption{Scalability: reruns of the best evolved individuals of the $16\times 16$ grid on grid sizes $L\in [11, 30]$ (\textit{black boxes}): (a)~fitness of reruns per grid size, (b)~solution quality (i.e., percentage of agents positioned in dominant structure) per grid size, and (c)~percentage of resulting structures of reruns of controllers specialized for grid size $L_s=16$ on grid sizes $L\times L$ with $L\in[11,30]$ with clustering (CL), aggregation (AG), loose grouping (LG), lines (LN), pairs (PR), triangular lattice (TL), and random dispersion (RD). Medians are indicated by the \textit{red bars}. 
    The values of the best evolved individuals of each grid size (see Figs.~\ref{fig:SAfitness_evo},~\ref{fig:SAquality}) are given for reference and visualized by \textit{gray boxes}~\cite{kaiser2021a}.}
    \label{fig:SAfitReRun}
\end{figure}

\subsection{Scalability with Swarm Density} \label{section:SAscalability}

In Sec.~\ref{section:SAbehaviors}, we found that swarm density influences the emergence of behaviors.  
High swarm densities lead to more grouping behaviors (i.e., aggregation, clustering, loose grouping) while low densities lead mainly to dispersion. 
In this section, we analyze the scalability of the resulting behaviors with respect to swarm density.
As before, we keep the swarm size $N=100$ fixed while modifying the grid size $L$ to change swarm density ($\frac{N}{L\times L}$). 
We restrict our study to square tori as we do not expect much influence of the torus geometry on grouping and dispersion
while we already have studied the effects of the torus geometry on formed line structures before~\cite{kaiser19c}.
In the following, we compare the runs of agent controllers that were evolved and specialized for a specific swarm density and grid size~$L_s$ on different not-trained grid sizes~$L\ne L_s$ with the specialized controllers run on~$L_s$ and with controllers that are specialized for~$L$.   
Since we have found the most behavioral diversity on $L_s=16$, we exemplarily rerun each of the 50~best evolved individuals of the $16\times 16$~grid for 500~time steps on each grid size in $L\in [11,30]$ (hereafter referred to as reruns).
As visualized in Fig.~\ref{fig:SAfitReRun}, we compare fitness (Eq.~\ref{equ:fitness}), solution quality (i.e., percentage of agents positioned in dominant structure), and behavior distributions of these reruns (represented by black boxes in the box plots) with the original (specialized) evolutionary runs on the $16\times 16$ grid (framed in green; gray boxes for $L=16$ in the box plots) to determine if the performance of the best evolved individuals is independent of swarm density.
In a second step, we compare the reruns to the best evolved specialized controllers (represented by gray boxes in the box plots) of the respective grid sizes to evaluate 
the added value of evolving controllers specialized for given swarm densities.

\subsubsection{Comparison of the Best Evolved Individuals of the $16\times 16$~Grid with their Reruns in Different Swarm Densities}

First, we compare fitness and solution quality of the best evolved individuals of the $16\times 16$~grid (i.e., $L_s=16$; gray box for $L=16$) 
with their reruns on grid sizes $L\in[11,30]$ (black boxes).
We find statistically significantly lower fitness (Fig.~\ref{fig:SAfitness_rerun}) for the reruns on grid sizes $L\in [11,13]$ than for the original runs on the $16\times 16$~grid (MW-U with BC, $p<0.001$). 
There are no significant differences in fitness for all other grid sizes. 
Reruns on grid sizes $L\in [11,12]$ have statistically significantly better solution quality than the original runs on the $16\times 16$~grid, while we do not find significant differences for reruns on other grid sizes (MW-U with BC, $p<0.05$).
High swarm densities on small grid sizes $L\in [11, 13]$ allow only for grouping behaviors (i.e., clustering, aggregation, loose grouping).
Due to the limited amount of free grid cells, most agents are forced to be part of the forming grouping structure, which results in high solution quality. 
Behaviors that require free grid cells between some or all agents, like pairs, lines, or triangular lattices, cannot form during reruns in high swarm densities. 
For those behaviors, the predictor outputs do not match the formed structures anymore and we find lower fitness.

Next, we study the influence on behavior distributions when running robot controllers in swarm densities that they are not evolved for. 
We find no statistically significantly different behavior distributions (Fig.~\ref{fig:SAbehavrerun}) for reruns on grid sizes $L\in [13,24]$ compared to the behavior distribution on the $16\times 16$ grid (FE  with BC, $p<0.05$).
Reruns on even smaller ($L<13$) and even larger grids ($L>24$), that is, in higher and lower swarm densities, lead to significantly different behavior distributions. 
As described before, high swarm densities only allow for grouping behaviors and the original variety of patterns cannot form. 
Low densities, however, lead to agents likely being randomly dispersed and far apart from each other due to the initial uniformly random positioning.
Here, a combination of the initial agent positions, the evaluation length, and the behavior determined by the best evolved actor may influence whether the structure formed on the $16\times 16$~grid reforms in the rerun. 
Analyzing the scalability of each best evolved individual of $L=16$ individually indicates different structure formation approaches of the actors.
Half of the best evolved individuals lead to the formation of the same pattern or a pattern of the same group (e.g., to aggregation in the rerun when clustering on the $16\times 16$~grid) in the reruns on grid sizes $L\in [13,30]$.  
The other half of the best evolved individuals only forms the initial pattern in some of the other swarm densities. 
In these cases, the reruns in higher swarm densities mostly lead to grouping behaviors while the reruns in lower densities frequently result in random dispersion. 
In both groups, we find the same variety of formed structures on the $16\times 16$~grid and thus the best evolved individuals seem to vary in their approach to structure formation by adapting more or less to and exploiting swarm densities during evolution.  
In summary, the non-specialized controllers keep their clustering, aggregation, and line behaviors even in low swarm densities. Hence, this scaling ability can for example be exploited to provoke desired behaviors (see Sec.~\ref{section:SAengineeredSO}) also in low-density situations. 

Overall, we find similar fitness, solution quality, and behavior distributions for reruns on grid sizes $L\in [14,24]$ and the original runs of best evolved individuals on the $16\times 16$~grid.
The best evolved individuals scale well with swarm density and can be reused in other swarm densities, but the initial performance cannot be maintained for all behaviors for high and low densities.

\subsubsection{Comparison of Reruns with the Best Evolved Individuals of the Respective Swarm Density}

Previously, we compared the best evolved individuals of the $16\times 16$~grid with their reruns on other grid sizes $L\in [11,30]$.
Next, we compare reruns of best evolved individuals of the $16\times 16$~grid (black boxes) with best evolved specialized individuals of respective grid sizes (gray boxes) for each grid size~$L$. 
We find statistically significantly greater fitness (Fig.~\ref{fig:SAfitness_rerun}) for best evolved individuals of the respective grid size than for the reruns on that grid size except for $L\in [14,17]$ (MW-U with BC, $p<0.05$). 
There are no statistically significant differences in solution quality (Fig.~\ref{fig:SAquality_rerun}) for best evolved individuals and reruns on grid sizes $L\in [11,15]$ and $L\in[20,24]$ (MW-U with BC, $p<0.05$). 
For grid sizes $L\in [17,19]$, we find significantly better solution quality in reruns than in best evolved individuals of the respective grid size and for grid sizes $L\in [25,30]$ solution quality is significantly lower for reruns. 
There is no significant difference in behavior distributions (Fig.~\ref{fig:SAbehavrerun}) of reruns and best evolved individuals on grid sizes $L\in\{11, 14, 15, 17\}$ (FE with BC, $p<0.001$).  
All other behavior distributions are statistically significantly different. 

Overall, we find that for grid sizes $L\in [14,17]$ reruns and best evolved specialized individuals for these grid sizes are competitive in fitness, solution quality, and behavior distributions. 
This was expected as we have also not found any significant differences between the evolutionary runs for those swarm densities (see Sec.~\ref{section:SAbehaviors}). 
Thus, best evolved individuals of one swarm density can easily be reused in similar swarm densities. 
For grid sizes $L\in [17,19]$, reruns lead to higher solution quality and similar or lower fitness than best evolved individuals of the respective grid sizes. 
The evolutionary runs on these grid sizes produced the lowest solution quality of all scenarios (see~Sec.~\ref{section:SAbehaviors}), which implies that the evolution of behaviors leading to forming the defined patterns (see Sec.~\ref{section:SAmetrics}) is especially challenging for these swarm densities.
Here, the rerun of best evolved individuals of lower swarm density (i.e., in this scenario of the $16\times 16$ grid) is beneficial when high solution quality is required and prediction accuracy (i.e., fitness) is unimportant.
This is the case when we use the minimize surprise approach in an offline manner~\cite{doncieux2015evolutionary}, that is, we separate optimization and operational phase. 
In the optimization phase, prediction accuracy is used as the means to guide evolution towards interesting swarm behaviors. 
In the operational phase, actor-predictor pairs will no longer be optimized. 
As the actor of an ANN pair determines the agent's next action independently from the predictor, the latter is not required for the mere execution of an evolved swarm behavior. 

For significantly higher or lower swarm densities, the best evolved actor-predictor ANN pairs of the respective grid size are better adapted to each other and to the environment and thus reach higher fitness and solution quality. 
Nevertheless, reruns of best individuals of another swarm density can be beneficial for structures that may not or only rarely form by evolution in a density, such as lines on the $30\times 30$~grid.

\subsection{Engineered Self-Organization} \label{section:SAengineeredSO} 

In all previous experiments, we run minimize surprise with complete freedom, that is, we rely completely on the innate motivation by prediction accuracy. 
This way we have no direct influence on the emergent variety of behaviors.
Here, we present an approach to push evolution towards the emergence of desired structures by predefining some or all of the agents' predictor outputs $p_0,\dots,p_{R-1}$ to fixed, desired values~\cite{kaiser19c}, that is, we indirectly specify a task-specific reward. 
High fitness can then only be reached by behaviors that create sensor input satisfying these fixed sensor predictions.
However, as before, the unfixed ones can still be adapted by the predictor. 
As an example, we aim for the emergence of lines (see Fig.~\ref{fig:SAline}) on grid sizes $15\times 15$ and $20\times 20$ in our self-assembly scenario (see Sec.~\ref{section:SAscenario}). 
A~study on the effects of the torus geometry (i.e., rectangular grids) on the orientation of the evolved line formations when predefining predictions can be found in~\cite{kaiser19c}.  
Furthermore, the evolution of grouping behaviors with predefined predictions can be found in~\cite{kaiser20b}. 
We do 50~independent evolutionary runs per scenario. 

\begin{figure}[!t]
    \centering
    \includegraphics[width=0.8\linewidth]{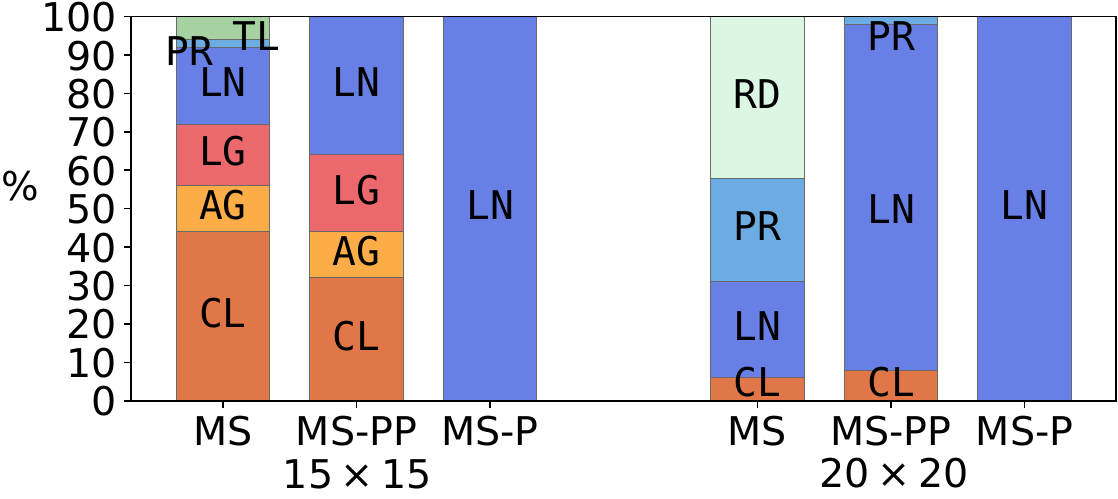}
    \caption{Percentage of resulting structures in minimize surprise~(MS), minimize surprise with partially predefined predictions~(MS-PP), and minimize surprise with predefined predictions~(MS-P) aiming for the formation of lines with clustering~(CL), aggregation~(AG), loose grouping~(LG), lines~(LN), pairs~(PR), triangular lattice~(TL), and random dispersion~(RD) for grid sizes $15\times 15$ and $20\times 20$~\cite{kaiser2021a}.}
    \label{fig:SAMSMSP}
\end{figure}

First, we partially predefine the predictions of the sensors directly in front and behind the agent to~1 (i.e., $s_0=s_3=s_8=s_{11}=1$; see~Fig.~\ref{fig:SAsensormodel}). 
The predictor still has to predict the other ten sensor values. 
Hence, we facilitate the emergence of lines and grouping behaviors. 
Predicting no more neighbors than the predefined ones (i.e., predicting 0 for the ten unfixed predictor outputs) matches exactly the predictions for line structures, but predicting several more neighbors matches the predictions for grouping behaviors. 
All other behaviors would need some predefined predictions to be 0 (instead of 1). 
We keep all parameters as before (see Table~\ref{tab:SAparameters}) except for the mutation rate for the $20\times 20$~grid, which we set to $0.3$ to obtain a converging fitness curve.
This was necessary probably because task difficulty may be increased on the $20\times 20$~grid.  
While we already find a majority of lines and grouping behaviors on the $15\times 15$~grid when running the approach with complete freedom, only $31\%$ of the runs lead to such behaviors on the $20\times 20$~grid. 
Thus, we push evolution towards the search for solutions that are otherwise rarely found in this scenario by partially predefining predictions. 
We find $64\%$ grouping behaviors (i.e., clustering, aggregation, loose grouping) and $36\%$ lines on the $15\times 15$ grid (Fig.~\ref{fig:SAMSMSP}). 
Compared to the runs with complete freedom (see Sec.~\ref{section:SAbehaviors}),  formations of line structures increase by $16$~pp and grouping behaviors decrease by $8$~pp. 
Pairs and triangular lattices do not emerge anymore. 
On the $20\times 20$~grid, we find $90\%$ lines, $2\%$ pairs, and $8\%$ cluster. 
Thus, lines increase by $65$~pp and clustering by $2$~pp, while pairs decrease by $25$~pp and random dispersion does not emerge anymore. 
As expected, only lines and grouping behaviors emerge except for one pair structure. 
Hence, we could push emergence towards lines and grouping behaviors by predefining some sensor values, but the system still depends on swarm density. 

Next, we predefine all predictor outputs to push evolution further towards the emergence of line structures.
As before, we predefine the sensors directly in front and behind the agent to~1 (i.e., $s_0=s_3=s_8=s_{11}=1$). 
Additionally, we fix all other predictor outputs to~0. 
As all predictor outputs are fixed, we remove the predictor.  
The actor still receives indirect selective pressure as we continue to reward `prediction' accuracy (Eq.~\ref{equ:fitness}). 
In this case, line structures form in all runs on both grids. 
We can interpret this approach as a special way of defining a task-specific fitness function that leads to the evolution of a desired structure without requiring global information. 

\subsection{Hyperparameter Optimization} 
\label{section:SAhyperparameter}

\begin{table}[!t]
\caption{Tested values for hyperparameter optimization.\label{tab:SAoptimization}}
\centering
\begin{tabular}{lll}
\hline 
 \textbf{parameter} & \textbf{value range} & \textbf{step size} \\ \hline 
 population size~P & [10, 80] & 10 \\ 
 number of generations~G & [50, 100] & 10\\ 
 evaluation length $T$ (time steps)  & 50, [100, 500] & 100 \\ 
 \# of sim. runs per fitness evaluation & $[1, 10]$ & 1 \\  
 mutation rate &  [0.001, 0.4] & 0.004 - 0.1 \\  \hline 
\end{tabular}
\end{table}

\begin{figure}[!t]
    \centering
     \includegraphics[width=\linewidth]{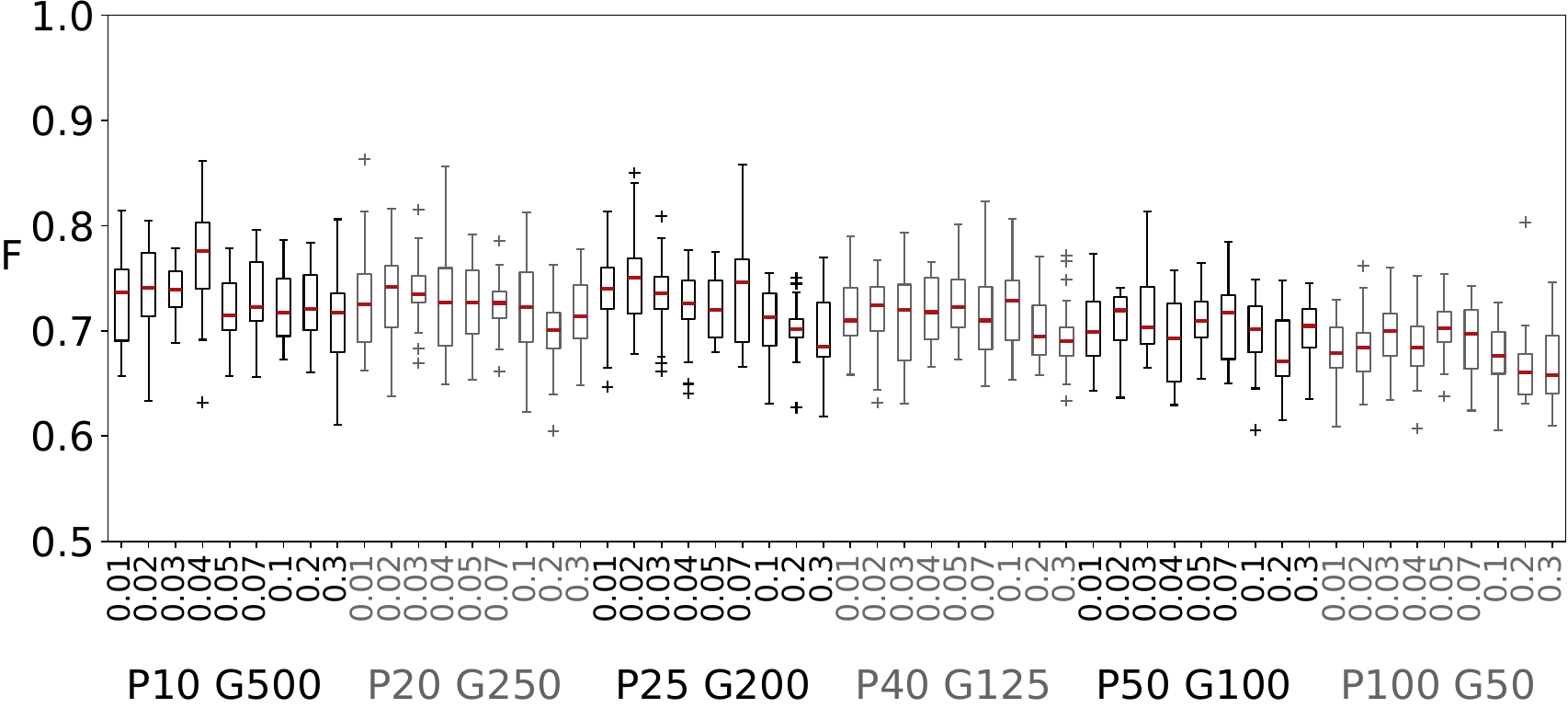}
    \caption{Hyperparameter tests: Best fitness per combination of mutation rate, population size (P), and number of generations (G) with $P\times G = 5{,}000$ evaluations per setup on the $15\times 15$~grid. Tested mutation rates 
    per combination of population size~P and number of generations~G are from left to right: 0.01, 0.02, 0.03, 0.04, 0.05, 0.07, 0.1, 0.2, 0.3. Medians are indicated by red bars~\cite{kaiser2021a}. \label{fig:hyperopti}}
\end{figure}

We have used a fixed set of hyperparameters to evolve swarm behaviors with minimize surprise. 
Here, we test how the number of simulation runs per fitness evaluation (repetitions), evaluation length, population size, number of generations, and mutation rate influence the resulting fitness (Eq.~\ref{equ:fitness}). 
We do 20~independent evolutionary runs per tested parameter combination.
We test two different grid sizes, $15\times 15$ and $20\times 20$, as swarm density affects the emergence of behaviors and thus different parameter sets may be optimal. 

First, we vary one hyperparameter while fixing the rest to their initial values (see Table~\ref{tab:SAparameters}). 
Table~\ref{tab:SAoptimization} summarizes the tested values. 
For one to ten repetitions, we do not find significantly different fitness for both grid sizes (KW, $p<0.05$).
Evaluation lengths between 50 and 500 time steps do not lead to significant differences in fitness on the $20\times 20$ grid, but this only holds true for 200 to 500 time steps on the $15\times 15$~grid (MW-U with BC, $p<0.05$). 
Numbers of generations between 50 and 100 do not lead to significantly different fitness on the $15\times 15$~grid (KW, $p<0.05$). 
On the $20\times 20$~grid, we find significantly better fitness for 90~generations than for numbers of generations between 50 and 70 but no significant differences for numbers of generations between 80 and 100 (MW-U with BC, $p<0.05$).
Population sizes between ten and 80 do not lead to significant differences on the $15\times 15$~grid. 
On the $20\times 20$~grid, population sizes of 50 and more have significantly better fitness than population size ten (MW-U with BC, $p<0.05$). 
We do not find statistically significant differences for mutation rates between $0.01$ and $0.3$ on both grid sizes (MW-U with BC, $p<0.05$). 
Other mutation rates, however, may lead to significantly lower fitness. 
In total, we find that we can speed up evolution by reducing repetitions and evaluation length without loss in fitness. 
The total number of evaluations (i.e., $\text{population size} \times \text{number of generations}$) doesn't impact fitness on the $15\times 15$~grid significantly, but 4,000 or more evaluations should be done on the $20\times 20$~grid. 

Next, we test different combinations of mutation rate, population size, and number of generations. 
We do 5,000 evaluations divided between population sizes of 10 to 100 and 50 to 500~generations. 
Thereby, the population size influences the width (i.e., exploration of the solution space) and the number of generations influences the depth of the search.
For all combinations of population size and generations, we vary mutation rates between $0.01$ and $0.3$, which results in a total 54~hyperparameter combinations per grid size. 
We do two simulation runs of 200~time steps each per fitness evaluation. 
Fig.~\ref{fig:hyperopti} visualizes best fitness for the different hyperparameter combinations on the $15\times 15$~grid. 
On both grid sizes, we find statistically significantly different fitness for the different hyperparameter combinations (KW, $p<0.01$).
We find the maximum mean best fitness for population size~10 and 500~generations, that is, $0.78$ for mutation rate~$0.04$ on the $15\times 15$~grid and $0.84$ for mutation rate~$0.05$ on the $20\times 20$~grid. 
In total, we find no significant differences for population sizes of ten to 40 and 125 to 500~generations for all mutation rates (MW with BC, $p<0.01$). 
Higher population sizes and thus fewer generations may lead to significantly lower fitness than some other hyperparameter settings. 
In summary, we find that for both grid sizes (i.e., swarm densities) low population sizes and many generations are best for evolving self-assembly behaviors with minimize surprise. 
All other parameters seem to be insignificant.

\section{Realistic Simulations and Real Robot Experiments} \label{section:reality}

\subsection{Approach} \label{section:REALapproach}

In this section, we present the robot platform (Sec.~\ref{section:RealRobot}) and evolution architecture (Sec.~\ref{section:realEvo}) used in our experiments in realistic simulations and with the real robots. 

\subsubsection{Robot} \label{section:RealRobot}

\begin{figure}[!t]
    \centering
    \subfloat[\label{fig:simThymio}]{\includegraphics[width=0.38\linewidth]{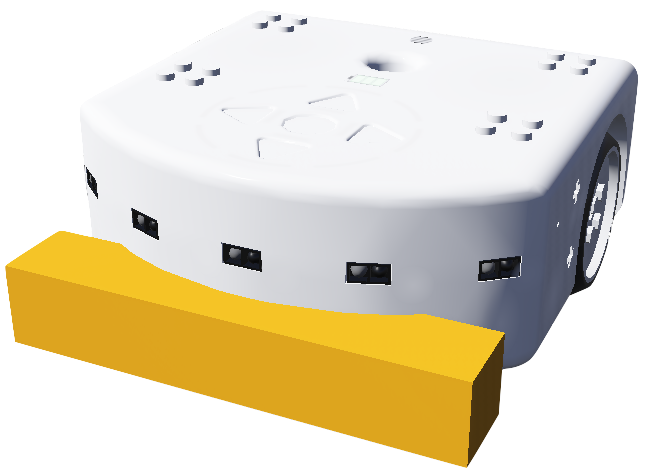}} \hspace{3mm}
    \subfloat[\label{fig:sensors}]{\includegraphics[width=0.29\linewidth]{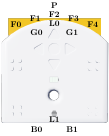}}\\
    \subfloat[\label{fig:realThymio}]{ \includegraphics[width=0.45\linewidth]{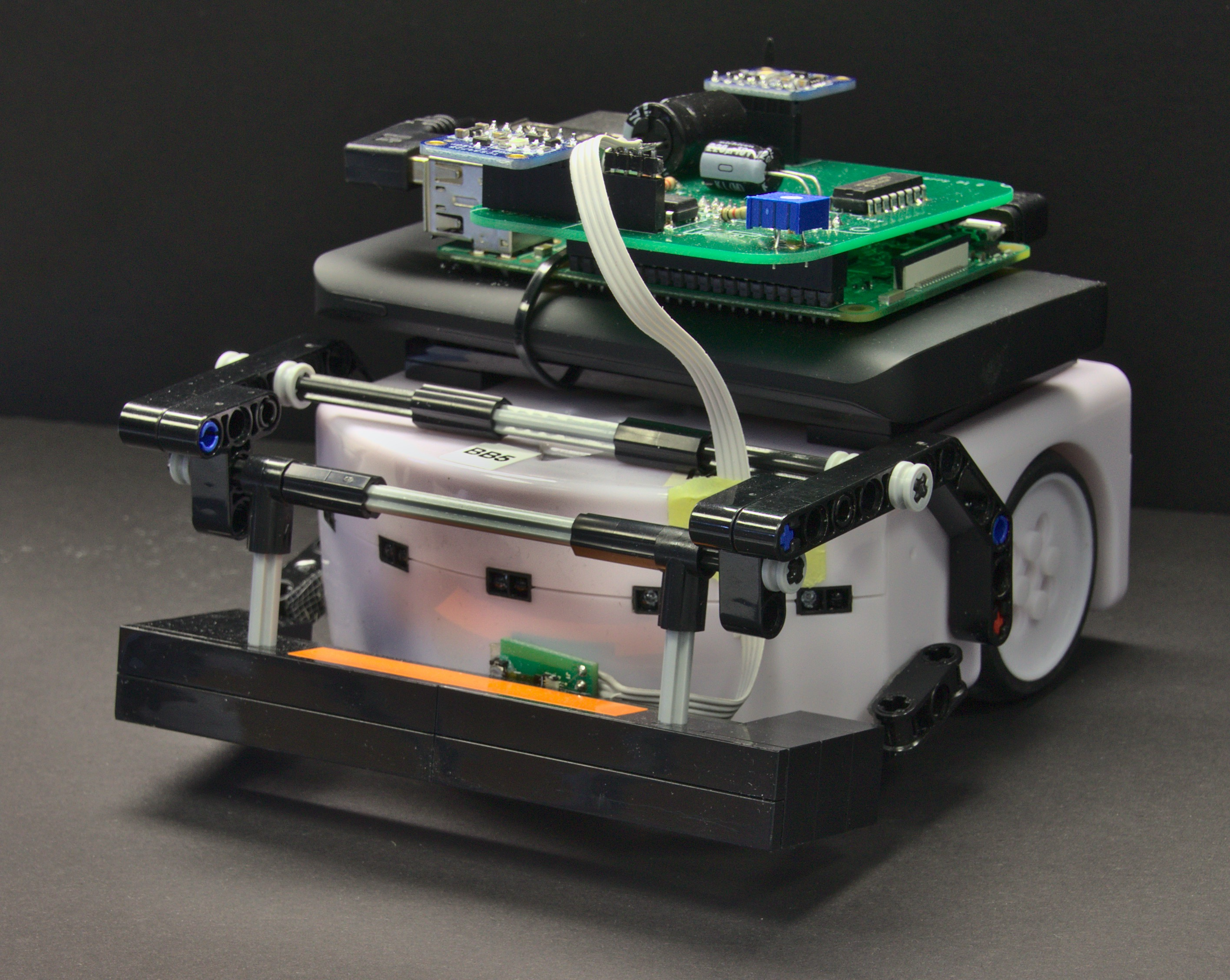}}
    \caption{Extended Thymio II robot (a)~in the Webots simulator and (c)~in reality. (b)~Sensor positions of the robot's 7~horizontal IR sensors ($F0, \dots, F4$, $B0$, $B1$), 2~ground IR sensors ($G0, G1$), 2~light sensors ($L0, L1$; invisible in simulation), and 1 force sensor~($P$). 
    Reprinted from~\cite{kaiser2020}.
    \label{fig:Thymio}}
\end{figure}

We use the differential drive mobile robot Thymio~II~\cite{Riedo2013} both in the Webots~\cite{Webots04} simulator and in our real robot experiments. 
It has a maximum linear velocity of~$20\;\frac{\text{cm}}{\text{s}}$, but we restrict the maximum speed to $12.6\;\frac{\text{cm}}{\text{s}}$ in simulation and to $10.6\;\frac{\text{cm}}{\text{s}}$ on the real robots to reduce wear of the motors. 
The Thymio~II has, amongst other sensors, seven horizontal infrared (IR) proximity sensors whereof five are located at the front ($F0,\dots,F4$), and two are located at the back ($B0$ and $B1$) of the robot and two IR ground sensors ($G0$ and $G1$). 
The IR sensors are updated every $100~\text{ms}$ on the real robot and every $10~\text{ms}$ (simulated time) in Webots. 
Thus, time is discretized into steps of $100~\text{ms}$ and $10~\text{ms}$ respectively and we can calculate fitness (Eq.~\ref{equ:fitness}) in time steps as before. 
We extend the real Thymio~II with a Raspberry Pi~3B (RPi) that is connected via USB and the D-Bus interface to the Thymio~II and an external battery. 
This enables us to program the robot with Python~3 and to add additional sensors. 
For the experiments in Sec.~\ref{section:REALbp}, we extend the robot with two light sensors ($L0$ and $L1$) and a pressure sensor ($P$) that are connected to the RPi.
The light sensors (TSL45315) are on top of the robot and detect light gradients. 
The force sensor (HSFPAR303A) measures forces when the Thymio~II pushes objects. 
For this purpose, we add a bulldozer blade in a bumper style that is built out of LEGO\textsuperscript{\textregistered} parts and mounted to the LEGO\textsuperscript{\textregistered} attachment points on the Thymio~II, see Fig.~\ref{fig:realThymio}.
In Webots, we modify the open-source PROTO-files of the Thymio~II Webots model to add the sensors and the blade, see Fig.~\ref{fig:simThymio}. 
Fig.~\ref{fig:sensors} illustrates the sensor positions on the extended Thymio~II robot.

\subsubsection{Evolution Architecture} \label{section:realEvo}

In Sec.~\ref{section:AnalysisApproach}, we evolved swarm behaviors using a genetic algorithm that conducted over 5,000~evaluations per evolutionary run in simulation.  
This would be too time consuming and would wear down our hardware on the real robots~\cite{doncieux2015evolutionary}. 
Therefore, we switch to a genetic algorithm in an online and onboard architecture with (1+1)-selection~\cite{Eiben2015,bredeche2010} for our experiments here~\cite{kaiser21}. 
As before, robots are equipped with an actor-predictor ANN pair (see Sec.~\ref{section:approach}). 
The actor ANN (i.e., robot controller) outputs two normalized speeds $v \in [-1.0,1.0]$ for the two differential drive motors with every sensor update. 
The normalized speeds~$v$ are scaled with the maximum speed~$v_\text{max}$ when sent to the robot's motors.  
ANN inputs are the current sensor values that are normalized by their maximum possible value. 
Additionally, the actor receives the last set of normalized speeds and the predictor network the next normalized speeds as input. 
Online evolution allows for infinitely continued adaptation, but we stop evolution after 1,000 evaluations in simulation and 350 on the real robots due to limitations in battery life. 
We evaluate each individual for 10~s, that is 1,000 time steps in simulation (sensor update rate of 100~Hz) and 100~time steps on the real robots (sensor update rate of 10~Hz).
Inputs are fed to the networks with every sensor update.
Robots start an evaluation at the last position of the previous evaluation. 
With a $20\%$ chance, the current best individual is re-evaluated. 
Its updated fitness is calculated as an exponentially weighted mean $F_t=\alpha \hat{f_t} +(1-\alpha)F_{t-1}$ with $\alpha=0.2$, the fitness value~$\hat{f_t}$ reached during re-evaluation, and the previous best fitness~$F_{t-1}$.
Otherwise, offspring is created by adding a uniformly random value in the range~$[-0.4, 0.4]$ to each value with a mutation rate of~$0.1$ per value, and evaluated.  
To evaluate the variety of behaviors that emerges during evolution, we re-run the best individual of the last generation for $100~\text{s}$ to store sensor values, predictions and, in simulation, the robot trajectories giving the robot poses $(x,y,\theta)$ over runtime. 
We reset the arena for this post-evaluation by placing robots at random positions (Sec.~\ref{section:REALbasic}) or their initial positions (Sec.~\ref{section:REALbp}). 
Table~\ref{tab:parameters} summarizes the parameters. 

\begin{table}[!t]
\centering
  \caption{Parameters for (1+1)-evolution in realistic simulations and real robot experiments.}
  \label{tab:parameters}
  \begin{tabular}{lll}
    \hline
    \textbf{parameter}&\textbf{simulation}&\textbf{real robots}\\
    \hline
    mutation rate & 0.1 & 0.1\\
    max. evaluations & 1000 & 350 \\ 
    evaluation length (time steps) & 1,000 & 100 \\
    post-evaluation length (time steps) $T$ \hspace{0.5em} & 10,000 & 1,000 \\
    re-evaluation probability & 0.2 & 0.2 \\ 
    re-evaluation weight $\alpha$ & 0.2 & 0.2 \\ 
    \hline
\end{tabular}
\end{table}
 
 \begin{figure}[t]
    \centering
    \includegraphics[width=0.85\linewidth, trim={0.3cm 0.4cm 0.4cm 0.3cm},clip]{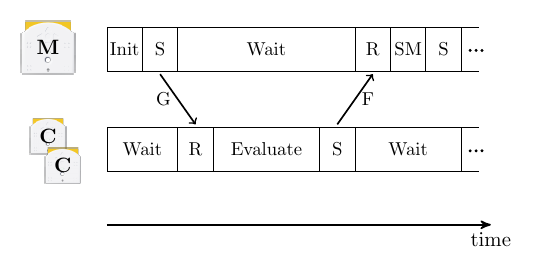}
    \caption{Centralized online evolution architecture. The external master robot~(M) runs the evolutionary process: it initializes the first genome (G) and sends (S) it to the clients (C). Clients evaluate the received (R) genome and send their individual fitnesses (F) back. The master determines the overall fitness, selects the current best individual, decides if it will be reevaluated or creates offspring by mutation (SM), and sends the respective genome to the clients.
    The process continues until terminated. 
    Reprinted from~\cite{kaiser2020}.
    }\label{fig:onlineEvo}
\end{figure}

For the experiments with real robots and in Webots, we use a centralized online evolution architecture~\cite{Eiben2015}. 
One Thymio~II robot serves as a central master to guide the evolutionary process.
The master distributes genomes, collects individual fitnesses, calculates the overall fitness, selects the best individual, and creates offspring.  
We place the master outside the arena (i.e., it is not evaluating genomes itself) to prevent experiment abortions due to hardware defects of the master in the real robot experiments. 
Still, evolution is running fully onboard on the real robots. 
The swarm members (clients) receive the genomes from the master, evaluate them, and send their individual fitnesses back. 
Fig.~\ref{fig:onlineEvo} illustrates this interplay schematically. 
The communication between master and clients is realized in the simulation with Webots' built-in Emitter and Receiver nodes. 
On the real robots, we use WiFi and a TCP connection. 
Genomes are evaluated again in the next evaluation in the case of transmission errors. 
We run experiments with simulated and real Thymio~II robots in arenas with carpeted floor. 
The robots use their ground IR sensors to detect arena boundaries, which are the edges of the carpeted floor in simulation and mirror film in the real arena.
Robots driving backwards detect the arena boundaries only when they are almost completely out of the arena as their ground IR sensors are located in their front.
Therefore, we add walls in distance to the real arena boundaries so that robots cannot leave the arena completely. 
We do 20~independent evolutionary runs per scenario (i.e., different arenas) in simulation and five independent evolutionary runs with the real robots. 
As before, we compare the evolutionary runs to 20~randomly generated ANN pairs per scenario in simulation (see Sec.~\ref{section:SAeffectivity}).

\subsection{Simple Swarm Behaviors} \label{section:REALbasic} 

In a first step, we aim to evolve basic swarm behaviors in the Webots simulator with a swarm of standard Thymio~II mobile robots. 
We describe the experimental setup first and present the results in simulation afterwards. 
A video of emergent behaviors is online~\cite{kaiser2021a}. 

\subsubsection{Experimental Setup}

We use a swarm of $N=10$~Thymio~II robots and one master robot in simulation. 
Swarm density is varied by placing the ten robots into square arenas with side lengths $L\in \{0.7\,\text{m},0.8\,\text{m},0.9\,\text{m},1.0\,\text{m},1.5\,\text{m}\}$. 
In this scenario, the walls in distance to the arena boundaries are low so that they are not detectable by the horizontal IR sensors.
As discussed above, these walls prevent the robots from leaving the arena completely.
We add a simple hardware protection layer to prevent robot damage. 
Robots stop if they detect other robots with their front or back horizontal IR sensors.
They also stop when detecting the arena boundaries with their ground IR sensors while aiming to drive straight. 
By keeping the hardware protection simple, we give the robots freedom to evolve their own obstacle avoidance behaviors, but we prevent only serious crashes between robots. 

We classify the resulting behaviors based on the robots' rotation and the mean cluster size in the last $\tau~=~5{,}000$~time steps of the post-evaluation run. 
Consequently, we give the robots time in the first half of the run to cover some distance and initiate their dominant behavior.
We do not consider the distance covered by robots here as it is low in all runs. 
We define rotation~$c_o$ as the absolute accumulated difference in orientation over the last $\tau$~time steps as given by 
\begin{equation}
c_o = |\frac{1}{N}\sum^{N-1}_{n=0}\sum^{T-1}_{t=T-\tau} (\theta_n(t+1)-\theta_n(t))|\, ,
\end{equation}

with $N$~robots, run time of $T$~time steps, and orientations~$\theta(t)$ and~$\theta(t+1)$ of robot~$n$.  
Robots can reach a maximum rotation of $126~\text{rad}$ when constantly turning with maximum rotational speed. 
Using an empirical approach, we then differentiate between stopped ($c_o < 5$), diverse, and circling ($c_o > 55$; i.e., robots turn on spot) behaviors.
Diverse behaviors can range from robots turning very slowly or in large circles to swarms which are partially stopped and partially circling. 
Furthermore, we define that clusters are formed out of robots that have max. twice the robot radius (i.e., $\approx 15.75\,\text{cm}$) distance to a minimum of one other cluster member. 
We differentiate between behaviors leading to grouped and dispersed robots based on a threshold for mean cluster size of $1.43$ which corresponds to five grouped robots.

\subsubsection{Results}

\begin{figure}[!t]
    \centering
     \includegraphics[width=0.85\linewidth]{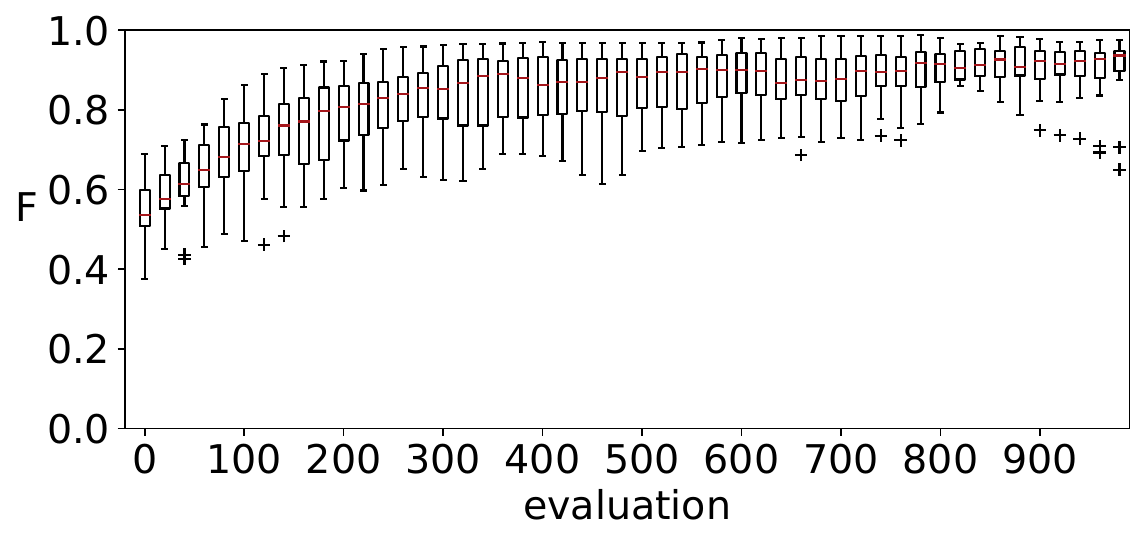}
    \caption{Realistic simulation: current best fitness F over evaluations for the $1.0\,\text{m} \times 1.0\,\text{m}$ arena in simulation over 20 runs. For clearer illustration, only every 20th evaluation is printed. Medians are indicated by red bars~\cite{kaiser2021a}. \label{fig:REALfit_curves_sim}}
\end{figure}

\begin{figure}[!t]
\centering
    \subfloat[\label{fig:REALpairs}]{\includegraphics[width=0.4\linewidth]{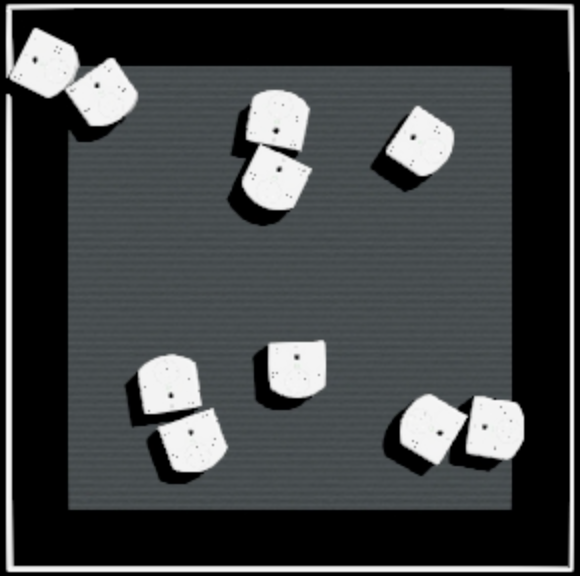}}\hspace{0.1em}
    \subfloat[\label{fig:REALclusters}]{\includegraphics[width=0.4\linewidth]{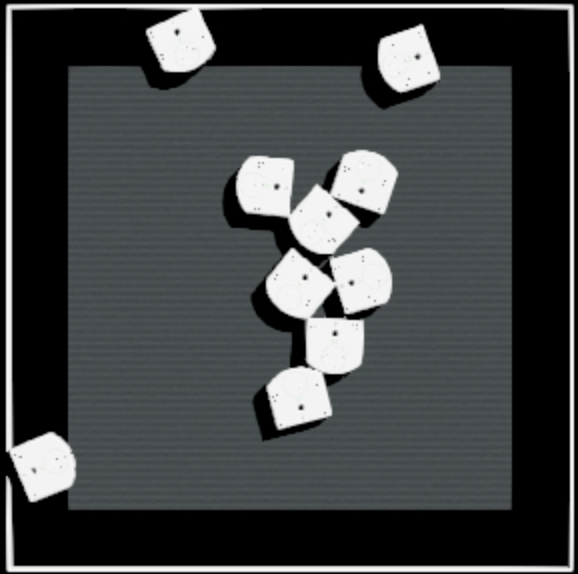}}
    \caption{Realistic simulation: robot positions for (a)~pairs and (b)~seven grouped robots at the end of two post-evaluation runs in the $0.9\,\text{m}\times 0.9\,\text{m}$ arena in simulation~\cite{kaiser2021a}.  \label{fig:REALcluster}}
\end{figure}

Fig.~\ref{fig:REALfit_curves_sim} shows the increase of best fitness over evaluations for the $1\,\text{m}\times 1\,\text{m}$ arena exemplarily for all runs. 
We reach a median best fitness of at least $0.88$ for the evolutionary runs and of maximum $0.58$ for the randomly generated ANN pairs for all arena sizes. 
Consequently, the best evolved individuals clearly outperform the random individuals in prediction accuracy.

\begin{figure}[!t]
\centering
    \subfloat[]{\includegraphics[width=0.8\linewidth]{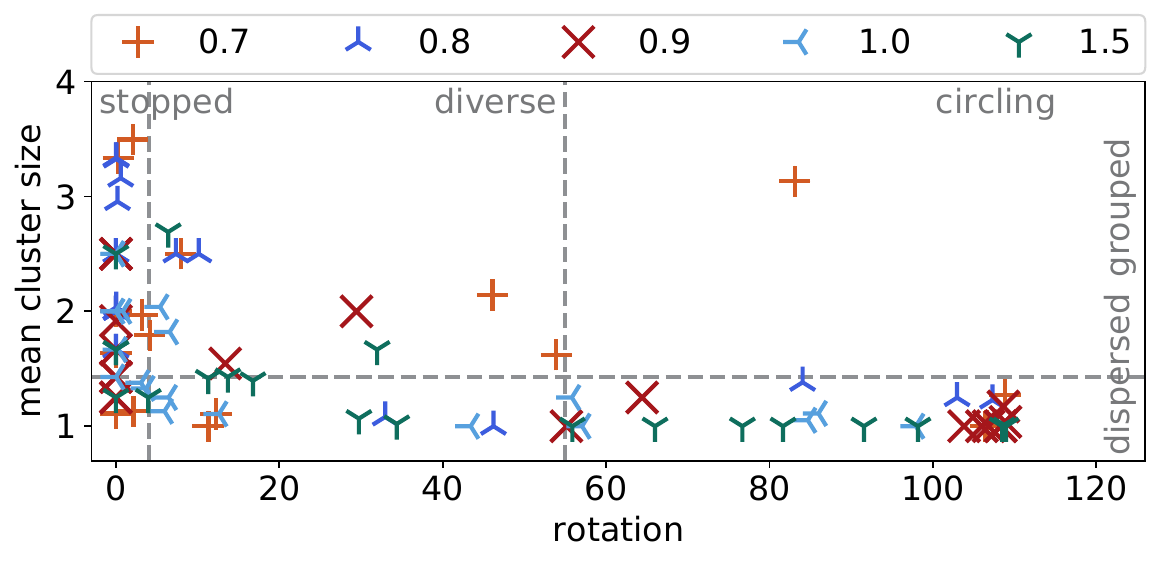}} \\
    \subfloat[]{\includegraphics[width=0.8\linewidth]{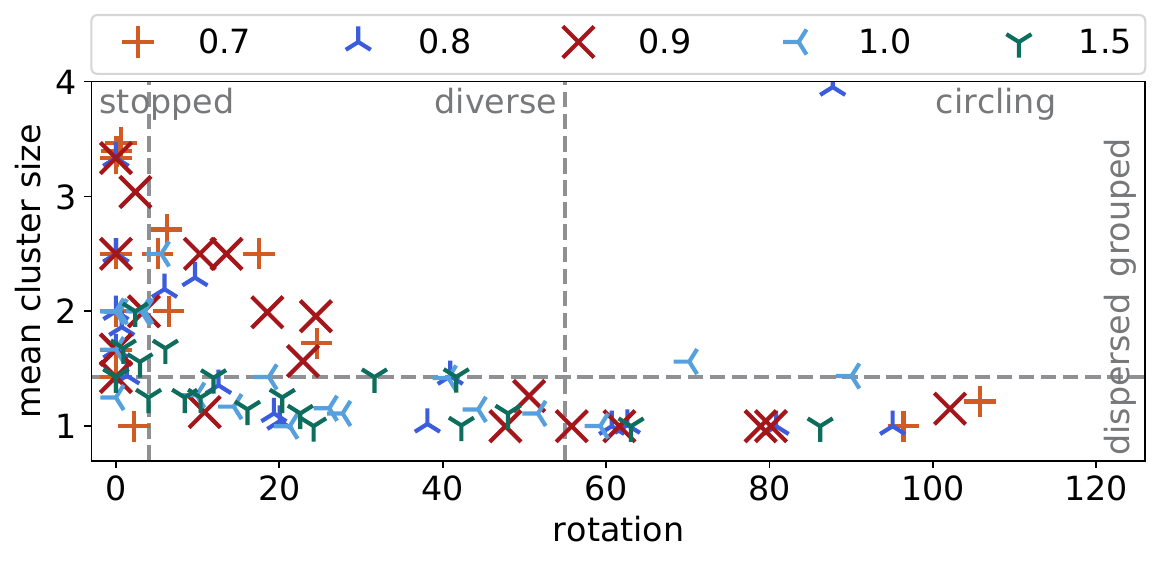}}
    \caption{Realistic simulation: rotation vs. mean cluster size over the last $5{,}000$ time steps for (a)~the best evolved individuals and (b)~the randomly generated individuals with 20~independent evolutionary runs per arena size. The gray lines categorize the resulting behaviors~\cite{kaiser2021a}. \label{fig:rot_clust}}
\end{figure}

We analyze the behavior diversity by post-evaluating the best evolved individuals and by comparing to randomly generated ANN pairs.
Over all arena sizes, we find more diverse behaviors for randomly generated ANN pairs than for the best evolved individuals, see Fig.~\ref{fig:rot_clust}.
The majority of diverse behaviors lead to swarms that are partially stopped and partially moving. 
We assume that predicting sensor inputs is harder as sensor values are time variant.
By contrast, evolution mostly leads to easy-to-predict behaviors with constant sensor values. 
Most behaviors lead to stopped robots that are grouped or dispersed, or to dispersed circling robots. 
Grouped robots (Fig.~\ref{fig:REALcluster}) are stopped by hardware protection in most cases guaranteeing constant sensor input. 
Rare cases lead to robots staying in their current position by quickly switching turning direction or robots turning constantly on spot while still remaining close enough to be considered as grouped in high swarm densities. 
Both dispersed circling and stopped swarms have constant low sensor values.
We find that robots frequently distribute over the arena by employing an emergent obstacle avoidance behavior in both variants. 
In the case of dispersed stopped swarms, the robots drive to different parts of the arena boundary where they are stopped by hardware protection. 

Overall, we find a few simple swarm behaviors when running minimize surprise on the simulated Thymio~II robots. 
Thus, we aim for more complex behaviors in the next step.

\subsection{Object Manipulation Behaviors} \label{section:REALbp}

Next, we aim for behaviors leading to robot-robot and robot-environment interactions by placing boxes that can be pushed by robots in the environment.
We describe the experimental setup (Sec.~\ref{section:REALbpSetup}) and our results in simulation (Sec.~\ref{section:REALbpSim}) and in real robot experiments (Sec.~\ref{section:REALbpReal}). 
A more detailed study of the results including the study of a second box density can be found in~\cite{kaiser21}. 

\subsubsection{Experimental Setup} \label{section:REALbpSetup}

\begin{figure}[t]
    \centering
    \subfloat[\label{fig:realArena}]{\includegraphics[width=0.52\linewidth]{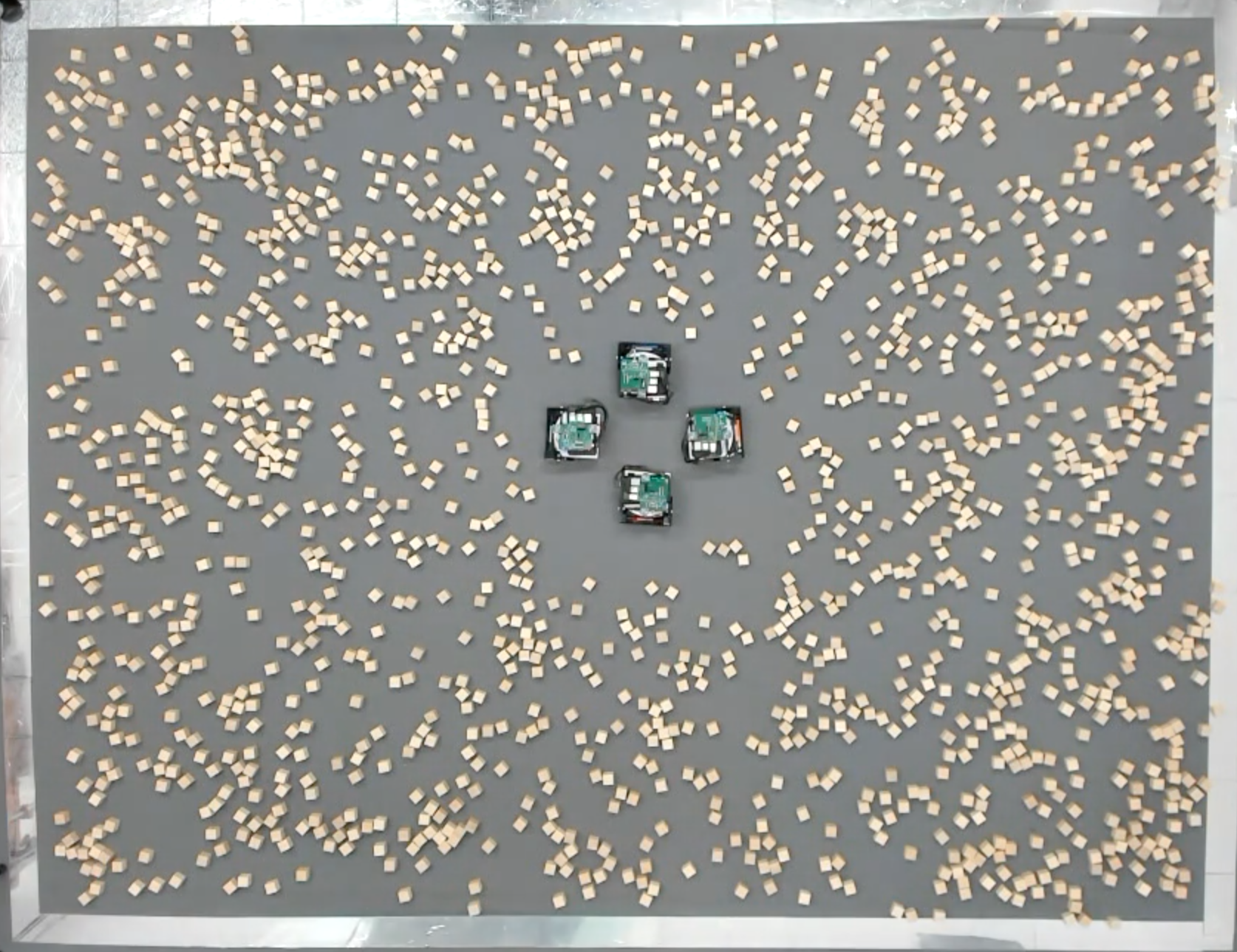}} \hspace{0.25em}
    \subfloat[\label{fig:gradientWorld}]{ \includegraphics[width=0.4\linewidth]{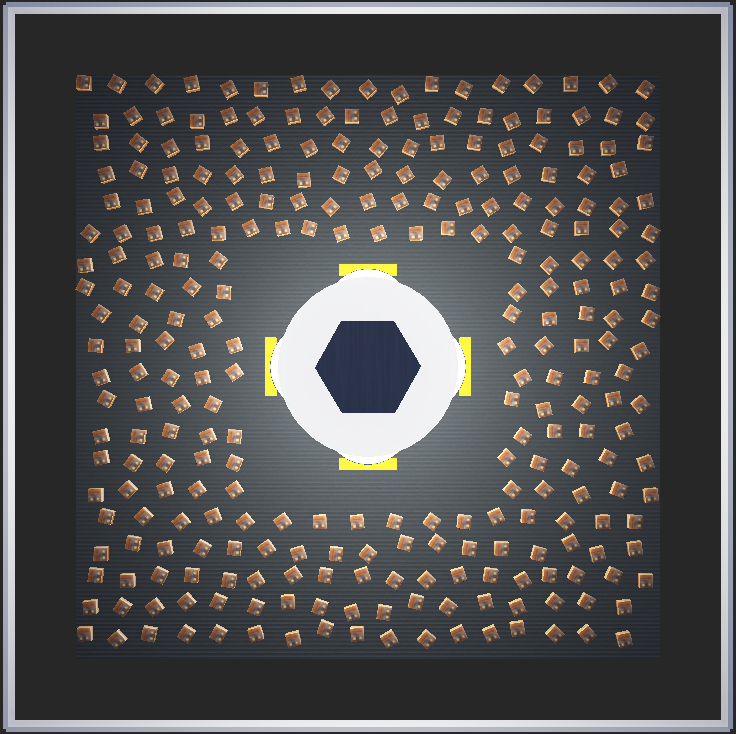}} 
    \caption{Standard arena in reality~(a) and gradient arena in simulation~(b). Robots have fixed starting positions in all scenarios. Boxes are distributed randomly at the beginning of the evolutionary run. Reprinted from~\cite{kaiser2020}.}
\end{figure}

We use a swarm of $N=4$~extended Thymio~II robots (see Sec.~\ref{section:RealRobot}) and one master robot. 
The robots are initially placed in the arena center. 
We use a $1.1\,\text{m} \times 1.1\,\text{m}$ arena that allows for fast simulation in Webots and a real arena of $2.2\,\text{m} \times 2.9\,\text{m}$.
Wooden cubes (boxes) of $2.5\,\text{cm} \times 2.5\,\text{cm} \times 2.5\,\text{cm}$ that weigh about $10\,\text{g}$ each are randomly distributed in the arena which results in a box density of ca.~$14\%$ (i.e., $\approx 220 \frac{\textrm{boxes}}{\textrm{sqm}}$). 
We adjust the weight in simulation to $2\,\text{g}$ to compensate for differences between simulation and real world that we determined in preliminary investigations. 
The boxes are too low to be detected by the horizontal IR sensors and are only detected by the pressure sensors.  
We run experiments with uniform light conditions both in simulation (standard arena) and in the real arena (Fig.~\ref{fig:realArena}). 
Here, the light sensors are not used. 
In simulation, we run additional experiments in an arena (gradient arena) that has a simulated light bulb above its center (Fig.~\ref{fig:gradientWorld}), resulting in decreasing light intensity towards the arena boundaries to investigate the influence of light on the emerging behaviors.

We prevent robot damage by adding a hardware protection layer. 
The detection of too close obstacles with a robot's horizontal IR sensors triggers an escape behavior. 
We prevent robots from leaving the arena by forcing them to turn on spot when they detect the arena boundaries with their ground IR sensors. 
Additionally, walls in distance to the arena boundaries force the robots to drive back into the arena by triggering the robots' escape behavior. 
In the real arena, robots are either kept in the arena by the rim formed by mirror film or, if necessary, the experimenter will trigger the escape behavior to force robots back into the arena. 
To avoid motor damage, we limit the maximum amount of pushed boxes to about 10~boxes. 
When a higher pushing force is detected by the pressure sensor, the robot turns on spot away from the boxes. 
As we cannot measure pushing forces when robots are driving backwards, we stop them after $9\,\text{s}$ of constantly driving backwards.  
We reset this limit each time positive motor values occur. 

\subsubsection{Simulation Results} \label{section:REALbpSim}

We reach median best fitness of at least $0.97$ for the evolutionary runs and of maximally $0.58$ for the randomly generated ANN pairs in both arenas, that is, evolution improves fitness over generations.
As before, we analyze the diversity of emergent behaviors in post-evaluation of the best evolved individuals and compare it to the randomly generated ANN pairs. 
A video is online~\cite{kaiser2021a}. 
We classify the behaviors based on distance covered by robots and box displacement.
The distance covered by robots
\begin{equation}
   d_R = \frac{1}{N} \sum^{N-1}_{n=0}\sum^{T-1}_{t=0}||l_n(t+1)-l_n(t)||_2\;\;,
   \label{equ:robomovement}
\end{equation}

is the mean accumulated robot displacement of the $N$ robots over runtime $T$ with positions $l_n(t)$ and $l_n(t+1)$ of robot~$n$.  
Robots can cover a maximum distance $d_R$ of $12.6\,\text{m}$ when constantly driving with maximum linear speed. 
We define box displacement $d_B$ as the mean Euclidean distance between the starting positions~$l_b(0)$ and final positions~$l_b(T)$ of $B$ boxes. 
The theoretical maximum displacement of a box is the arena's diagonal.

\begin{figure}[t]
\centering
    \subfloat[]{\includegraphics[width=0.8\linewidth]{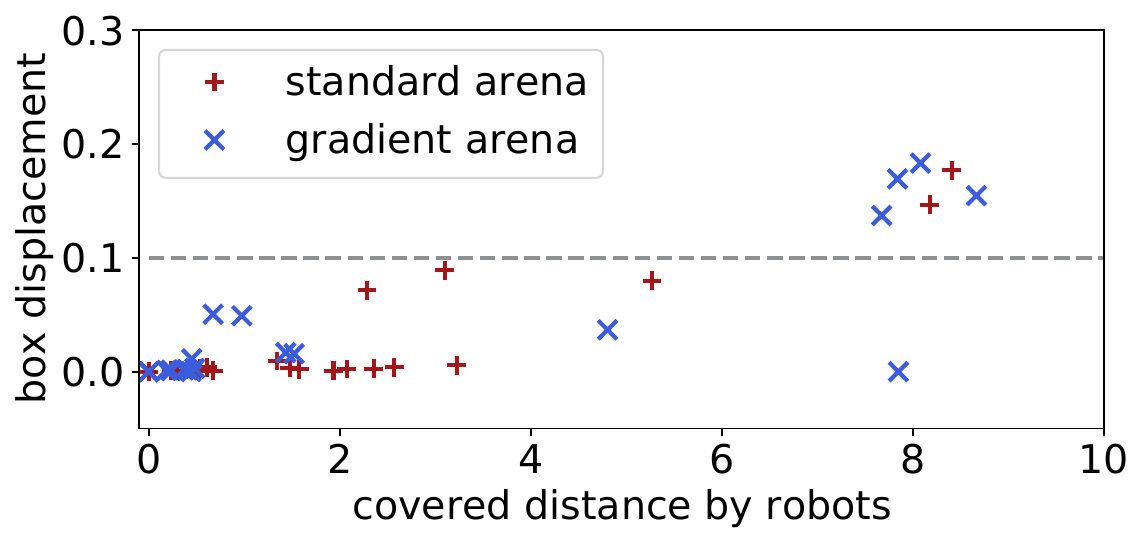}} \\
    \subfloat[]{\includegraphics[width=0.8\linewidth]{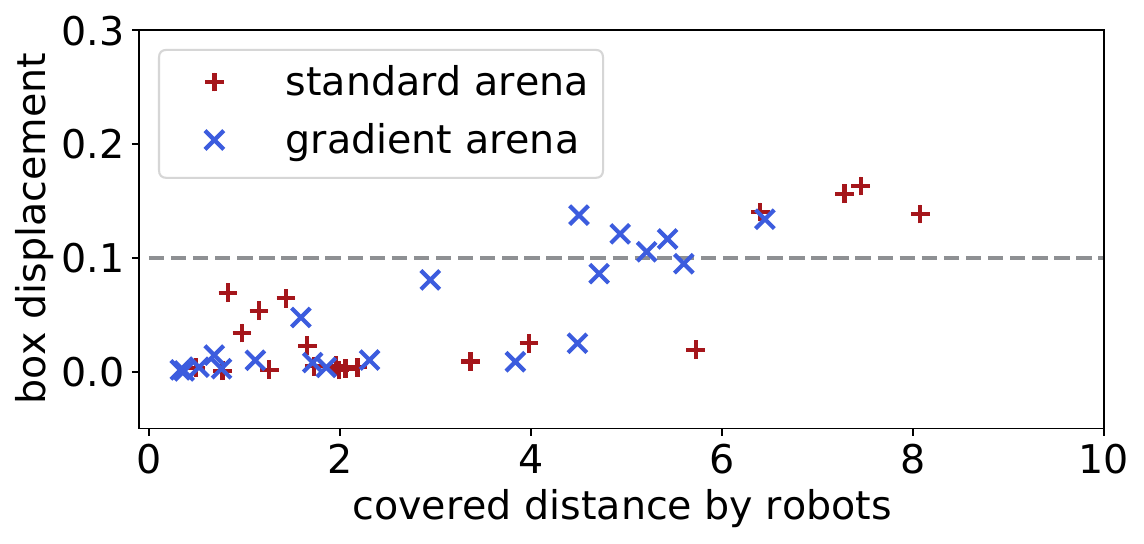}}
    \caption{Realistic simulation - object manipulation: distance covered by robots~$d_R$ (Eq.~\ref{equ:robomovement}) vs. box displacement~$d_B$ for (a)~the best evolved individuals~\cite{kaiser21,kaiser2020} and (b)~the randomly generated individuals~\cite{kaiser2021a} with 20~independent evolutionary runs per arena size. The dashed gray line marks a threshold between behaviors leading to the pushing of boxes and other behaviors. \label{fig:rot_clust2}}
\end{figure}

\begin{figure}[t]
    \centering
    \subfloat[\label{fig:turning}]{\includegraphics[width=0.4\linewidth]{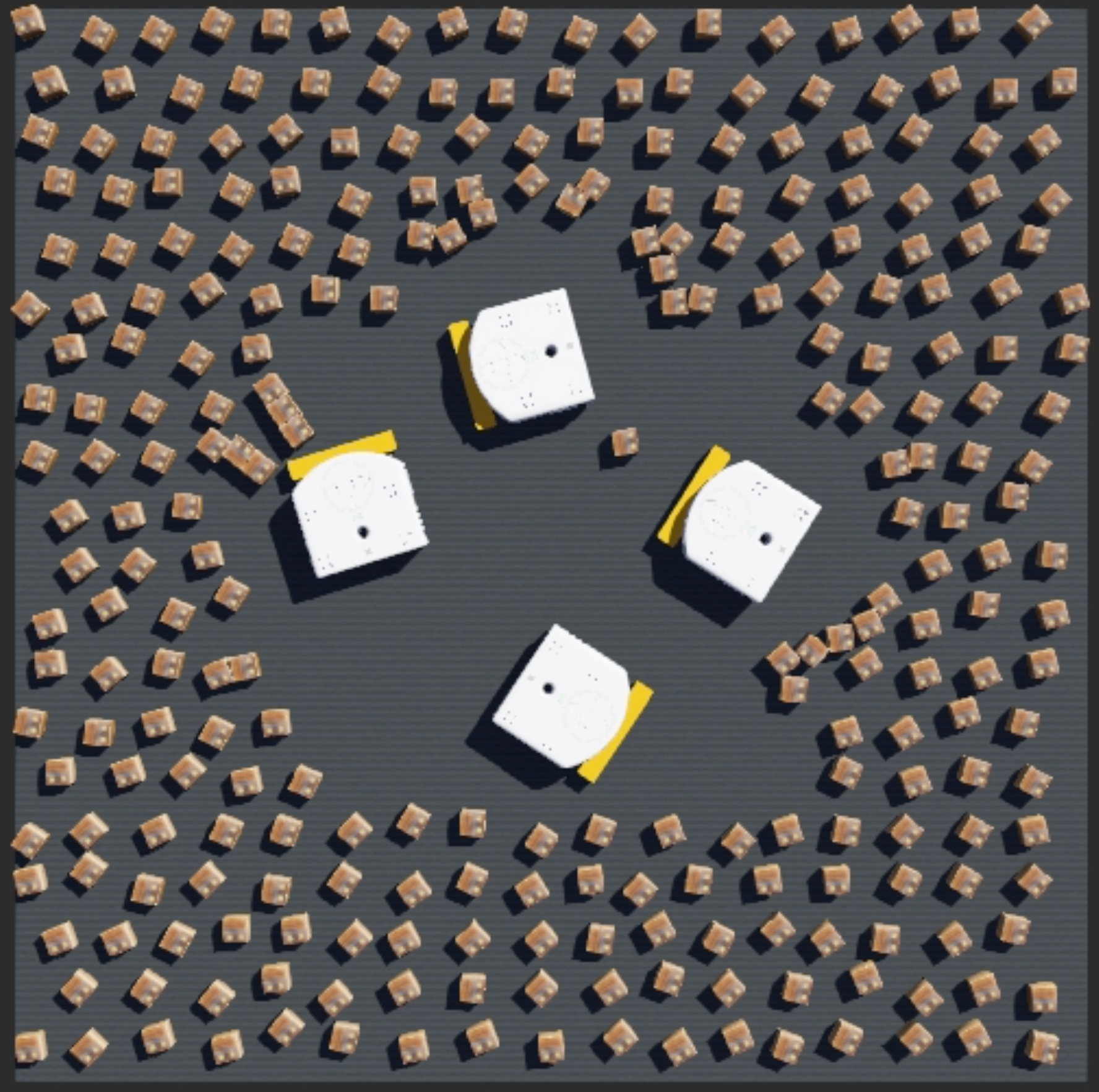}}\hspace{0.1em}
    \subfloat[\label{fig:push}]{\includegraphics[width=0.415\linewidth]{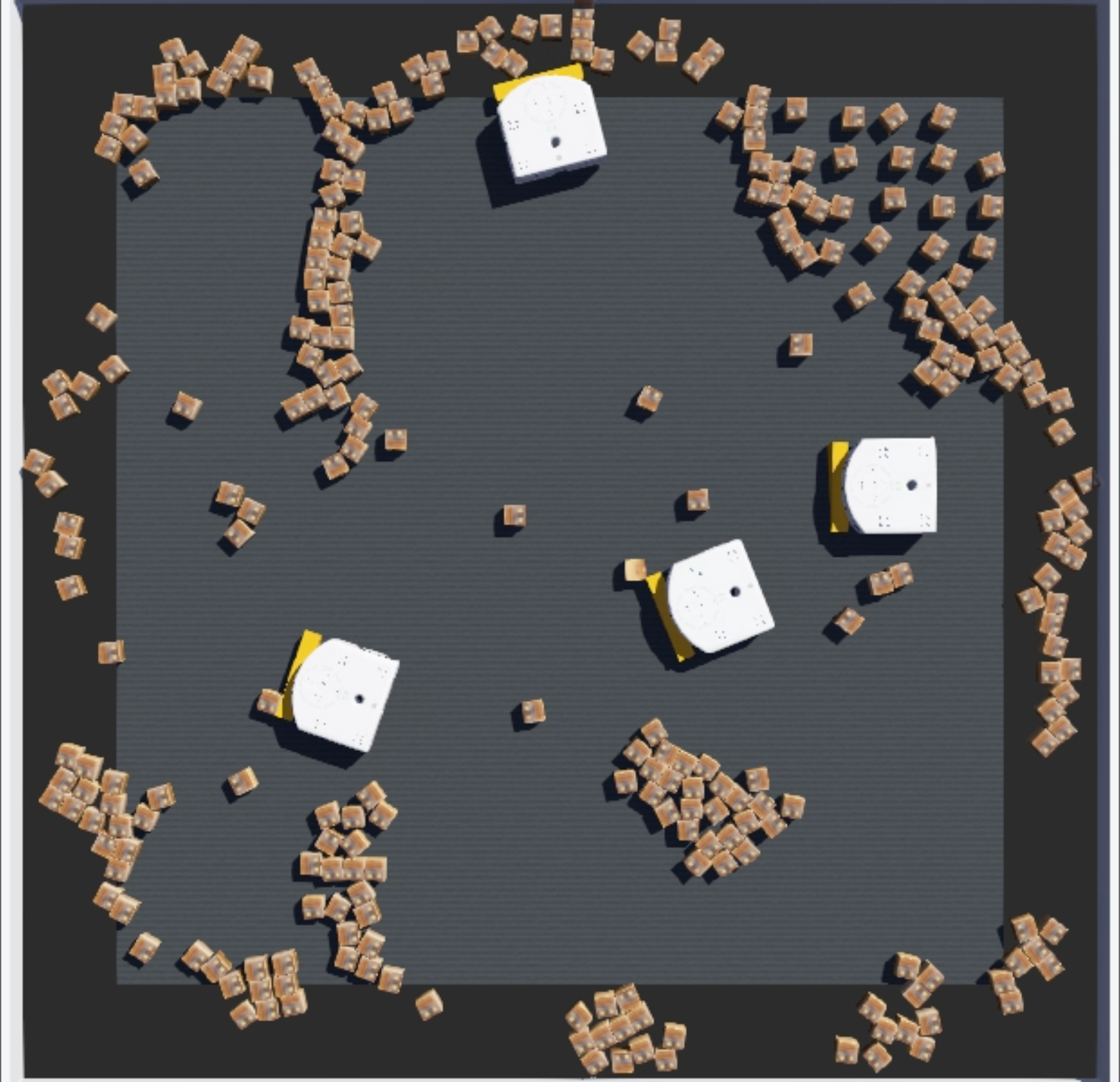}}
    \caption{Realistic simulation - object manipulation: final robot and box positions for (a)~circling and (b)~pushing of boxes at the end of the post-evaluation runs
    in simulation. Reprinted from~\cite{kaiser2020}.
    \label{fig:obj_man}}
\end{figure}

We find three classes of behaviors: circling, reverse driving, and behaviors that lead to the pushing of boxes. 
By qualitative analysis, we find that a threshold of $0.1$ in box displacement~$d_B$ distinguishes behaviors that lead to the pushing of boxes from behaviors with limited or no box manipulation, see Fig.~\ref{fig:rot_clust2}. 
The majority of behaviors in evolution and of the randomly generated ANN pairs lead to circling (robots go in small circles). 
Robots distribute themselves by using an emergent obstacle avoidance behavior or the hardware protection's escape behavior which leads to the pushing of few boxes. 
In the evolutionary runs, we also find few cases were robots follow each other (`circle dance').
Reverse driving robots drive backwards until they are stopped by hardware protection. 
This leads to a static, easily predictable environment and only little pushing of boxes. 
Except for randomly generated ANN pairs in the gradient arena, behaviors leading to the pushing of boxes exploit the hardware protection's boundary avoidance behavior to execute a random walk behavior while pushing boxes through the arena. 
Small box clusters form (Fig.~\ref{fig:push}) probably due to hardware protection as robots turn away when they exceed the threshold of maximum pushed boxes.
By contrast, randomly generated ANN pairs in the gradient arena push boxes while driving large circles and trying to avoid each other. 
This behavior leads also to shorter covered distances $d_R$ than a random walk behavior and is probably hard to predict.

\begin{figure}[!t]
    \centering
     \subfloat[\label{fig:meansensor}]{
    \includegraphics[width=0.8\linewidth]{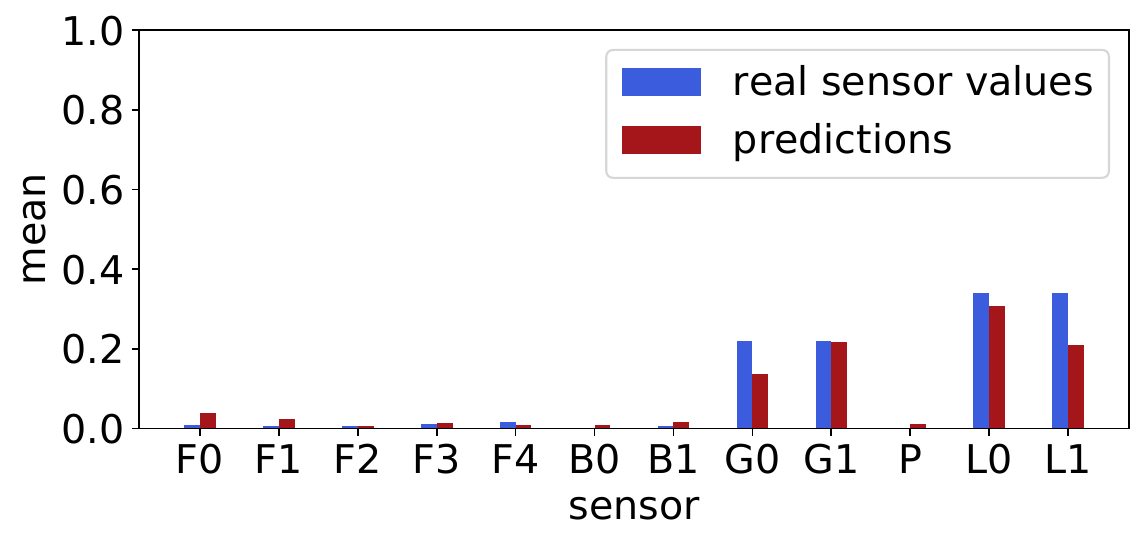}}\\
    \subfloat[\label{fig:nontrivialpred}]{\includegraphics[width=0.8\linewidth]{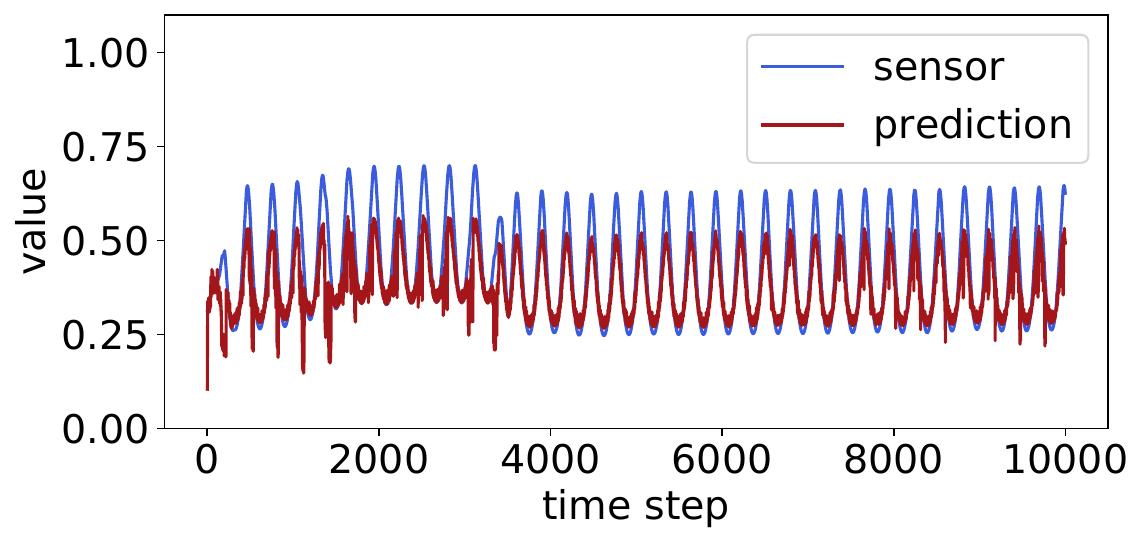}} 
    \caption{Realistic simulation - object manipulation: mean predicted and real sensor values~(a) of one behavior that lets robots drive in circles in the gradient arena with front IR sensors $F0, \dots, F4$, back IR sensors $B0$ and $B1$, ground IR sensors $G0$ and $G1$, pressure sensor $P$, and light sensors $L0$ and $L1$ and (b)~non-trivial predictions of L0. Reprinted from~\cite{kaiser2020}. \label{fig:sample19_cone_dense}}
\end{figure}

As before (see Sec.~\ref{section:SApredictions}), sensor value predictions and real sensor values may explain the best evolved behaviors.  
Fig.~\ref{fig:meansensor} illustrates mean sensor values and predictions of a circling behavior in the gradient arena as a representative example. 
Overall, robots neither sense nor predict nearby robots (i.e., $F0,\dots,F4,B0,B1 \approx 0$) except for the `circle dance' behavior.
Here, robots detect other robots with their front left IR sensors. 
Furthermore, reverse driving behaviors lead to the detection of the outer arena walls with the back IR sensors. 
The ground IR sensors ($G0, G1$) lead to real and predicted values matching the reflected light from the arena's carpet. 
Pressure sensor values ($P$) and predictions are low as hardware protection forces robots to turn when pushing more than~$10$ boxes, that is, $P > 0.25$. 
All sensors used in the standard arena setting allow for trivial predictions.  
But in the gradient arena, light intensity fluctuates based on the distance from the arena's center and thus the light sensors ($L0, L1$) do not allow for trivial predictions. 
We find rather complex and adapted sensor predictions in rare cases. 
Fig.~\ref{fig:nontrivialpred} illustrates oscillations of light intensity that are triggered by repetitive behaviors, such as circling. 
We find those oscillations for light intensity levels lower than detected by the sensors for most behaviors, but in few cases we find sophisticated predictor outputs as depicted in Fig.~\ref{fig:nontrivialpred} for $L0$. 
Furthermore, reverse driving behaviors lead to constant, low light intensity as robots are usually stopped by hardware protection at the arena's boundaries.

\subsubsection{Real Robot Experiments} \label{section:REALbpReal}

\begin{figure}[!t]
    \centering
    \includegraphics[width=0.85\linewidth]{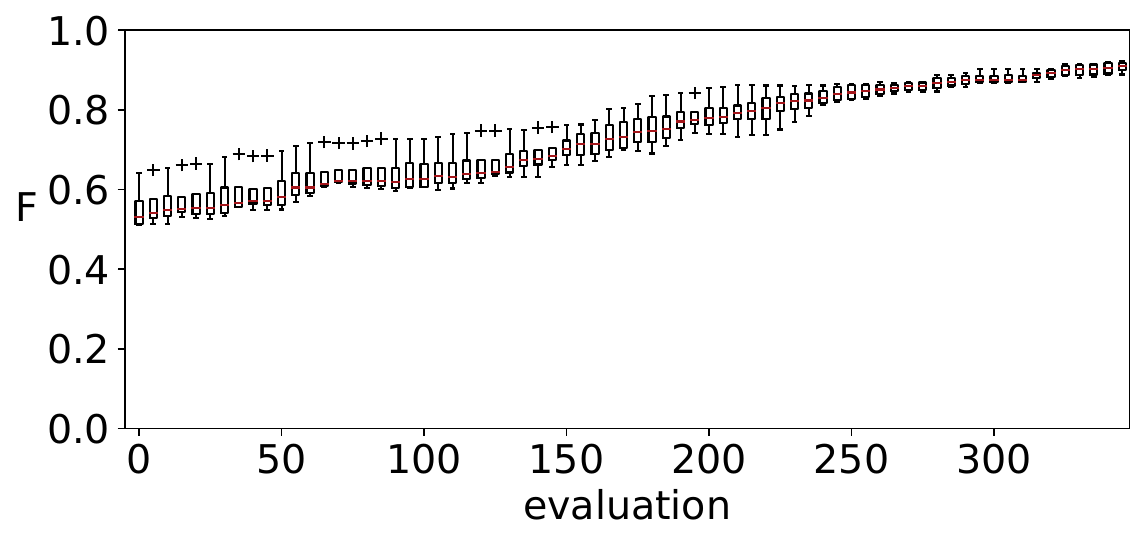}
    \caption{Real robot experiments with 4 robots: current best fitness over evaluations in the real arena of 4~runs. We print boxes for every 5 evaluations for a clearer illustration. Red bars indicate the median. Reprinted from~\cite{kaiser2020}. \label{fig:fitness_curve_real}}
\end{figure}

Fig.~\ref{fig:realArena} shows the initial robot and box positions in the real robot experiments (see Sec.~\ref{section:REALbpSetup}). 
We run five independent evolutionary runs with real robots with a total duration of approx. 65~min each.  
One of those five runs terminated early due to connection errors and is excluded from our evaluation. 
Fig.~\ref{fig:fitness_curve_real} shows the increase of best fitness over 350~evaluations reaching a median best fitness of~$0.91$ in the last generation. 
A video of one complete evolutionary run is online~\cite{kaiser2020}.
One run leads to robots driving backwards, but most  of the runs (i.e., three out of four) lead to robots driving in circles as in simulation. 
This allows for easy sensor predictions as robots clear space from boxes early in the evolutionary runs. 
Sensor value predictions are similar to the results in simulation.
All horizontal IR sensors ($F0, \dots,F4$, $B0$, $B1$) are approximately zero most of the time and thus predictions are also low. 
The real values and predictions for the ground IR sensors match the light reflected from the arena's carpet which is approx. $0.4$ here. 
As robots do not push boxes in the emergent circling behaviors, the pressure sensor~$P$ has low values and predictions.

\section{Discussion and Conclusion} \label{section:conclusion}

We have shown that our approach successfully evolves reactive swarm behaviors by minimizing surprise (i.e., maximizing prediction accuracy) in simulation and in real robot experiments. 
This approach may not explicitly push towards exploration or interesting behaviors~\cite{friston12}, but a careful configuration of the environment (e.g., including dynamic components) and the robot (e.g., sensor model and possible actions) enables the emergence of a variety of behaviors. 
We used simple environments that allowed for interaction only between robots and more complex environments with passive objects that could be manipulated by robots. 
Simple swarm behaviors relying on robot-robot interactions emerged as well as swarm behaviors with robot-environment interactions. 
In experiments with real robots, we have shown that the minimize surprise approach can also bridge the reality gap by directly evolving controllers online in the real world.
This online evolution approach has the potential to implement a multi-robot system that stays fully adaptive to dynamic environments in an open-ended way. 
Our first experiments showed that robots adapt to the changes in the environment caused by their manipulation of passive objects (i.e., boxes). 
In future work, we aim to extend the real robot experiments to evolve increasingly complex swarm behaviors and to investigate the adaptive power of our approach in more detail.
We also plan to objectively classify emergent behaviors using methods of unsupervised machine learning. 
In addition, we want to go beyond reactive swarm behaviors and try to generate more complex behaviors.
For example, predictor outputs can be used for action selection basing behaviors on the current and the expected future state of the environment.

In our in-depth study of a self-assembly scenario in simple simulations, we showed that our approach is effective and outperforms random search.  
The emergence of behaviors depends on swarm density, but the best individuals are robust to damage and scalable with swarm density.
This scalability allows for fast evolution of swarm behaviors with low computational resources that can then be deployed on larger swarms. 
While novelty search leads to more behavioral variety, we find solutions of higher quality using minimize surprise.
The drawback of lower solution quality in novelty search may be overcome by the use of quality-diversity algorithms. 
We want to minimize surprise with MAP-Elites to evolve diverse and high-quality solutions with a task-independent fitness function in future work. 

We pushed emergence towards desired behaviors by predefining some or all sensor predictions. 
This provides an intuitive way to specify (partially) task-dependent fitness functions. 
Our minimize surprise approach can be run either with complete freedom (i.e., no predefined sensor predictions), as a special kind of task-dependent approach (i.e., predefining all sensor predictions), or anything in between with partially predefined sensor predictions. 

In all our experiments so far, we used simple, fixed ANN structures. 
We aim to investigate the influence of
different state-of-the-art methods, such as LSTMs~\cite{hochreiter1997} or Transformers~\cite{NIPS2017_3f5ee243}, and the ANN topology on the emergent behaviors in future work. 
This also includes combining actor and predictor into one ANN as done in the work by Nolfi and Spalanzani~\cite{nolfi_learning_2002} and evolving both weights and network structure, for example, using NEAT~\cite{stanley:ec02}. 

In future work, we want to evolve large-scale multi-robot system behaviors with minimize surprise in dynamic, high entropy environments to prove that our approach can lead to open-ended life-long adaptation. 
We expect that the minimize surprise approach will be a useful tool to govern the increasing complexity of our future autonomous systems and an instrumental contribution to increase the robustness, adaptivity, and scalability of future real-world multi-robot systems.

\bibliographystyle{IEEEtran}  
\bibliography{references}

\begin{IEEEbiographynophoto}{Tanja Katharina Kaiser} received a MSc degree in computer science from Technische Universit\"at Berlin, Germany in 2017. She is currently a doctoral candidate at the Universit\"at zu L\"ubeck, Germany. Her main research interests are swarm robotics and evolutionary robotics. 
\end{IEEEbiographynophoto}

\begin{IEEEbiographynophoto}{Heiko Hamann}
received his doctorate in engineering from the University of Karlsruhe, Germany in 2008. He did his postdoctoral training in swarm robotics, modular robotics, and evolutionary robotics at the Zoology department of the University of Graz, Austria. He was assistant professor of swarm robotics at the University of Paderborn, Germany from 2013 until 2017. Since 2017 he is professor for service robotics at the University of L\"ubeck, Germany. His main research interests are swarm intelligence, swarm robotics, evolutionary robotics, bio-hybrid systems, and modeling of complex systems.
\end{IEEEbiographynophoto}

\end{document}